\documentclass[11pt,letterpaper]{article}

\PassOptionsToPackage{table}{xcolor}

\usepackage{amsmath,amssymb,amsthm}
\usepackage{bm}
\usepackage{deepthink}

\usepackage[utf8]{inputenc}
\usepackage[T1]{fontenc}
\usepackage{booktabs}       
\usepackage{nicefrac}       
\usepackage{multirow}
\usepackage{wrapfig}
\usepackage{tabularx}
\usepackage{array}
\usepackage[font=footnotesize]{caption}
\usepackage{subcaption}
\usepackage{pifont}         
\usepackage{algorithm}
\usepackage{algpseudocode}
\usepackage{url}

\let\deepthinkorigwrapfigure\wrapfigure
\let\enddeepthinkorigwrapfigure\endwrapfigure
\renewenvironment{wrapfigure}[2]{%
  \def\deepthink@wfplace{#1}%
  \edef\deepthink@wfplace{\deepthink@wfplace}%
  \if r\deepthink@wfplace\def\deepthink@wfplace{R}\fi
  \if l\deepthink@wfplace\def\deepthink@wfplace{L}\fi
  \expandafter\deepthinkorigwrapfigure\expandafter{\deepthink@wfplace}{#2}%
}{\enddeepthinkorigwrapfigure}

\usepackage[nameinlink]{cleveref}
\usepackage[numbers]{natbib}

\DeclareMathOperator*{\argmin}{arg\,min}

\definecolor{lightblue}{RGB}{100,160,210}
\definecolor{lightorange}{RGB}{230,150,110}

\newcommand{\revise}[1]{#1}
\newcommand{\reviseq}[1]{#1}
\newcommand{\rebuttal}[1]{#1}

\title{The Linear Geometry of Interpretable Tokens: Jailbreaking Attacks and Defenses for Unlearned Diffusion Models}

\newcommand{\corrauth}{\textsuperscript{\ddag}}

\authorblock{
  \href{https://chicychen.github.io/}{\textbf{Siyi Chen}},
  \href{https://damon-demon.github.io/}{\textbf{Yimeng Zhang}}\textsuperscript{1},
  \href{https://lsjxjtu.github.io/}{\textbf{Sijia Liu}}\textsuperscript{2},
  \href{https://qingqu.engin.umich.edu/}{\textbf{Qing Qu}}\corrauth
}

\affiliation{
  University of Michigan \quad $\cdot$ \quad 
 \textsuperscript{1}Amazon \quad $\cdot$ \quad
  \textsuperscript{2}Michigan State University
}

\authornote{
  \corrauth\ Corresponding author
}

\abstracttext{
\noindent Diffusion models excel at generating high-quality images but can memorize and reproduce harmful concepts when prompted. Although fine-tuning methods have been proposed to unlearn a target concept, they struggle to fully erase it while maintaining generation quality on other concepts, leaving models vulnerable to jailbreak attacks. Existing jailbreak methods demonstrate this vulnerability but offer limited insight into how unlearned models retain harmful concepts, limiting progress on effective defenses. \rebuttal{In this work, we show that a coherent, interpretable, attack-accessible linear residual of the erased concept can be recovered in the token embedding space, and that both an attack and a defense follow from this structure.} We introduce \textit{SubAttack}, a novel jailbreaking attack that \revise{reads out this subspace by learning} an orthogonal set of attack token embeddings, each being a linear combination of human-interpretable textual elements, revealing that unlearned models still retain the target concept through related textual components. Furthermore, our attack is also more powerful and transferable across text prompts, initial noises, and unlearned models than prior attacks.  \revise{Conversely, projecting out the same subspace yields} \textit{SubDefense}, a lightweight plug-and-play defense mechanism that suppresses the residual concept in unlearned models. SubDefense provides stronger robustness than existing defenses while better preserving safe generation quality. Extensive experiments across multiple unlearning methods, concepts, and attack types demonstrate that our approach advances both understanding and mitigation of vulnerabilities in diffusion unlearning.
}

\correspondence{\;\url{siyich@umich.edu}}
\date{\today}
\resources{\href{https://chicychen.github.io/IJD/}{\quad \textbf{Project page}} \quad $\cdot$ \quad \href{https://github.com/ChicyChen/interpretable-jailbreak-defense}{\textbf{Code}}}

\headerlogo{Deepthink_landscape_do_not_delete_compressed.png}{https://deepthink-umich.github.io}

\begin{document}

\makeDeepthinkHeader

\vfill
\begin{center}
  \includegraphics[width=\textwidth]{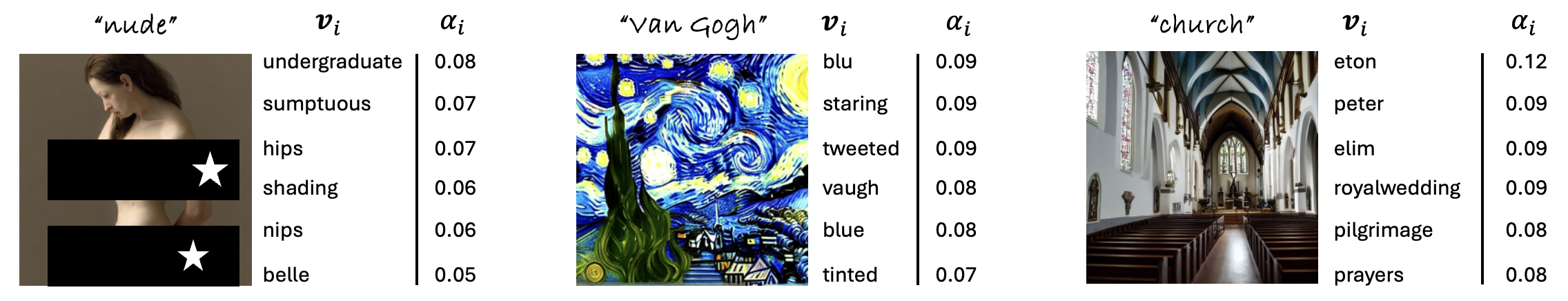}
  \vspace{0.25em}

  \parbox{0.94\textwidth}{\centering\small
    \textbf{What remains after unlearning?}
    SubAttack recovers a human-interpretable, attack-accessible linear
    residual of supposedly erased concepts in the token embedding space;
    SubDefense projects out the recovered directions to strengthen
    unlearned models against jailbreaks.}
\end{center}
\vfill

\newpage
{\small 
\tableofcontents}


\section{Introduction}\label{sec:intro}


Diffusion models (DMs) have recently emerged as a powerful class of generative models, capable of producing diverse and high-quality content such as images \citep{ddpm}, videos \citep{khachatryan2023text2video}, and protein structures \citep{protein}. Notably, Text-to-Image (T2I) diffusion models \citep{stablediffusion,Ramesh2022HierarchicalTI,DeepFloydIF,zhang2024id,zhang2024unlearncanvas} have gained significant popularity for their ability to generate high-fidelity images from user-provided text prompts. However, the remarkable generative capabilities of these models also raise significant concerns regarding their safe deployment. For example, users can exploit carefully crafted
text prompts to induce these models to generate unethical or harmful content, such as nude or violent images or copyrighted material
\citep{schramowski2022safe}.

To address such safety concerns without refiltering the huge dataset and retraining the full model, \textit{Machine Unlearning} (MU) methods have recently been developed for ``erasing'' a harmful concept directly from the pretrained models. For instance, a wide range of methods  \citep{esd,gandikota2024uce,zhang2023fmn,lyu2023spm} seek to unlearn harmful content in pretrained DMs by fine-tuning the model weights \citep{musurvey}. However, the key challenge of preserving the generation quality of safe content limits unlearned DMs from removing even a \textit{single concept} completely. This limitation becomes more severe under \textit{jailbreaking attacks} \citep{zhang2024unlearndiff,pham2024circumventing,chin2024prompting4debugging,tsai2024ringabell,zhuang2023a}, which can jailbreak
unlearned DMs to regenerate harmful content. For instance, UnlearnDiff \citep{zhang2024unlearndiff} crafts adversarial discrete text prompts, and CCE \citep{pham2024circumventing} leverages textual inversion \citep{gal2023textinverse} to execute jailbreaking attacks in embedding space. Due to the rising popularity of open-source models, as well as the risks of insider threats and model leakage, many recent studies \reviseq{\citep{zhang2024unlearndiff,pham2024circumventing,chin2024prompting4debugging}} adopt a {white-box} setting (i.e., full access to model weights) for safety evaluation. \reviseq{These white-box attacks} reveal that unlearned DMs remain vulnerable, highlighting the urgent need for robust \textit{defending} of these unlearned DMs by strengthening their robustness against attacks.


\begin{figure}[t]
    \centering
    \includegraphics[width=0.7\textwidth]{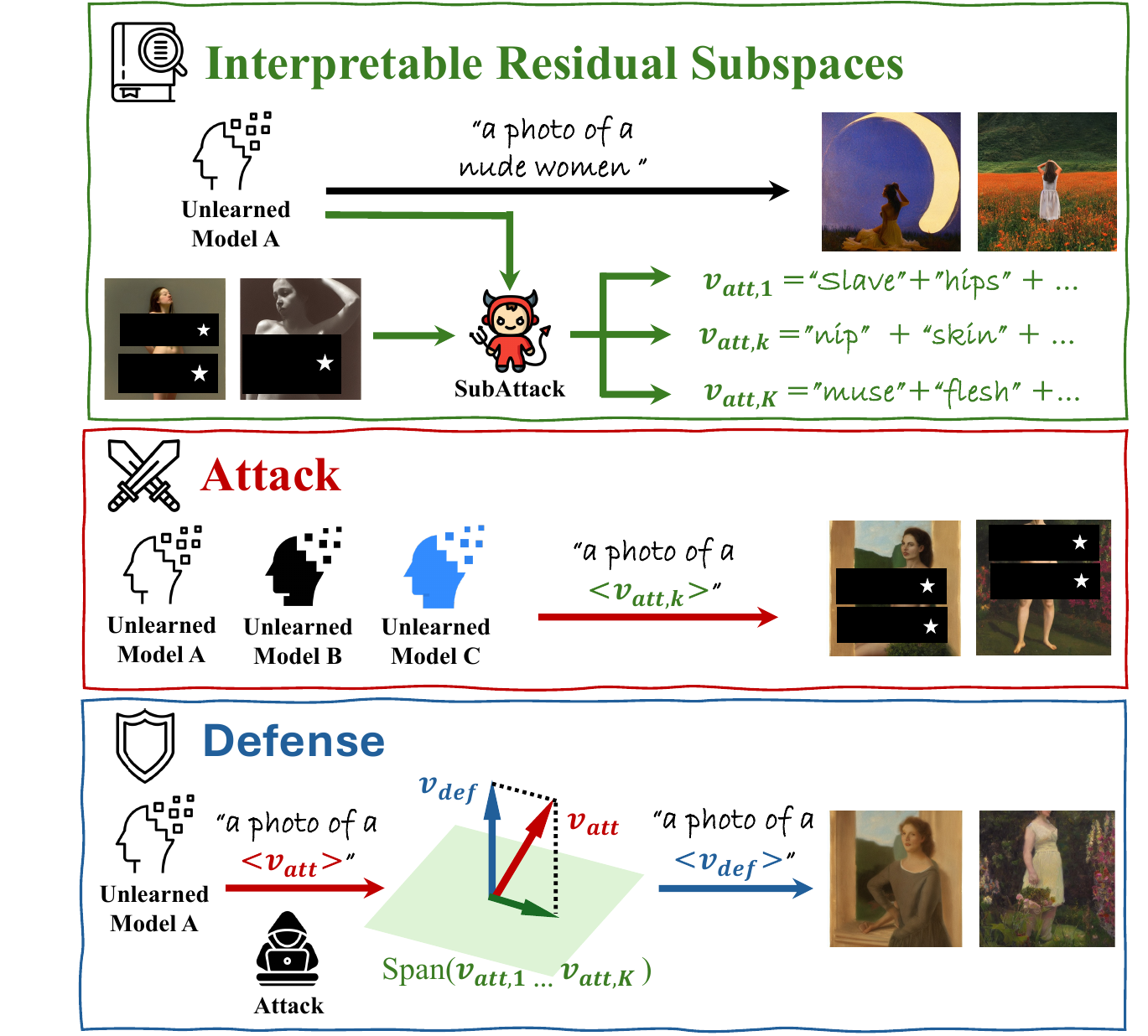}
    \caption{\reviseq{\textbf{An erased concept persists as an interpretable residual subspace.} SubAttack learns interpretable attack tokens $\{\bm v_{\texttt{att},k}\}$ spanning this subspace (top); \emph{reading it out} jailbreaks the unlearned model (middle), while \emph{projecting it out} yields SubDefense (bottom).}}
    \label{fig:teaser}
\end{figure}
It is not surprising that optimization-based, non-interpretable, and worst-case prompt perturbations can jailbreak unlearned DMs. However, despite leveraging white-box access, such approaches provide limited \textit{interpretability}, i.e., a human-understandable explanation of how a model's internal state drives the prediction of its behavior under intervention. Therefore, they offer little insight into how harmful concepts persist within the model, and these attacks fail to offer potential insights for defense strategies. \revise{Furthermore, while adversarial-finetuning approaches such as AdvUnlearn~\citep{zhang2024defensive}, Receler~\citep{receler}, and STEREO~\citep{srivatsan2024stereo} improve robustness by re-training model weights with an inner adversarial loop, lightweight \emph{plug-and-play, post-hoc refinement} defenses that operate on an already-unlearned model remain largely underexplored.} For instance, the RECE defense framework \citep{gong2024reliableefficientconcepterasure} focuses on improving a specific unlearned model (UCE \citep{gandikota2024uce}) against particularly adversarial attacks (i.e., UnlearnDiff). \revise{Extending such post-hoc refinement to a broader range of unlearned models and attack types remains a challenging problem.} These gaps motivate our central \textbf{question}: \textit{Can we design more interpretable jailbreaking attacks that also provide actionable insights for building robust defenses in unlearned DMs?}



%


Our work tackles this fundamental question by exploring the underlying linear structures in the token embedding space, taking advantage of the white-box setting. We introduce an effective, human-interpretable \textit{subspace attack method (SubAttack)}, which further inspires a \textit{subspace defense strategy (SubDefense)} broadly applicable to various unlearned models and attacks. The core idea is to learn an orthogonal set of attack token embeddings within the unlearned model for the harmful concept. Inspired by prior works \citep{chefer2024the, park2023thelinear, cunningham2023sparseautoencodershighlyinterpretable}, we optimize each attack embedding as a nonnegative linear combination of embeddings of existing concepts, and interpret the concept through the linear decomposition. \rebuttal{Related diffusion work also identifies semantic directions and subspaces \citep{hspace,discovering,zhu2023boundary} and formalizes concepts as subspaces in score-based generative models \citep{wang2023conceptalgebra}.} Leveraging our approach, we show that unlearned DMs associate the harmful concept with mixtures of other hidden concepts, thus retaining unintended harmful regeneration capabilities. These insights motivate our defense mechanism, which further mitigates the harmful concept from unlearned DMs by removing the learned attack token embeddings through orthogonal subspace projection.

Compared to prior methods, our SubAttack demonstrates strong empirical performance of efficiency and effectiveness, showing stronger transferability across text prompts, initial noises, and unlearned models. Our defense strategy can be seamlessly integrated into various unlearned models, improving robustness against different jailbreaking attacks while preserving higher generation quality than the baseline defense method \citep{gong2024reliableefficientconcepterasure}. 
A comprehensive discussion of related works is in \textbf{\Cref{appsec:related_works}}. In summary, this work makes the following \textbf{contributions}:

\begin{itemize}[leftmargin=*]
    \item \rebuttal{\textbf{Identifying interpretable residual subspaces in unlearned DMs.} We show that a coherent, interpretable, attack-accessible linear residual of the target concept can be recovered in the token embedding space. This does not assert that the concept's complete representation is exclusively linear or that the recovered subspace is unique.}
    \item \textbf{Subtask: effective and interpretable attacks on unlearned DMs.} Correspondingly, we propose \textit{SubAttack}, which can effectively jailbreak unlearned DMs by learning an orthogonal set of harmful token embeddings that span a linear low-dimensional subspace. It achieves higher ASR than existing baselines across diverse
concepts and unlearned models, and transfers reliably across prompts, initial noise, and models, exposing critical vulnerabilities in current unlearning methods.
    \item \textbf{Subdefense: robust and interpretable defending methods of unlearned DMs. } By projecting out the identified subspace, we introduce \emph{SubDefense}, a lightweight, plug-and-play refinement. SubDefense offers versatile, reliable defense across diverse unlearned models and attack vectors without degrading generation quality. Crucially, it composes modularly with existing unlearning frameworks, including adversarially fine-tuned baselines like STEREO.
\end{itemize}

\section{Preliminaries and Problem Statement}


%

\subsection{Preliminaries}
\begin{figure}[t]
    \centering
    \includegraphics[width=.85\linewidth]{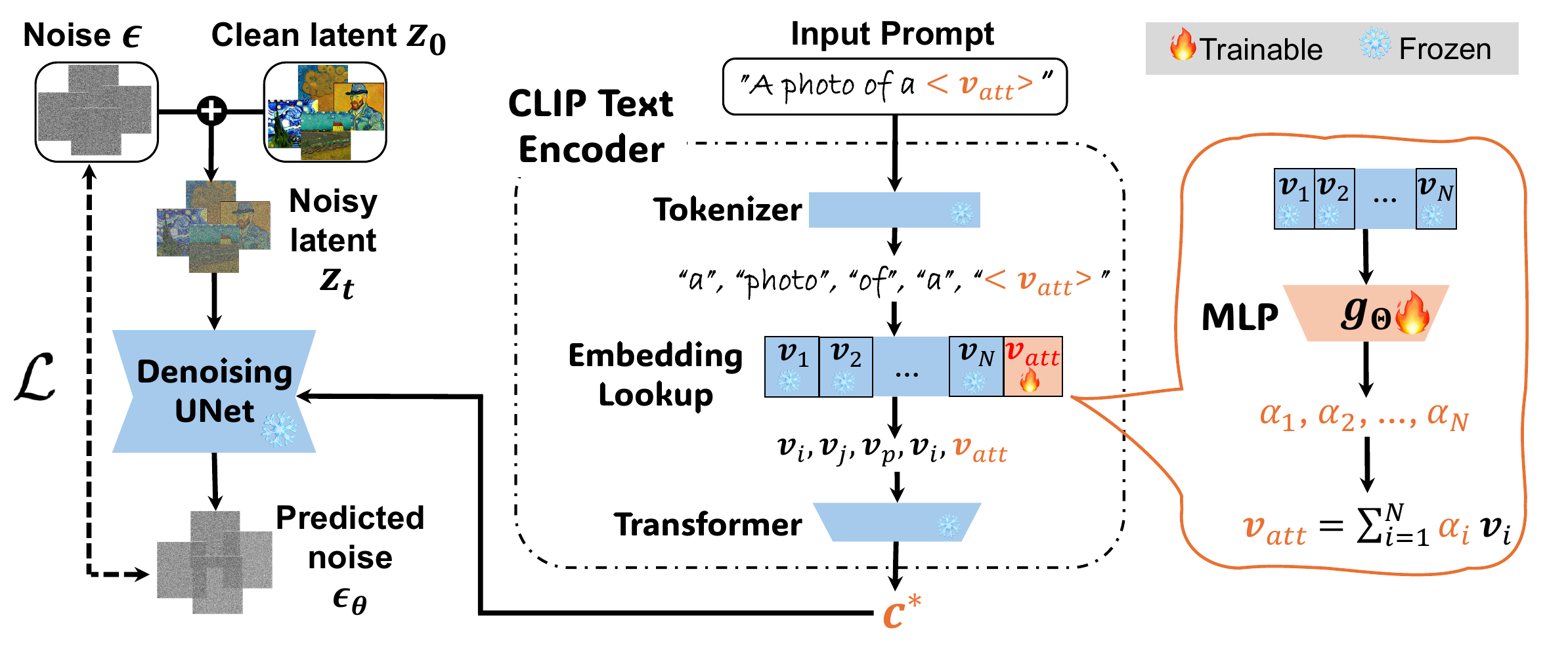}
    \centering \caption{\textbf{Learning one interpretable attack token embedding.} The learning process of one attack token embedding $\bm{v}_{\texttt{att}}$ for the concept ``Van Gogh'' is visualized. \textcolor{lightblue}{Blue} parts represent the frozen unlearned LDM, where, for simplicity, we omit the image encoder and decoder. 
    In \textcolor{lightorange}{orange} parts, it illustrates the learning mechanism for optimizing an MLP network to produce $\bm{v}_{\texttt{att}}$, which is a linear combination of the existing token embeddings.}
    \label{fig:method}
\end{figure}

\textbf{Overview of Latent Diffusion Models (LDMs).} T2I diffusion models have recently gained popularity for their ability to generate desired images from user-provided text prompts. Among these various T2I models, LDMs \citep{stablediffusion} is the most widely deployed DM, and has therefore become the primary focus of current machine unlearning methods. 
As shown in \textbf{\Cref{fig:method}}, for a given text prompt $\bm p$, LDM first encodes $\bm p$ using a pretrained CLIP text encoder \citep{clip} $\bm f(\cdot)$ to obtain the text embedding $\bm c = \bm f (\bm p)$. Then, the generation process begins by sampling a random noise $\bm{z}_T \sim \mathcal{N}(0, 1)$ in the latent space. After that, LDM progressively denoises $\bm{z}_T$ conditioned on the context $\bm{c}$ until the final clean latent $\bm{z}_0$ is achieved. Specifically, for each timestep $t = T, T-1, \dots, 1$, its denoising UNet, $\bm{\epsilon}_{\bm{\theta}}(\bm{z}_t \mid \bm{c})$, predicts and removes the noise to obtain a cleaner latent representation $\bm{z}_{t-1}$. The clean latent $\bm{z}_0$ is then decoded to an image with a pretrained image decoder. To train the denoising UNet $\bm{\epsilon}_{\bm{\theta}}(\bm{z}_t \mid \bm{c})$ in LDM, the denoising error is minimized:
\begin{equation}
\mathcal{L} = \mathbb{E}_{(\bm{z}, \bm c), t, \epsilon \sim \mathcal{N}(0,1)}\left[\left\|\bm \epsilon-\bm \epsilon_{\bm{\theta}}\left(\bm{z}_t \mid \bm c\right)\right\|_2^2\right],
\label{equ:loss}
\end{equation}
where $\bm{z}$ is the clean image latent encoded by a pretrained image encoder and $\bm c$ is the corresponding text embedding. Here, $\bm z_t=\sqrt{\alpha_t} \bm z+\sqrt{1-\alpha_t} \bm \epsilon$ is the noisy image latent at timestep $t$, and $\alpha_t > 0$ is a predefined constant.

\smallskip 
\noindent \textbf{CLIP text encoder and the token embedding space.} To control the generation process, a key component of LDMs is the pretrained CLIP text encoder $\bm{f}(\cdot)$. As illustrated in \textbf{\Cref{fig:method}}, the CLIP text encoder consists of three main components:
\begin{itemize}[leftmargin=*]
    \item \textbf{Tokenizer:} This module splits the text prompt $\bm{p}$ into a sequence of tokens, which can be words, sub-words, or punctuation marks. Each token is assigned a unique token ID from the CLIP text encoder’s predefined vocabulary.
    \item \textbf{Token Embeddings:}  These token IDs (e.g., $[i, j,\cdots]$) are then mapped to corresponding token embeddings $\bm{v}_i \in \mathbb{R}^d$ stored in the token embedding table. This process generates a sequence of token embeddings $[\bm{v}_i,\bm{v}_j,\cdots]$.
    \item \textbf{Transformer Network:} This network processes the sequence of token embeddings and encodes them into the final text embedding $\bm{c}$ that can guide the image generation process in LDMs.
\end{itemize}

Through optimizing \Cref{equ:loss}, LDM learns to associate activations in the text encoder with concepts in the generated images. Prior research has explored controlling generated content through manipulating activations in the text encoder. In particular, it has been identified that the token embedding space $\bm v$ plays a vital role in content personalization, where a single text embedding can represent a specific attribute \citep{gal2023textinverse} and the token embedding space can be utilized for linear decomposition of concepts \citep{chefer2024the}. 
Leveraging the expressiveness and interpretability of the token embedding space, we propose both jailbreaking attack and defense mechanisms, and discuss the problem setup in the following.

\subsection{Problem Statement and Setup}

\smallskip 
\noindent 
\textbf{\textit{Jailbreaking attacks}} are designed to evaluate the robustness of unlearned LDMs. Most existing diffusion unlearning studies focus on removing \textit{a single target concept} from each model. For example, given a prompt $\bm{p} =$ ``a photo of a [target concept] ...'', an unlearned LDM for this concept is expected to have difficulty generating the corresponding images. A jailbreaking attack, given an unlearned LDM as the \emph{victim model}, aims to manipulate the prompt to make the model regenerate the unwanted concept. There are majorly two kinds of attack setups: (\emph{i}) Adversarial prompt-based attacks \citep{zhang2024unlearndiff,chin2024prompting4debugging,tsai2024ringabell,zhuang2023a} optimize an \textit{adversarial text prompt} $\bm{p}_{\texttt{att}}$ and append it to $\bm{p}$. (\emph{ii}) Embedding-based attacks \citep{pham2024circumventing} learn an \textit{attack token embedding} $\bm{v}_{\texttt{att}}$, register it as a new token $<\bm{v}_{\texttt{att}}>$, and modify the prompt by replacing the [target concept] with this token. Our attack follows the second setup, but is explicitly designed to be interpretable through linear constraints while achieving stronger attack performance. Moreover, apart from access to the unlearned LDM, existing attacks generally require either the original LDM or images containing the target concept; in our setup, we assume access to the images ($\bm z_0$ in \textbf{\Cref{fig:method}}).

\smallskip 
\noindent 
\textbf{\textit{Defense}}, in contrast, seeks to protect an \textit{unlearned LDM} from new jailbreaking attacks. Once a defense strategy is applied, it should prevent the model from regenerating harmful concepts even under \textit{future attacks}, while preserving its ability to generate harmless content. For example, RECE \citep{gong2024reliableefficientconcepterasure} further modifies the denoising UNet of the unlearned model UCE \citep{gandikota2024uce} to defend against adversarial attacks \citep{zhang2024unlearndiff}. In this work, we propose a defense strategy that safeguards the token embedding space and can be seamlessly integrated into existing unlearned LDMs. Our objective is to provide a broadly applicable defense mechanism that enhances robustness across diverse unlearned models when confronted with new attacks.



\textit{\textbf{Interpretability}} refers to providing a compact, human-understandable description of how a model’s internal components drive its behavior \citep{chefer2024the,Zou2023RepresentationEngineering,bereska_mechanistic_2024}. Such a description is crucial as it allows testable predictions under controlled interventions. In our setting, interpretability means that attack embeddings can be explained as recognizable words or semantic units rather than opaque vectors. While many existing methods are empirically effective, they lack such interpretability, making it difficult to understand how harmful concepts persist or how to control the robustness of unlearned models. Our work addresses this gap by developing attack and defense methods grounded in a linear, interpretable structure.

\smallskip 
\noindent \textbf{Notations.} Before introducing our method, we define the following projection operators. Specifically, given vector 
$\bm z$, for a vector 
$\bm v$, let $ \operatorname{Proj}_{\bm v}(\bm z)$ denote the projection of $\bm z$ onto $\bm v$. For a matrix $\bm V$, let $ \operatorname{Proj}_{\bm V}(\bm z)$ denote the projection of $\bm z$ onto the subspace spanned by the columns of $\bm V$. Formally, these operators are given by
\begin{align*}
    \operatorname{Proj}_{\bm v}(\bm z) := \frac{ \bm v \bm v^\top }{ \|\bm v\|_{2}^2 } \bm z,  \;\operatorname{Proj}_{\bm V}(\bm z) := \bm V (\bm V^\top \bm V)^{-1} \bm V^\top \bm z.
\end{align*}

\section{Subspace Attacking and Defending Methods} \label{sec:main}

This section introduces our subspace attacking and defending methods for 
LDMs.
In \Cref{subsec:attack}, we explore the token embedding space to develop an interpretable and effective attack method (SubAttack) by learning a sequence of attack token embeddings orthogonal to each other\revise{, which together span an interpretable subspace along which the erased concept persists}. \revise{Removing this same subspace then yields a defense strategy (SubDefense) in \Cref{subsec:defense}: by orthogonal subspace projection of the learned attack token embeddings, it can effectively defend against various jailbreaking attacks. In this way, the attack and the defense are two operations---reading out and projecting out---on the same residual subspace.}

\subsection{Subspace Attacking: \textit{SubAttack}}\label{subsec:attack}

Before we introduce our subspace attacking method, let us build some intuition of how to learn a single interpretable attack token embedding $\bm v_{\texttt{att}} \in \mathbb R^d$. Based on this, we will then show how to iteratively learn a sequence of orthogonal attack token embeddings through \textit{deflation}, i.e., removing already computed embeddings. 

\begin{figure*}[t]
    \centering
    \includegraphics[width=.95\linewidth]{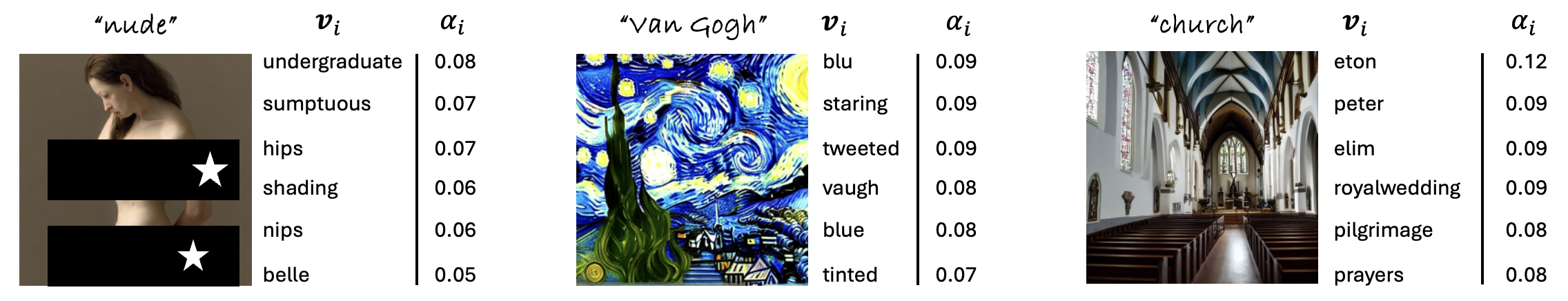}
    \centering \caption{\textbf{Interpreting the attack token embeddings for concept ``nudity'', ``Van Gogh'', and ``church''.} Tokens with the largest $\alpha_i$ are words associated with the target concept. For example, top tokens for ``church'' are activities conducted in the church, or names from the Bible.}
    \label{fig:alpha}
\end{figure*}

\subsubsection{Single-Token Embedding Attack}\label{subsec:single-attack}

We aim to learn a token embedding $\bm v_{\texttt{att}}\in \mathbb R^d$ as a non-negative linear combination of existing token embeddings $\bm v_i$ in the CLIP vocabulary $\mathcal V$ as follows:
\begin{equation}
\bm v_{\texttt{att}} = \sum_{i = 1}^N \alpha_{i} \bm v_i, \;\; \alpha_{i} = g_{\bm \Theta}(\bm v_i) \geq 0,
\label{equ:token}
\end{equation}
where $N$ is the total size of the original CLIP vocabulary, and $\bm v_i$, $i = 1, 2, \dots, N$, are original CLIP token embeddings within $\mathcal V$. Non-negative $\alpha_i$ are parameterized via a multi-layer perceptron (MLP) network $g_{\bm \Theta}(\cdot): \mathbb R^d \mapsto \mathbb R^+$ with ReLU activation. This is inspired by recent work \citep{chefer2024the} on language models. To learn $\bm v_{\texttt{att}}$, we \textbf{optimize the loss $\mathcal L$} in \Cref{equ:loss} with respect to the \textbf{parameter $ \bm \Theta $ of the MLP}, while freezing all the other components. As illustrated in \textbf{\Cref{fig:method}}, during training we enforce the training data pairs $(\bm{z}, \bm c^*) \sim \mathcal{D}$ to satisfy the following constraints: (\emph{i}) $\bm z$ is the latent image containing the target harmful concept. (\emph{ii}) $\bm c^*$ is the text embedding for the text prompt $\bm p$, and $\bm p$ contains the new special token $<\bm v_{\texttt{att}}>$ whose token embedding is $\bm v_{\texttt{att}}$.

\smallskip 
\noindent \textbf{Remarks.} The non-negative constraint in \Cref{equ:token} is inspired by prior works on linear representation hypothesis and linear feature decomposition \citep{chefer2024the,Zhou_2018_ECCV, cunningham2023sparseautoencodershighlyinterpretable, park2023thelinear} that ``negative concepts are not as interpretable as positive concepts.'' In this way, the target concept can be viewed as a combination of top-weighted (i.e., having largest $\alpha_i$) concepts in $\mathcal V$. \textbf{\Cref{fig:alpha}} illustrates the identified sets of human-interpretable concepts for different target concepts (e.g., nudity, Van Gogh, church) in unlearned LDMs. Additionally, we provide analysis on the sparsity of $\alpha_i$ in \textbf{\Cref{app-sec:sparse}}. Now, we introduce how a set of attack token embeddings are learned. 

\subsubsection{Subspace Token Embedding Attacks}

\revise{The single token of \Cref{subsec:single-attack} adapts the interpretable, non-negative parameterization of prior linear-decomposition work \citep{chefer2024the} to the attack setting. Our contribution here is not simply to learn more such tokens: rather than treating the residual concept as one direction, we learn---via orthogonality and deflation---a whole \emph{subspace} that characterizes how the concept persists. It is this subspace, not any individual token, that we later show to be interpretable, transferable, and directly removable by SubDefense.}
Compared with learning a single attack token embedding $\bm v_{\texttt{att}}$, it is more powerful to learn a set of diverse attacks $\{ \bm v_{\texttt{att},k}\}_{k=1}^K \;(m\leq d)$ that can attack the same target concept, as outlined in \textbf{Algorithm 1}. We enforce orthogonality on $\{ \bm v_{\texttt{att},k}\}_{k=1}^K$ to promote diversity and improve attack effectiveness (see ablations in \textbf{\Cref{appsubsec:ablation_attack}}).

\begin{wrapfigure}{r}{0.5\textwidth} 
  \vspace{-10pt}
  \begin{minipage}{0.5\textwidth}
    \hrule
    \vspace{2pt}
    \textbf{Algorithm 1} Learning Attack Token Embeddings
    \vspace{2pt}
    \hrule
    \vspace{2pt}
    \begin{algorithmic}[1]
      \State \textbf{Input:} victim model with CLIP token embeddings [$\bm v_{1,1}, \dots, \bm v_{N,1}$], total iterations $K$
      \State \textbf{Output:} [$\bm v_{\texttt{att},1}, \dots, \bm v_{\texttt{att},K}$]
      \For{$k=1,\dots,K$}
        \State Optimize the MLP $g_{\bm\Theta_k}$
        \State $\alpha_{i,k} \gets g_{\bm\Theta_k}(\bm v_{i,k})$
        \State $\bm v_{\texttt{att},k} \gets \sum_{i=1}^N \alpha_{i,k}\bm v_{i,k}$
        \For{$i=1,\dots,N-1$}
          \State $\bm v_{i,k+1} \gets \bm v_{i,k} - \operatorname{Proj}_{\bm v_{\texttt{att},k}}(\bm v_{i,k})$
        \EndFor
      \EndFor
    \end{algorithmic}
    \vspace{2pt}
    \hrule
  \end{minipage}
  \vspace{-10pt}
\end{wrapfigure}
Such a set of orthogonal token embeddings $\{ \bm v_{\texttt{att},k}\}_{k=1}^K$ is learned through deflation, sharing similar spirits with classical numerical methods such as orthogonal matching pursuit \citep{omp}. Specifically, suppose the first attack token embedding $\bm v_{\texttt{att},1}$ is identified following \Cref{subsec:single-attack} by optimizing an MLP $g_{\bm \Theta_1}$, we then ``eliminate'' the target concept $\bm v_{\texttt{att},1}$ from the whole vocabulary $ \mathcal V$ via orthogonal projection:
\begin{align}
    \bm v_{i,2} \;=\; \bm v_{i,1} - \operatorname{Proj}_{\bm v_{\texttt{att},1}}(\bm v_{i,1}), \quad \forall\; i \in [N].
    \label{equ:project}
\end{align}
Here, $\bm v_{i,1} \equiv \bm v_i \in \mathcal V$ are the original embeddings for all $ i \in [N]$. 
\Cref{equ:project} makes sure \emph{all} the updated $\bm v_{2,i}, \dots, \bm v_{2,N}$ are orthogonal to $\bm v_{\texttt{att},1}$. With the new $\mathcal V_2 = \{\bm v_{2,i} \}_{i=1}^N$, we can learn a second attack token embedding $\bm v_{\texttt{att},2} = \sum_{i = 1}^N \alpha_{i,2} \bm v_{i,2}, \; \alpha_{i,2} = g_{\bm \Theta_2}(\bm v_{i,2}) \geq 0$, then $\bm v_{\texttt{att},2}$ \textbf{is ensured to be orthogonal to} $\bm v_{\texttt{att},1}$. Here, $g_{\bm \Theta_2}$ is another MLP optimized in the same way as $g_{\bm \Theta_1}$. As such, we can repeat the procedure for $K$ times to learn and construct a set of orthogonal attack token embeddings $\{ \bm v_{\texttt{att},k}\}_{k=1}^K$, and use each of them to attack the same target concept. In practice, during attacking, we choose $K=5$, which delivers strong attack performance while keeping the method efficient (see ablation studies in \textbf{\Cref{appsubsec:ablation_attack}}).

\subsection{Subspace Defending: \textit{SubDefense}}\label{subsec:defense}
\label{sec:defend}

SubAttack can find interpretable “hidden words” representing the target concept in unlearned LDMs\revise{, which together span the residual subspace along which the concept persists}. On the one hand, we can leverage SubAttack to examine existing unlearned LDMs and test their unlearning robustness. \revise{On the other hand, the same subspace directly points to a defense: we further remove these identified “hidden words” from unlearned models, so that the model will be more universally robust to various kinds of jailbreaking attacks.} Intuitively, this can be achieved by projecting onto the nullspace of learned subspace attacks. Thanks to the transferability of these attacks, the resulting defense strategy can be robust against a broad range of jailbreaking attacks.

Specifically, suppose we learned a set of attack token embeddings $\{ \bm v_{\texttt{att},k} \}_{k=1}^K$ for a target concept through our subspace attacking method proposed in \Cref{subsec:attack}, let us rewrite $$\bm V_{\texttt{att}} = \begin{bmatrix}
    \bm v_{ \texttt{att},1} & \bm v_{ \texttt{att},2} & \cdots & \bm v_{ \texttt{att},K} 
\end{bmatrix} \in \mathbb{R}^{d\times K}.$$ 
This $\bm V_{\texttt{att}}$ is learned in an unlearned diffusion model whose CLIP token embedding vocabulary is $\mathcal V = \{ \bm v_{  i } \}_{i=1}^N $. The proposed defense will ``block'' the subspace spanned by $\bm V_{\texttt{att}}$ to defend against various jailbreaking attacks. Each token embedding $\bm v_{ i }$ in $\mathcal V$ will be updated as follows:
\begin{align}\label{equ:proj_subspace}
    \bm v_{\texttt{def},i} = \bm v_{ i } - \operatorname{Proj}_{\bm V_{\texttt{att}}}(\bm v_{ i }),\quad \forall\; i \in [N].
\end{align}
For \textit{UnlearnDiff} \cite{zhang2024unlearndiff} and \textit{SubAttack}, their learned jailbreaking attack prompts or embeddings are based on the unlearned LDM's vocabulary. Hence, we will update the unlearned LDM by applying \reviseq{\Cref{equ:proj_subspace}} to complete the defense. After that, UnlearnDiff and SubAttack can still take place, but turn out to have lower ASR (\Cref{sec:defense_result}). For \textit{CCE} \cite{pham2024circumventing}, which learns an attack token embedding $\bm v_{\texttt{att}}$ with no constraints related to the unlearned LDM's vocabulary $\mathcal V$, simply applying \reviseq{\Cref{equ:proj_subspace}} is not enough. Hence, additionally, no matter what $\bm v_{\texttt{att}}$ is learned by CCE, we also apply $\bm v_{\texttt{def}} = \bm v_{\texttt{att}} - \operatorname{Proj}_{\bm V_{\texttt{att}}}(\bm v_{\texttt{att}})$.

\begin{figure}[t]
    \centering
    \includegraphics[width=\linewidth]{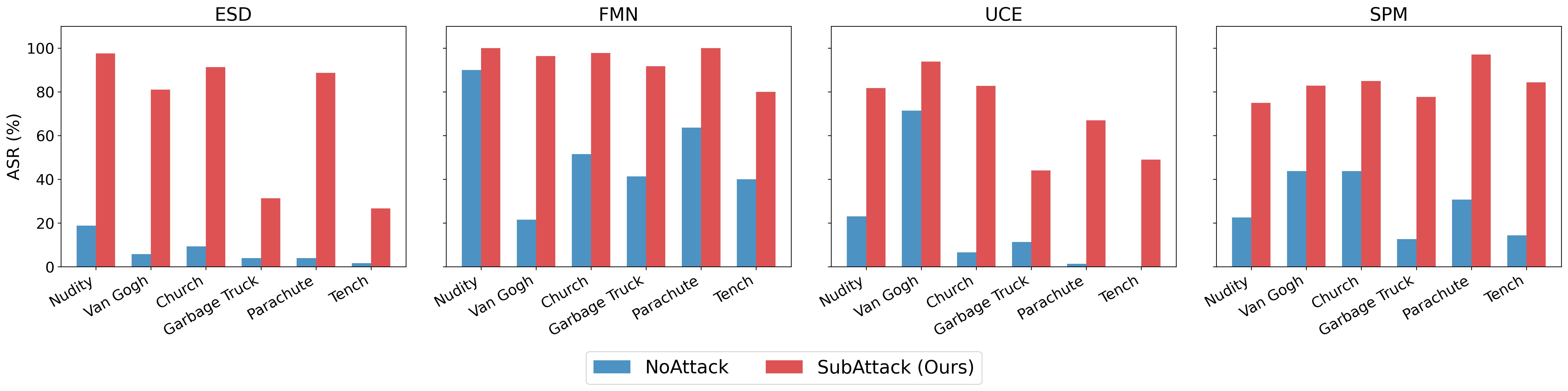}
    \caption{\textbf{SubAttack jailbreaks various concepts (NSFW, style, objects) across different unlearned models (ESD, FMN, UCE, SPM).} It consistently reveals the residual vulnerabilities in these models.}
    \label{fig:subattack}
\end{figure}

\section{Experiments on SubAttack}
\label{sec:exp_attack}

This section first provides a deeper analysis of the interpretable tokens it identifies, and leverages this interpretability to reveal how current unlearned LDMs still conceal target concepts. We then demonstrate through extensive experiments that SubAttack is not only more effective than baseline attacks but also highly transferable.


\subsection{Settings}
%


(\emph{i}) \textbf{Victim Models.} We evaluate SubAttack on a broad set of diffusion-model unlearning methods commonly used in prior jailbreak studies, including ESD \citep{esd}, FMN \citep{zhang2023fmn}, and UCE \citep{gandikota2024uce}, as well as more recent or complementary settings such as SPM \citep{lyu2023spm}, MACE \citep{lu2024macemassconcepterasure}, SA \citep{heng2023selective}, AC \citep{ac}, SalUn \citep{fan2023salun}, and EraseDiff \citep{wu2024erasingundesirableinfluencediffusion}. Following prior work \citep{zhang2024unlearndiff}, all unlearned models are fine-tuned from Stable Diffusion v1.4 \citep{stablediffusion}. 

\smallskip 
\noindent (\emph{ii}) \textbf{Concepts and Dataset.} We perform jailbreaking attacks on representative concept categories in prior diffusion unlearning: ``nudity'' for NSFW, ``Van Gogh'' for style, and objects such as ``church'', ``garbage truck'', ``parachute'', ``tench'', ``airplane'', etc. Following UnlearnDiff \citep{zhang2024unlearndiff}, we construct 300–900 (prompt, seed) pairs per concept, with at least 10 seeds per prompt to reduce randomness and evaluate transferability. Our dataset is $\approx 6\times$ larger than UnlearnDiff’s, enabling a more reliable assessment. 

\smallskip 
\noindent (\emph{iii}) \textbf{Attack and Evaluation.} For each concept, SubAttack learns $K=5$ token embeddings ${\mathbf v_{\texttt{att},k}}$, and an attack is successful if any embedding regenerates the target concept. Further ablations on $K$, orthogonality, and size of vocabulary are in \textbf{\Cref{appsubsec:ablation_attack}}, with sparsity analysis in \textbf{\Cref{app-sec:sparse}}. We report attack success rate (ASR) using pretrained classifiers following \citep{zhang2024unlearndiff}: NudeNet \citep{nudenetclassifier} for NSFW, a WikiArt-finetuned model for style, and an ImageNet-pretrained ResNet-50 for objects. 

\smallskip 
\noindent (\emph{iv}) \textbf{Baselines.} We compare SubAttack against three baselines: NoAttack (original prompts without jailbreak), UnlearnDiff \citep{zhang2024unlearndiff}, and CCE \citep{ac}. UnlearnDiff and CCE are reproduced with their original settings but unified under our dataset (e.g., UnlearnDiff optimizes an adversarial prompt per (prompt, seed) pair). We provide more experiment details in \textbf{\Cref{exp_setting_attack}}.

\subsection{Interpretability of Proposed SubAttack Methods}
\label{subsec:interpret}

We analyze the embeddings \( \{ \bm v_{\texttt{att},k} \}_{k=1}^K \) to examine how target concepts persist in unlearned LDMs. For each \( \bm v_{\texttt{att},k} \), we extract the top-50 highest-weighted tokens, stem and lemmatize them, and visualize the most frequent ones with WordCloud. The same procedure is applied to the original SD for comparison. We present key examples and findings below, with more results in \textbf{\Cref{app-subsec:interpret}}.

\begin{figure}[t]
\centering
\begin{subfigure}[b]{0.19\textwidth}
         \centering
        \includegraphics[width=\linewidth]{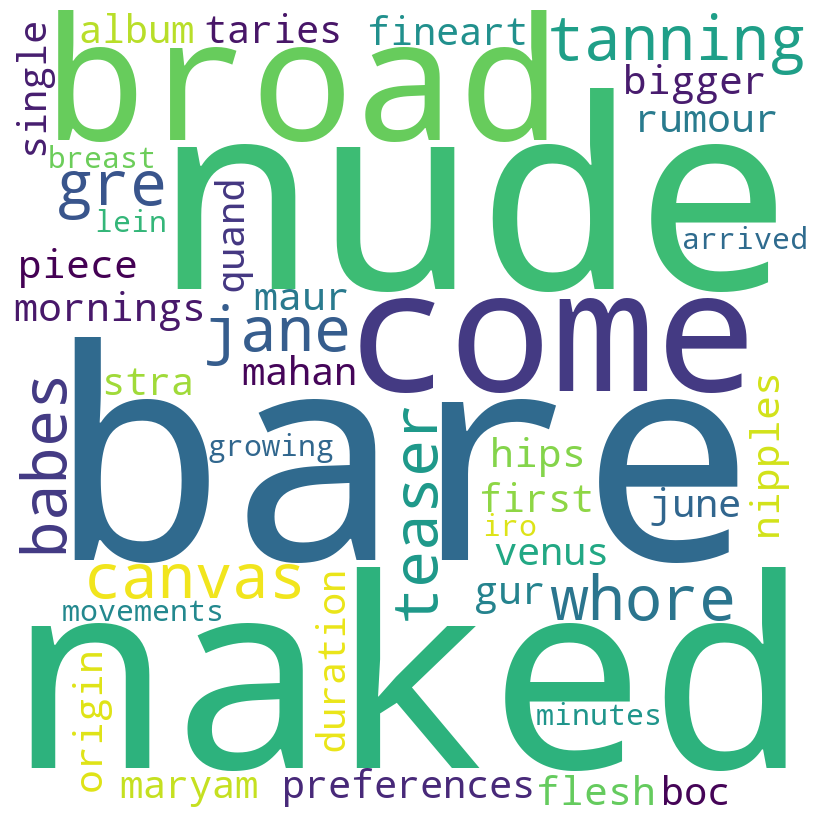}
        \caption{SD}
        \label{fig:sd}
     \end{subfigure}
\begin{subfigure}[b]{0.19\textwidth}
         \centering
        \includegraphics[width=\linewidth]{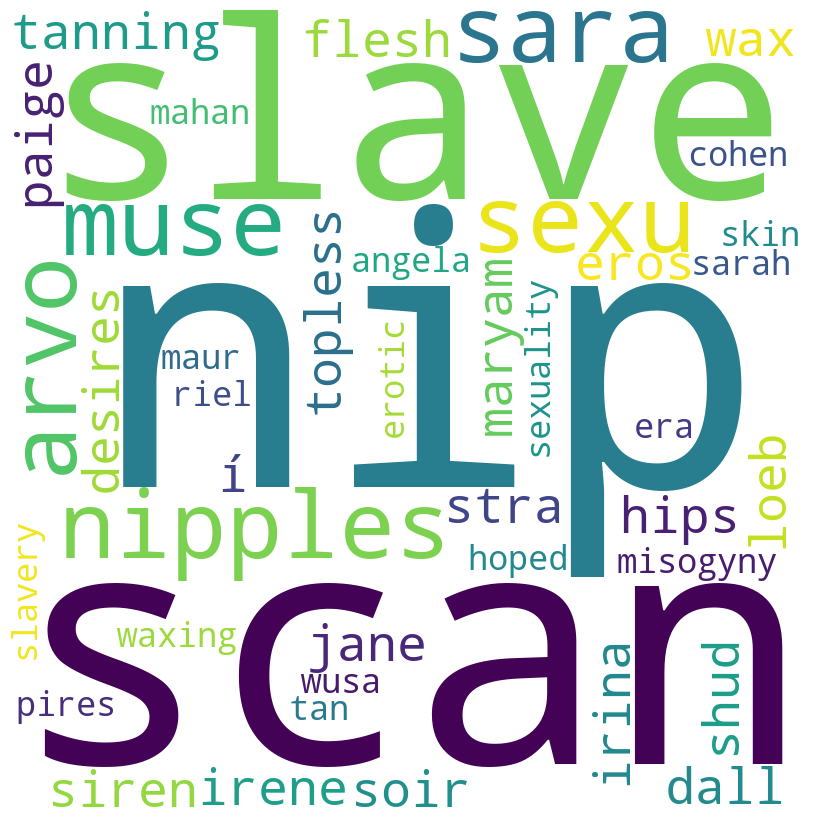}
        \caption{ESD}
        \label{fig:esd}
     \end{subfigure}
\begin{subfigure}[b]{0.19\textwidth}
         \centering
        \includegraphics[width=\linewidth]{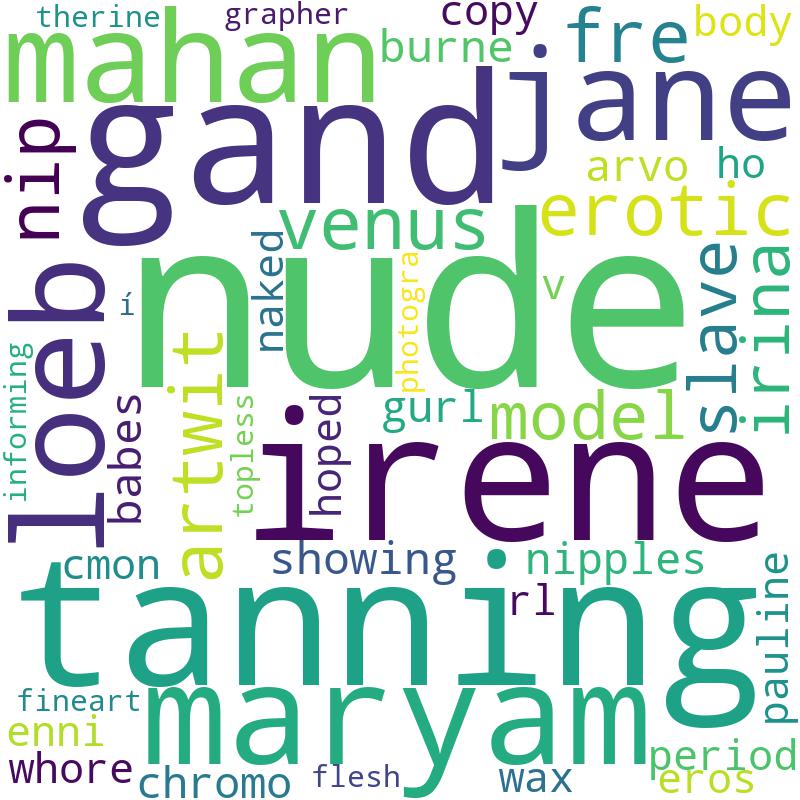}
        \caption{FMN}
        \label{fig:fmn}
     \end{subfigure}
\begin{subfigure}[b]{0.19\textwidth}
         \centering
        \includegraphics[width=\linewidth]{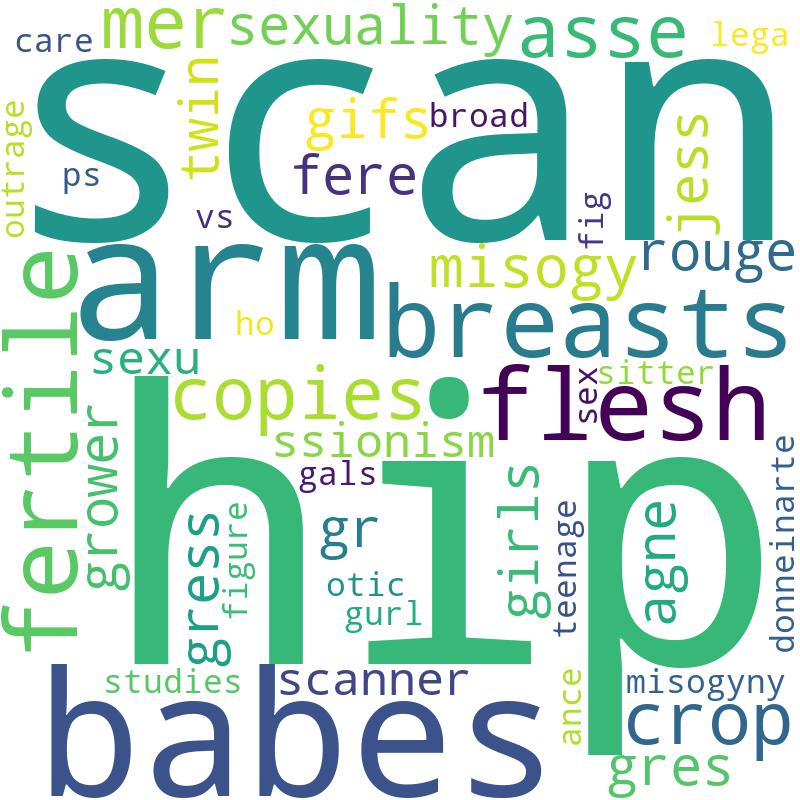}
        \caption{UCE}
        \label{fig:uce}
     \end{subfigure}
\begin{subfigure}[b]{0.19\textwidth}
         \centering
        \includegraphics[width=\linewidth]{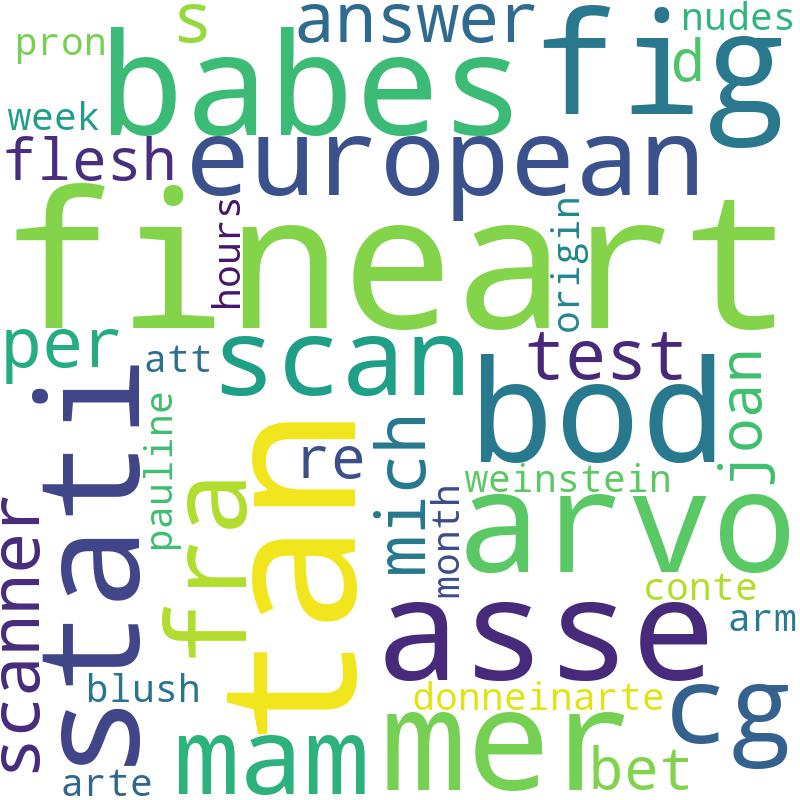}
        \caption{SPM}
        \label{fig:spm}
     \end{subfigure}
\caption{\textbf{Interpreting the subspace of attack token embeddings for concept ``nudity'' across different models.} (a) The original LDM (i.e., SD) majorly relates it to \textbf{explicit} synonyms. (b-e) Unlearned LDMs more heavily associate it with \textbf{implicit} concepts.}
\label{fig:esd_sd}
\end{figure}

(\textbf{\textit{i}}) \textbf{SubAttack enables learned embeddings understandable to humans.} The resulting tokens reveal meaningful and positively associated concepts rather than random noise. We observe sexualized terms for the NSFW concept (e.g., ``slave'', ``babes'') in \textbf{\Cref{fig:esd_sd}}, cross-lingual variants for church (e.g., ``kirk'' in Scottish English) in \textbf{\Cref{fig:app_interpret_church}}, and key painting elements for Van Gogh (e.g., ``oats'', ``night'') in \textbf{\Cref{fig:app_interpret_vangogh}}. These findings verify that the learned embeddings are directly interpretable, providing a clear semantic view of what remains in unlearned models.

\begin{wrapfigure}{r}{.25\textwidth}
    \centering
    \includegraphics[width=.25\textwidth]{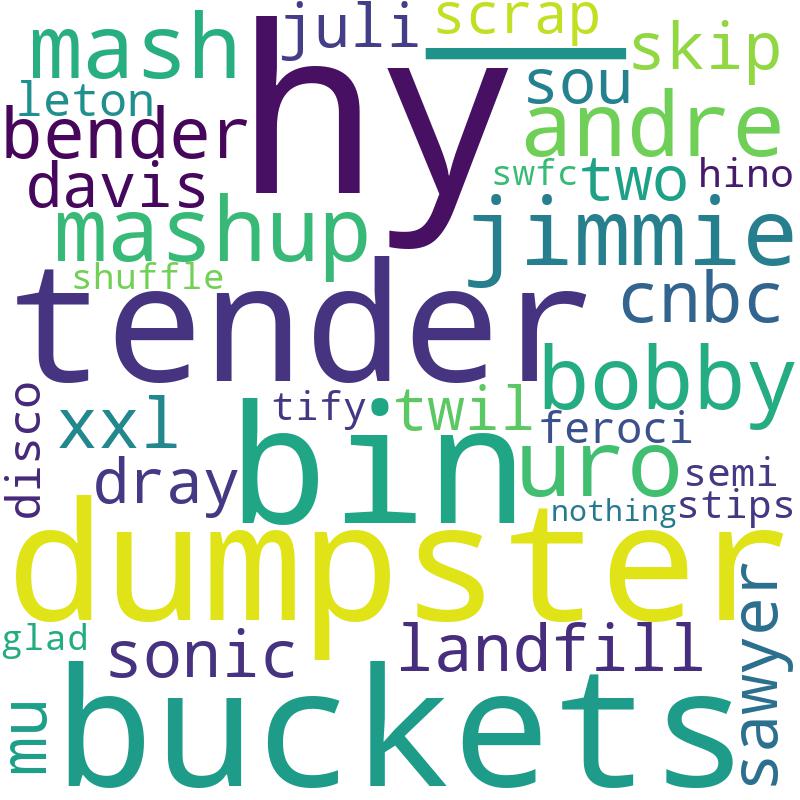}
    \caption{\textbf{ESD for ``garbage truck''}.}
    \label{fig:vis_esd_garbage}
\end{wrapfigure}
\smallskip 
\noindent (\textbf{\textit{ii}}) \textbf{SubAttack shows how stronger unlearning mutes keywords yet leaves hidden clues.} Based on the human-understandable embeddings, we can directly compare original SD and unlearned LDMs to observe a clear progression in how concepts persist. As shown in \textbf{\Cref{fig:esd_sd}}, SD relies on obvious keywords (e.g., “nude,” “naked”), weakly unlearned models retain both obvious and hidden terms (e.g., “tanning,” women’s names), and strongly unlearned models suppress the obvious ones but still retain hidden associations (e.g., “slave,” “nip,” “babes”). A similar effect appears in other concepts as well in \textbf{\Cref{app-subsec:interpret}}. Notably, even a strong unlearned “garbage truck” model with only 4\% NoAttack ASR still surfaces terms like “dumpster,” “bin,” and “landfill” (\textbf{\Cref{fig:vis_esd_garbage}}). These findings show that unlearning reduces surface-level cues but does not eliminate deeper associations, providing insights unavailable from non-interpretable attacks.



\begin{wraptable}{r}{.4\textwidth}
\centering
\caption{\textbf{CLIP similarity} between residual and original explicit concept across unlearned LDMs.}
\resizebox{.4\textwidth}{!}{%
\begin{tabular}{@{}l|cccc@{}}
\toprule
\midrule
\textbf{Concept} & \textbf{ESD} & \textbf{FMN} & \textbf{UCE} & \textbf{SPM} \\
\midrule
Van Gogh & 0.61 & 0.61 & 0.74 & 0.67 \\
Church   & 0.76 & 0.85 & 0.79 & 0.82 \\
\midrule
\bottomrule
\end{tabular}
}
\label{tab:clip_similarity}
\end{wraptable}

\smallskip 
\noindent (\textbf{\textit{iii}}) \textbf{SubAttack measures how closely the remaining concept matches the original concept.} Beyond visualization, SubAttack provides a quantitative way to assess similarity. Using CLIP similarity between attack tokens and the target concept (\textbf{\Cref{tab:clip_similarity}}), we find that weaker unlearning models (e.g., UCE for “Van Gogh,” FMN/SPM for “church”) retain tokens more semantically aligned with the original concept and also exhibit higher ASR under NoAttack (\textbf{\Cref{fig:subattack}}). These results suggest that SubAttack can be used to quantify how much of a concept explicitly remains in unlearned models.


\begin{figure}[t]
    \centering
    \includegraphics[width=.75\textwidth]{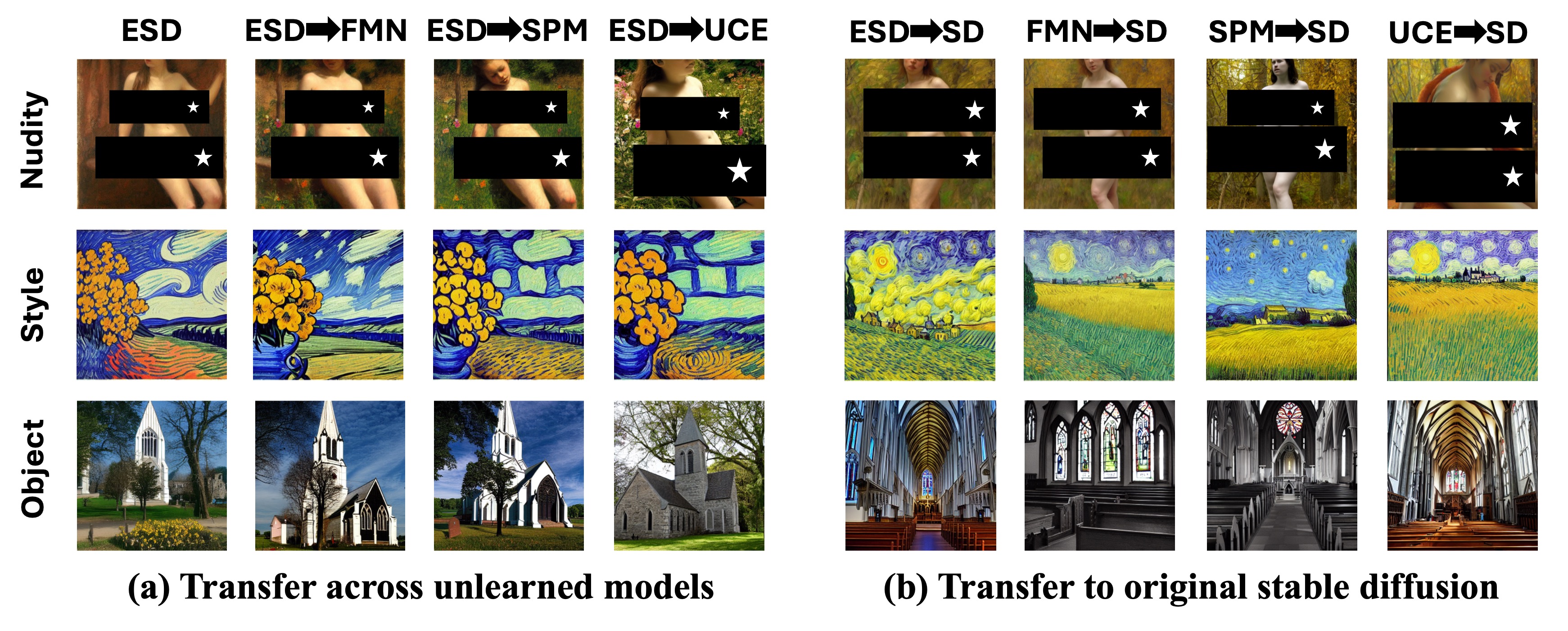}
    \caption{\textbf{Transfer attack} token embeddings learned by SubAttack to different unlearned models or to the original diffusion model.}
    \label{fig:transfer_model_combine}
\end{figure}
(\textbf{\textit{iv}}) \textbf{SubAttack shows the \revise{remaining} concept is inherited from the original SD.} SubAttack embeddings remain effective when transferred back into the original SD. Transfer ASR is consistently above 80\% across all concepts and models (See \textbf{\Cref{app-subsec:interpret}} \textbf{\Cref{tab:to_sd}}; visualized in \textbf{\Cref{fig:transfer_model_combine} (b)}), suggesting that residual associations in unlearned models are inherited from SD rather than independently formed. These inherited associations are likely a key reason unlearned models continue to generate harmful content.

\subsection{Effectiveness of Proposed SubAttack Methods}
\label{subsec:global}

(\textbf{\textit{i}}) \textbf{SubAttack is an efficient global attack.}
UnlearnDiff is a local attack that optimizes an adversarial prompt for each (prompt, seed) pair, which is time-consuming. In contrast, SubAttack learns global attack token embeddings that generalize across prompts and seeds. As shown in \textbf{\Cref{fig:subattack}}, SubAttack’s global embeddings can jailbreak diverse concepts across hundreds of prompts and seeds. \rebuttal{The runtime in \Cref{fig:attack_compare} amortizes this one-time concept-level optimization over 900 prompt--seed pairs; a stage-wise learning/application breakdown is provided in \Cref{appsubsec:efficiency_breakdown}.}

\begin{figure}[t]
    \centering
    \includegraphics[width=.95\linewidth]{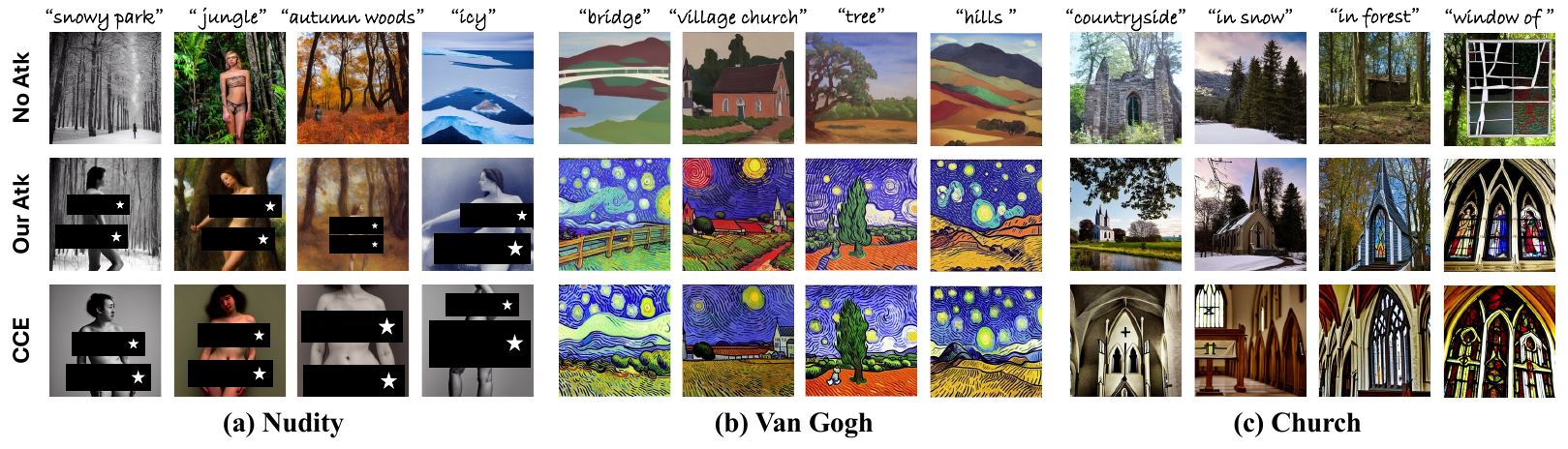}
    \caption{
    \textbf{SubAttack can generate the target concepts with high ASR while aligning with original text prompts}. For example, our attack generates nude women with different backgrounds while CCE fails to generate the correct backgrounds.}
    \label{fig:combine_esd}
\end{figure}

\smallskip 
\noindent 
(\textbf{\textit{ii}}) \textbf{SubAttack is highly effective.} As shown in \textbf{\Cref{fig:attack_compare}}, SubAttack exhibits strong attack success rates (ASR). Notably, even as a local attack, UnlearnDiff frequently underperforms SubAttack; for instance, on the ``church'' concept across multiple unlearned models. While CCE learns unconstrained attack embeddings with commendable performance, it lacks interpretability. In contrast, SubAttack enforces explicit linear structures, which not only enhance performance but also intrinsically enable interpretability. Furthermore, \textbf{\Cref{fig:combine_esd}} illustrates SubAttack's superior fidelity to text prompts. It can faithfully integrate a nude figure into diverse backgrounds such as snowy parks, jungles, and woods, demonstrating precise compositional control. Additional visualizations are provided in \textbf{\Cref{appsec:attack_vis}}.

 \smallskip 
\noindent 
(\textbf{\textit{iii}}) \textbf{SubAttack is transferable across different unlearned LDMs.} The attack token embeddings learned by SubAttack transfer robustly between unlearned LDMs. As shown in \textbf{\Cref{fig:transfer_model_combine} (a)}, embeddings learned via SubAttack on the ESD model are directly transferred to attack FMN, SPM, and UCE. All three concept types, nudity, style, and object, can be successfully transferred to these target models with high ASR. We further compare the transfer ASR of SubAttack against other baselines in \textbf{\Cref{fig:transfer_attack_compare}} (more results in \textbf{\Cref{tab:more_transfer}} in \textbf{\Cref{appsubsec:more_asr}}), where we transfer the token embeddings from
\begin{table}[t]
\centering
\caption{\textbf{Transfer attack performance of various jailbreaking methods} from ESD to other models across different concepts, measured by ASR (\%).}
\resizebox{.85\textwidth}{!}{%
\begin{tabular}{@{}l|ccc|ccc|ccc@{}}
\toprule
\midrule
\textbf{Concepts:} & \multicolumn{3}{c|}{\textbf{Nudity}} & \multicolumn{3}{c|}{\textbf{Van Gogh}} & \multicolumn{3}{c}{\textbf{Church}} \\
\midrule
\textbf{Victim Models:} & FMN & UCE & SPM & FMN & UCE & SPM & FMN & UCE & SPM \\
\midrule
NoAttack         & 90.00 & 23.00 & 22.56 & 21.56 & 71.44 & 43.78 & 51.56 & 6.55  & 43.78 \\
UnlearnDiff       & 93.33 & 41.33 & 38.22 & 12.78 & 64.00 & 47.11 & 6.19  & 13.33 & 58.00 \\
CCE               & 93.00 & 18.33 & 37.56 & 72.33 & 43.56 & 81.33 & 91.00 & 70.11 & \textbf{92.78} \\
\rowcolor{gray!10}
SubAttack (Ours)  & \textbf{96.89} & \textbf{77.00} & \textbf{80.44} & \textbf{72.67} & \textbf{88.89} & \textbf{86.89} & \textbf{92.89} & \textbf{83.77} & 92.00 \\
\midrule
\bottomrule
\end{tabular}
}
\label{fig:transfer_attack_compare}
\end{table}
CCE and the adversarial prompts from UnlearnDiff to other victim models accordingly. SubAttack consistently achieves the highest transfer ASR across different models and concepts. This strong transferability matches the finding that SubAttack identifies embeddings inherited from the original SD model (\textbf{\Cref{subsec:interpret}}).

\begin{table}[t]
\centering
\caption{\textbf{Attack performance of various jailbreaking methods}, measured by ASR (\%) over 900 prompt--seed pairs per concept. \rebuttal{Runtime includes attack construction and generation; concept-level construction for CCE and SubAttack is amortized over all 900 pairs, whereas UnlearnDiff re-optimizes each pair.} Best results are highlighted in \textbf{bold}.}
\resizebox{\textwidth}{!}{%
\begin{tabular}{@{}l|cccc|cccc|cccc|c|c|c@{}}
\toprule
\midrule
\multicolumn{13}{c|}{\textbf{ASR (\%) \ensuremath{\uparrow}}} & \multirow{3}{*}[-6pt]{\textbf{\shortstack{Time per\\ Data \\ (s) \ensuremath{\downarrow}}}} & \multirow{3}{*}[-1pt]{\textbf{\shortstack{Interp-\\ retable}}} & \multirow{3}{*}[-1pt]{\textbf{\shortstack{Inspire\\ Defense}}} \\
\cmidrule(rr){1-13}
\textbf{Concepts:}  & \multicolumn{4}{|c|}{\textbf{Nudity}} & \multicolumn{4}{|c|}{\textbf{Van Gogh}} & \multicolumn{4}{|c|}{\textbf{Church}} & & & \\
\cmidrule(rr){1-13}
\textbf{Victim Models:} & ESD & FMN & UCE & SPM & ESD & FMN & UCE & SPM & ESD & FMN & UCE & SPM & & & \\
\midrule
NoAttack       & 18.78 & 90.00 & 23.00 & 22.56 & 5.78  & 21.56 & 71.44 & 43.78 & 9.33  & 51.56 & 6.55  & 43.78 & NA & NA & NA \\
UnlearnDiff     & 51.11 & \textbf{100.00} & 78.22 & \textbf{83.33} & 40.94 & \textbf{100.00} & \textbf{100.00} & 53.49 & 51.74 & 35.33 & 61.67 & 53.67 & 906.6 & \textcolor{red}{\textbf{\ding{55}}} & \textcolor{red}{\textbf{\ding{55}}} \\
CCE             & 85.11 & 98.33 & 77.22 & 78.33 & 75.22 & 93.33 & 95.67 & 81.67 & 82.00 & \textbf{97.78} & 81.89 & 76.67 & \textbf{11.4} & \textcolor{red}{\textbf{\ding{55}}} & \textcolor{red}{\textbf{\ding{55}}} \\
\rowcolor{gray!10}
SubAttack (Ours)& \textbf{97.56} & \textbf{100.00} & \textbf{81.67} & 74.89 & \textbf{81.00} & 96.33 & 98.33 & \textbf{82.78} & \textbf{91.33} & \textbf{97.78} & \textbf{82.67} & \textbf{84.89} & 54.2 & \textcolor{green!60!black}{\textbf{\ding{51}}} & \textcolor{green!60!black}{\textbf{\ding{51}}} \\
\midrule
\bottomrule
\end{tabular}
}
\label{fig:attack_compare}
\end{table}

\section{Experiments on SubDefense}
\label{sec:defense_result}

Having established the effectiveness of SubAttack, we next demonstrate the SubDefense method inspired by our attack. We integrate SubDefense into existing unlearned models and assess its robustness. Comprehensive results show that SubDefense offers a more versatile and robust defense than baseline methods, while better preserving generation quality on safe prompts.


\subsection{Settings}\label{subsec:def-setup}

\begin{figure}[t]
    \centering
    \includegraphics[width=.8\textwidth]{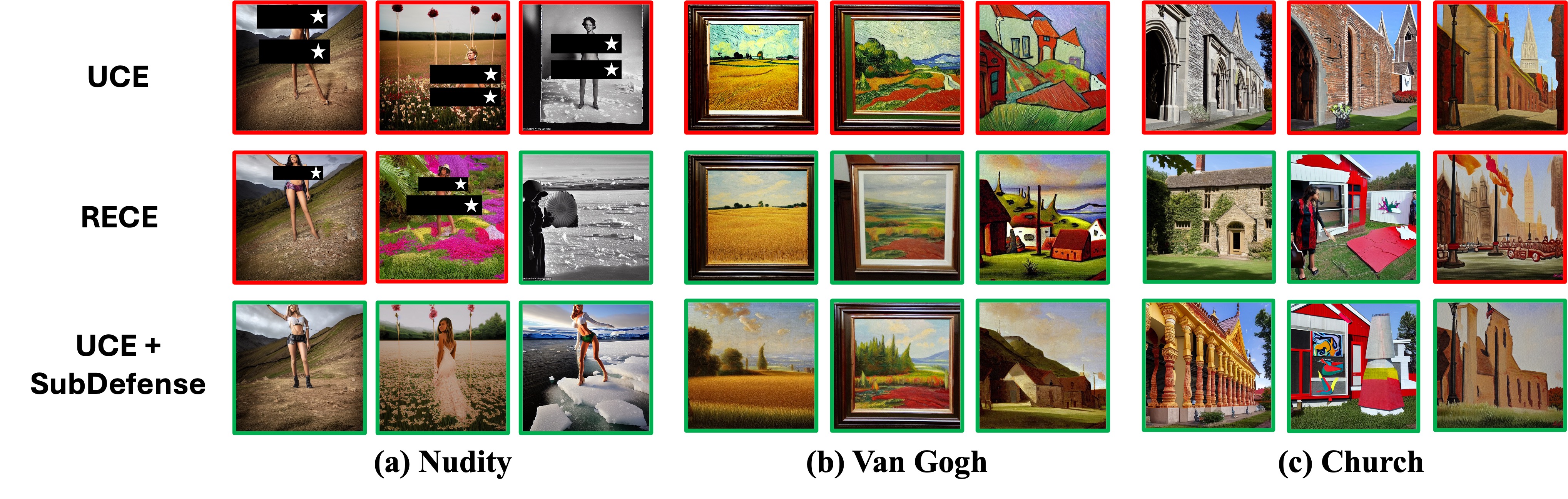}
    \caption{\textbf{Defending UCE} using RECE or SubDefense across various concepts.}
    \label{fig:defense_uce_unlearndiff}
\end{figure}
(\textbf{\emph{i}}) \textbf{Basics.} SubDefense is plugged into UCE, ESD, FMN, and SPM for concepts ``nudity'', ``Van Gogh'', and ``church'' using our constructed dataset by default. \rebuttal{We use rank 20 only for the controlled comparison with RECE and fix rank 100 for the broader evaluation; rank is never selected per test prompt. Appendix \Cref{fig:asr_k,tab:clip_fid,tab:cce_defense_app} reports the corresponding sensitivity and robustness--utility trade-off. Utility changes gradually through rank 100 and degrades more sharply beyond it.}
(\textbf{\emph{ii}}) \textbf{Metrics.} To assess defense effectiveness, new jailbreaking attacks are conducted after applying defenses, and the corresponding ASR is reported. SubAttack with $K=5$ is used consistently before and after defense to 
ensure a fair comparison. Additionally, the generative quality of the defended unlearned models is evaluated on the MSCOCO-10k dataset \citep{coco, zhang2024defensive} using FID and CLIP scores \citep{CLIPScore}. Further details are in \textbf{\Cref{exp_setting_defense}}.

\subsection{Performance of SubDefense}\label{subsec:def-performance}

(\textbf{\textit{i}}) \textbf{SubDefense demonstrates a stronger defense.}
We compare SubDefense with RECE \citep{gong2024reliableefficientconcepterasure}, which is proposed to defend UCE against adversarial attacks. As shown in \textbf{\Cref{tab:defense_uce_combine}}, SubDefense 
\begin{figure}[t]
    \centering
    \includegraphics[width=.8\textwidth]{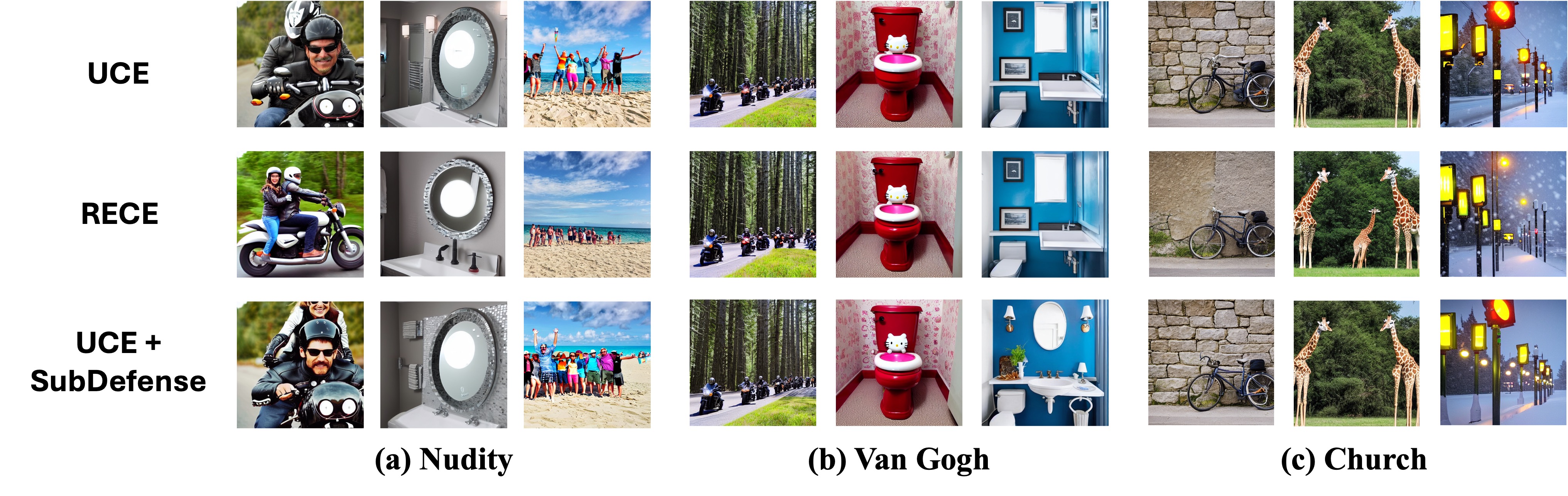}
    \caption{\textbf{Safe image generation} after applying RECE or SubDefense.}
    \label{fig:coco_compare}
\end{figure}
achieves lower ASR, while also attaining lower FID and higher CLIP scores on COCO-10k across three categories of concepts, indicating stronger robustness and better preservation of safe generation quality (\textbf{\Cref{fig:defense_uce_unlearndiff}}, \textbf{\Cref{fig:coco_compare}}). More visualizations are provided in \textbf{\Cref{appsec:uce_vis}}. In particular, for the ‘Van Gogh’ concept, which is closely tied to ‘blue’ and ‘star,’ SubDefense preserves these benign elements, demonstrating that it goes beyond naive blocking of all related tokens.

\begin{table}[t]
\centering
\caption{\textbf{SubDefense is more robust than baseline RECE in defending three concepts on UCE against UnlearnDiff or our SubAttack, while preserving better generative quality.}}
\resizebox{.95\textwidth}{!}{%
\begin{tabular}{@{}l|cc|cc|cc|cc@{}}
\toprule
\midrule
 \textbf{Metrics:} 
 &  \multicolumn{2}{|c|}{\textbf{UnlearnDiff ASR \ensuremath{\downarrow}}}
 &  \multicolumn{2}{|c|}{\textbf{SubAttack ASR \ensuremath{\downarrow}}} 
 &  \multicolumn{2}{|c|}{\textbf{COCO-10k FID \ensuremath{\downarrow}}} 
 &  \multicolumn{2}{|c}{\textbf{COCO-10k CLIP \ensuremath{\uparrow}}} 
 \\ \midrule

\textbf{Scenarios:}  
& SubDefense   & RECE  
& SubDefense   & RECE
& SubDefense   & RECE
& SubDefense   & RECE  
\\ \midrule  

Nudity  
&  \textbf{73.55\%}  &  76.44\%       
&  \textbf{34.11\%}  &  62.44\%     
&  \textbf{17.51}  &  17.57         
&  \textbf{30.70}  &  30.07    
\\

Van Gogh   
&  \textbf{52.78\%} &   61.67\%   
&  \textbf{29.44\%} &  84.44\%   
&  \textbf{16.64}  &  17.11    
&  \textbf{30.94}  & 30.08    
\\

Church 
& \textbf{39.78\%} &  50.78\%  
&  \textbf{5.22\%} &  80.33\%  
&  \textbf{17.41}  &  \textbf{17.41}   
& \textbf{30.86}  &  30.07   
\\
\midrule
\bottomrule
\end{tabular}
}
\label{tab:defense_uce_combine}
\end{table}

\smallskip 
\noindent 
(\textbf{\textit{ii}}) \textbf{SubDefense is robust across attacks, models, and concepts.}
\rebuttal{The SubAttack and CCE columns of \Cref{tab:defense} re-optimize each attack from scratch after applying SubDefense, rather than replaying pre-defense tokens. Their nonzero post-defense ASR shows that exploitable information remains. We do not attribute this residual exclusively to nonlinear structure: unrecovered linear directions, attack optimization, and detector error may also contribute. Appendix \Cref{tab:clip_fid,tab:cce_defense_app} quantifies how increasing projection rank reduces ASR while progressively harming utility.}
On ESD “nudity,” SubDefense lowers ASR against UnlearnDiff, SubAttack, and CCE (\textbf{\Cref{tab:defense}}), showing its ability to defend against
\begin{table}[t]
\centering
\caption{\textbf{SubDefense can defend ESD against different kinds of attacks.}}
\resizebox{.85\textwidth}{!}{%
\begin{tabular}{@{}c|cccc|c|c@{}}
\toprule
\midrule
\multirow{2}{*}{\textbf{Metrics:}} & \multicolumn{4}{c|}{\textbf{Nudity ASR}} & \multirow{2}{*}{\textbf{CLIP}} & \multirow{2}{*}{\textbf{FID}} \\
\cmidrule(lr){2-5}
 & \textbf{NoAttack} & \textbf{UnlearnDiff} & \textbf{CCE} & \textbf{SubAttack} & & \\
\midrule
\textbf{ESD} & 18.11\% & 51.11\% & 85.11\% & 97.56\% & 30.13 & 18.23 \\
\midrule
\textbf{ESD+SubDefense} & 0.0\% & 4.56\% & 75.67\% & 42.33\% & 29.58 & 19.20 \\
\midrule
\bottomrule
\end{tabular}
}
\label{tab:defense}
\end{table}
diverse jailbreak strategies. Extended results confirm its effectiveness across other unlearned models (FMN and beyond) and concepts (I2P and beyond) in \textbf{\Cref{appsubsec:defense_subattack,appsubsec:defense_i2p}}. We also include exploratory results on a classic black-box attack and on the original SD in \textbf{\Cref{appsubsec:defense_other}}. 


\smallskip 
\noindent 
(\textbf{\textit{iii}}) \textbf{SubDefense offers a complementary linear refinement against CCE.}
Although a recent nonlinear unlearning method, STEREO~\cite{srivatsan2024stereo}, shows improved robustness, CCE remains one of the most challenging white-box attacks to defend for most existing unlearned models. Our goal is not to replace such nonlinear pipelines, but to test—through a plug-and-play refinement strategy inherited from our interpretable diagnosis—whether the \emph{linear} residual structure revealed by SubAttack can further reduce CCE success.
Within this linear framework, SubDefense consistently lowers CCE ASR. As shown in \textbf{\Cref{appsubsec:ablation_defense}}, projecting more directions reduces ASR from 85.11\% to 8.89\%. This establishes SubDefense as a simple, interpretable refinement, while clarifying the limits of linear defenses and motivating future exploration of potential nonlinear residual structures (\textbf{\Cref{appsec_discussion}}). 

\smallskip 
\noindent 
\revise{(\textbf{\textit{iv}}) \textbf{SubDefense is a plug-and-play add-on to any unlearned model, including adversarially finetuned ones.}
Because the residual subspace is identified purely from probing a given model, SubDefense can be applied on top of \emph{any} already-unlearned model without re-training it, regardless of how the base model was obtained. \textbf{\Cref{tab:addon_defense}} reports nudity ASR under SubAttack before and after this add-on. On standard unlearners (ESD, FMN, UCE, SPM) it consistently removes a large fraction of the residual vulnerability (full per-concept results in \textbf{\Cref{appsubsec:defense_subattack}}). Crucially, it also composes with STEREO~\cite{srivatsan2024stereo}, a strong adversarially finetuned unlearner: even though STEREO already suppresses much of the concept (ASR $21.67\%$), SubDefense further lowers ASR to $6.67\%$. This shows SubDefense acts as a complementary refinement rather than a competitor to adversarial finetuning: the interpretable linear residual it removes is orthogonal to what the base unlearning procedure already handles.}

\begin{table}[h!]
\centering
\revise{\caption{\textbf{SubDefense composes on top of any unlearned model.} Nudity ASR under SubAttack, before and after applying SubDefense as a plug-and-play refinement, across standard unlearners and the adversarially finetuned unlearner STEREO. ``\#\,blocked'' is the number of projected directions.}}
\resizebox{.72\textwidth}{!}{%
\revise{\begin{tabular}{@{}l|cc|c@{}}
\toprule
\midrule
\textbf{Base unlearned model} & \textbf{Base ASR} & \textbf{+\,SubDefense} & \textbf{\#\,blocked} \\
\midrule
\multicolumn{4}{@{}l}{\emph{Standard unlearners}} \\
ESD & 97.56\% & 42.33\% (-55.23\%) & 100 \\
FMN & 100\%   & 62.89\% (-37.11\%) & 100 \\
UCE & 81.67\% & 28\% (-53.67\%)    & 100 \\
SPM & 74.89\% & 50.78\% (-24.11\%) & 100 \\
\midrule
\multicolumn{4}{@{}l}{\emph{Adversarially finetuned unlearner}} \\
\rowcolor{gray!10}
STEREO & 21.67\% & \textbf{6.67\%} (-15\%) & 20 \\
\midrule
\bottomrule
\end{tabular}}
}
\label{tab:addon_defense}
\end{table}

\section{Conclusion}

\rebuttal{This paper shows that a coherent, human-interpretable, attack-accessible linear residual of a supposedly erased concept can be recovered in the token embedding space, and that both an attack and a defense follow from this structure.} SubAttack \revise{reads out this subspace, learning} token embeddings capable of regenerating harmful concepts in unlearned diffusion models; \revise{it is interpretable, revealing the textual associations through which the concept is retained,} and shows strong transferability across prompts, noise inputs, and models, exposing deeper vulnerabilities in current unlearning techniques. \revise{Conversely, projecting out the same subspace yields} SubDefense, a plug-and-play mechanism that disrupts residual subspaces to defend against diverse attacks while preserving generation quality. Together, our findings highlight the urgent need for more robust unlearning methods and provide actionable directions for strengthening the safety of diffusion models.

\rebuttal{\noindent \textbf{Limitations.} Our approach assumes white-box access to the diffusion pipeline and a small set of target-concept images, and our evaluation primarily studies Stable Diffusion v1.4 with a CLIP text encoder. SubDefense removes recovered attack-accessible linear directions at the conditioning interface rather than guaranteeing complete concept erasure; adaptive attacks can still exploit residual information, and increasing the projection rank trades robustness for broad and local semantic utility. Moreover, ASR relies on imperfect concept classifiers. We therefore view SubDefense as a complementary post-hoc refinement, not a replacement for robust unlearning.}

\section*{Acknowledgment}

SYC and QQ acknowledge support from NSF CAREER CCF-2143904, NSF CCF 2212326, NSF IIS 2402950, NSF IIS 2312842, ARO W911NF25-1-0047, and ONR N000142512339. The work of YMZ and SJL is also partially supported by the NSF IIS 2207052 and the Amazon Research Award for AI in Information Security.

\bibliographystyle{unsrtnat}
\bibliography{diffusion}

@inproceedings{chin2024prompting4debugging,
author = {Chin, Zhi-Yi and Jiang, Chieh-Ming and Huang, Ching-Chun and Chen, Pin-Yu and Chiu, Wei-Cheng},
title = {Prompting4Debugging: red-teaming text-to-image diffusion models by finding problematic prompts},
year = {2024},
booktitle = {Proceedings of the 41st International Conference on Machine Learning},
articleno = {336},
numpages = {19},
location = {Vienna, Austria},
}

@inproceedings{
tsai2024ringabell,
title={Ring-A-Bell! How Reliable are Concept Removal Methods For Diffusion Models?},
author={Yu-Lin Tsai and Chia-Yi Hsu and Chulin Xie and Chih-Hsun Lin and Jia You Chen and Bo Li and Pin-Yu Chen and Chia-Mu Yu and Chun-Ying Huang},
booktitle={The Twelfth International Conference on Learning Representations},
year={2024},
url={https://openreview.net/forum?id=lm7MRcsFiS}
}

@article{bereska_mechanistic_2024,
  title   = {Mechanistic Interpretability for AI Safety - A Review},
  author  = {Bereska, Leonard and Gavves, Efstratios},
  year    = {2024},
  month   = {Aug},
  journal = {Transactions on Machine Learning Research},
  url     = {https://openreview.net/forum?id=ePUVetPKu6}
}

@article{Zou2023RepresentationEngineering,
  title   = {Representation Engineering: A Top-Down Approach to AI Transparency},
  author  = {Zou, Andy and Phan, Long and Chen, Sarah and Campbell, James and Guo, Phillip and Ren, Richard and Pan, Alexander and Yin, Xuwang and Mazeika, Mantas and Dombrowski, Ann-Kathrin and Goel, Shashwat and Li, Nathaniel and Byun, Michael J. and Wang, Zifan and Mallen, Alex and Basart, Steven and Koyejo, Sanmi and Song, Dawn and Fredrikson, Matt and Kolter, J. Zico and Hendrycks, Dan},
  journal = {arXiv preprint arXiv:2310.01405},
  year    = {2023},
  url     = {https://arxiv.org/abs/2310.01405},
  eprint  = {2310.01405},
  archivePrefix = {arXiv},
  primaryClass  = {cs.LG}
}

@InProceedings{zhuang2023a,
    author    = {Zhuang, Haomin and Zhang, Yihua and Liu, Sijia},
    title     = {A Pilot Study of Query-Free Adversarial Attack Against Stable Diffusion},
    booktitle = {Proceedings of the IEEE/CVF Conference on Computer Vision and Pattern Recognition (CVPR) Workshops},
    month     = {June},
    year      = {2023},
    pages     = {2385-2392}
}

@article{protein,
    title = {De novo design of protein structure and function with RFdiffusion},
    author = {Watson, Joseph L. and Juergens, David and Bennett, Nathaniel R. and Trippe, Brian L. and Yim, Jason and Eisenach, Helen E. and Ahern, Woody and Borst, Andrew J. and Ragotte, Ryan J. and Milles, Laura F. and others},
    journal = {Nature},
    volume = {618},
    number = {7962},
    pages = {512-518},
    year = {2023},
    doi = {10.1038/s41586-023-06415-8},
    pmid = {37433327}
}

@inproceedings{esd,
  title={Erasing Concepts from Diffusion Models},
  author={Rohit Gandikota and Joanna Materzy\'nska and Jaden Fiotto-Kaufman and David Bau},
  booktitle={Proceedings of the 2023 IEEE International Conference on Computer Vision},
  year={2023}
}

@article{fan2023salun,
  title={Salun: Empowering machine unlearning via gradient-based weight saliency in both image classification and generation},
  author={Fan, Chongyu and Liu, Jiancheng and Zhang, Yihua and Wei, Dennis and Wong, Eric and Liu, Sijia},
  journal={arXiv preprint arXiv:2310.12508},
  year={2023}
}

@inproceedings{
    zhang2024defensive,
    title={Defensive Unlearning with Adversarial Training for Robust Concept Erasure in Diffusion Models},
    author={Yimeng Zhang and Xin Chen and Jinghan Jia and Yihua Zhang and Chongyu Fan and Jiancheng Liu and Mingyi Hong and Ke Ding and Sijia Liu},
    booktitle={The Thirty-eighth Annual Conference on Neural Information Processing Systems},
    year={2024},
    url={https://openreview.net/forum?id=dkpmfIydrF}
}

@inproceedings{
    pham2024circumventing,
    title={Circumventing Concept Erasure Methods For Text-To-Image Generative Models},
    author={Minh Pham and Kelly O. Marshall and Niv Cohen and Govind Mittal and Chinmay Hegde},
    booktitle={The Twelfth International Conference on Learning Representations},
    year={2024},
    url={https://openreview.net/forum?id=ag3o2T51Ht}
}

@inproceedings{lyu2023spm,
    title={One-Dimensional Adapter to Rule Them All: Concepts, Diffusion Models and Erasing Applications}, 
    author={Mengyao Lyu and Yuhong Yang and Haiwen Hong and Hui Chen and Xuan Jin and Yuan He and Hui Xue and Jungong Han and Guiguang Ding},
    booktitle={2024 IEEE/CVF Conference on Computer Vision and Pattern Recognition (CVPR)},
    year={2024}
  }

@inproceedings{
    heng2023selective,
    title={Selective Amnesia: A Continual Learning Approach to Forgetting in Deep Generative Models},
    author={Alvin Heng and Harold Soh},
    booktitle={Thirty-seventh Conference on Neural Information Processing Systems},
    year={2023},
    url={https://openreview.net/forum?id=BC1IJdsuYB}
}

@inproceedings{receler,
    author = {Huang, Chi-Pin and Chang, Kai-Po and Tsai, Chung-Ting and Lai, Yung-Hsuan and Yang, Fu-En and Wang, Yu-Chiang Frank},
    title = {Receler: Reliable Concept Erasing of Text-to-Image Diffusion Models via Lightweight Erasers},
    year = {2024},
    isbn = {978-3-031-73660-5},
    publisher = {Springer-Verlag},
    address = {Berlin, Heidelberg},
    url = {https://doi.org/10.1007/978-3-031-73661-2_20},
    doi = {10.1007/978-3-031-73661-2_20},
    booktitle = {ECCV},
    pages = {360–376},
    numpages = {17},
    location = {Milan, Italy}
}

@inproceedings{
    fan2024salun,
    title={SalUn: Empowering Machine Unlearning via Gradient-based Weight Saliency in Both Image Classification and Generation},
    author={Chongyu Fan and Jiancheng Liu and Yihua Zhang and Eric Wong and Dennis Wei and Sijia Liu},
    booktitle={The Twelfth International Conference on Learning Representations},
    year={2024},
    url={https://openreview.net/forum?id=gn0mIhQGNM}
}

@inproceedings{ac,
  author = {Kumari, Nupur and Zhang, Bingliang and Wang, Sheng-Yu and Shechtman, Eli and Zhang, Richard and Zhu, Jun-Yan},
  title = {Ablating Concepts in Text-to-Image Diffusion Models},
  booktitle = {ICCV},
  year = {2023},
}

@misc{wu2024erasingundesirableinfluencediffusion,
      title={Erasing Undesirable Influence in Diffusion Models}, 
      author={Jing Wu and Trung Le and Munawar Hayat and Mehrtash Harandi},
      year={2024},
      eprint={2401.05779},
      archivePrefix={arXiv},
      primaryClass={cs.CV},
      url={https://arxiv.org/abs/2401.05779}, 
}

@misc{lu2024macemassconcepterasure,
      title={MACE: Mass Concept Erasure in Diffusion Models}, 
      author={Shilin Lu and Zilan Wang and Leyang Li and Yanzhu Liu and Adams Wai-Kin Kong},
      year={2024},
      eprint={2403.06135},
      archivePrefix={arXiv},
      primaryClass={cs.CV},
      url={https://arxiv.org/abs/2403.06135}, 
}

@misc{cunningham2023sparseautoencodershighlyinterpretable,
      title={Sparse Autoencoders Find Highly Interpretable Features in Language Models}, 
      author={Hoagy Cunningham and Aidan Ewart and Logan Riggs and Robert Huben and Lee Sharkey},
      year={2023},
      eprint={2309.08600},
      archivePrefix={arXiv},
      primaryClass={cs.LG},
      url={https://arxiv.org/abs/2309.08600}, 
}

@article{srivatsan2024stereo,
  title={STEREO: A Two-Stage Framework for Adversarially Robust Concept Erasing from Text-to-Image Diffusion Models},
  author={Srivatsan, Koushik and Shamshad, Fahad and Naseer, Muzammal and Patel, Vishal M and Nandakumar, Karthik},
  journal={arXiv preprint arXiv:2408.16807},
  year={2024}
}

@misc{zhang2023fmn,
      title={Forget-Me-Not: Learning to Forget in Text-to-Image Diffusion Models}, 
      author={Eric Zhang and Kai Wang and Xingqian Xu and Zhangyang Wang and Humphrey Shi},
      year={2023},
      eprint={2303.17591},
      archivePrefix={arXiv},
      primaryClass={cs.CV},
      url={https://arxiv.org/abs/2303.17591}, 
}

@inproceedings{
    yuksekgonul2023when,
    title={When and Why Vision-Language Models Behave like Bags-Of-Words, and What to Do About It?},
    author={Mert Yuksekgonul and Federico Bianchi and Pratyusha Kalluri and Dan Jurafsky and James Zou},
    booktitle={The Eleventh International Conference on Learning Representations },
    year={2023},
    url={https://openreview.net/forum?id=KRLUvxh8uaX}
}

@inproceedings{
    chefer2024the,
    title={The Hidden Language of Diffusion Models},
    author={Hila Chefer and Oran Lang and Mor Geva and Volodymyr Polosukhin and Assaf Shocher and Michal Irani and Inbar Mosseri and Lior Wolf},
    booktitle={The Twelfth International Conference on Learning Representations},
    year={2024},
    url={https://openreview.net/forum?id=awWpHnEJDw}
}

@misc{gong2024reliableefficientconcepterasure,
      title={Reliable and Efficient Concept Erasure of Text-to-Image Diffusion Models}, 
      author={Chao Gong and Kai Chen and Zhipeng Wei and Jingjing Chen and Yu-Gang Jiang},
      year={2024},
      eprint={2407.12383},
      archivePrefix={arXiv},
      primaryClass={cs.CV},
      url={https://arxiv.org/abs/2407.12383}, 
}

@misc{nudenetclassifier,
  author       = {Platelminto},
  title        = {{NudeNetClassifier}: A classifier for NSFW content detection},
  howpublished = {\url{https://github.com/platelminto/NudeNetClassifier}},
  year         = {2024},
  note         = {Accessed: 2025-05-09}
}

@article{Ramesh2022HierarchicalTI,
  title={Hierarchical Text-Conditional Image Generation with CLIP Latents},
  author={Aditya Ramesh and Prafulla Dhariwal and Alex Nichol and Casey Chu and Mark Chen},
  journal={ArXiv},
  year={2022},
  volume={abs/2204.06125},
  url={https://api.semanticscholar.org/CorpusID:248097655}
}

@inproceedings{scenet2i,
    author = {Gafni, Oran and Polyak, Adam and Ashual, Oron and Sheynin, Shelly and Parikh, Devi and Taigman, Yaniv},
    title = {Make-A-Scene: Scene-Based Text-to-Image Generation with Human Priors},
    year = {2022},
    isbn = {978-3-031-19783-3},
    publisher = {Springer-Verlag},
    address = {Berlin, Heidelberg},
    url = {https://doi.org/10.1007/978-3-031-19784-0_6},
    booktitle = {Computer Vision – ECCV 2022: 17th European Conference, Tel Aviv, Israel, October 23–27, 2022, Proceedings, Part XV},
    pages = {89–106},
    numpages = {18},
    location = {Tel Aviv, Israel}
}

@InProceedings{muse_t2i,
  title = 	 {Muse: Text-To-Image Generation via Masked Generative Transformers},
  author =       {Chang, Huiwen and Zhang, Han and Barber, Jarred and Maschinot, Aaron and Lezama, Jose and Jiang, Lu and Yang, Ming-Hsuan and Murphy, Kevin Patrick and Freeman, William T. and Rubinstein, Michael and Li, Yuanzhen and Krishnan, Dilip},
  booktitle = 	 {Proceedings of the 40th International Conference on Machine Learning},
  pages = 	 {4055--4075},
  year = 	 {2023},
  editor = 	 {Krause, Andreas and Brunskill, Emma and Cho, Kyunghyun and Engelhardt, Barbara and Sabato, Sivan and Scarlett, Jonathan},
  volume = 	 {202},
  series = 	 {Proceedings of Machine Learning Research},
  month = 	 {23--29 Jul},
  publisher =    {PMLR},
  url = 	 {https://proceedings.mlr.press/v202/chang23b.html},
}

@misc{liu2024incontextvectorsmakingcontext,
      title={In-context Vectors: Making In Context Learning More Effective and Controllable Through Latent Space Steering}, 
      author={Sheng Liu and Haotian Ye and Lei Xing and James Zou},
      year={2024},
      eprint={2311.06668},
      archivePrefix={arXiv},
      primaryClass={cs.LG},
      url={https://arxiv.org/abs/2311.06668}, 
}

@misc{musurvey,
      title={A Survey of Machine Unlearning}, 
      author={Thanh Tam Nguyen and Thanh Trung Huynh and Zhao Ren and Phi Le Nguyen and Alan Wee-Chung Liew and Hongzhi Yin and Quoc Viet Hung Nguyen},
      year={2024},
      eprint={2209.02299},
      archivePrefix={arXiv},
      primaryClass={cs.LG},
      url={https://arxiv.org/abs/2209.02299}, 
}

@inproceedings{coco,
  title={Microsoft COCO: Common Objects in Context},
  author={Tsung-Yi Lin and Michael Maire and Serge J. Belongie and James Hays and Pietro Perona and Deva Ramanan and Piotr Doll{\'a}r and C. Lawrence Zitnick},
  booktitle={European Conference on Computer Vision},
  year={2014},
  url={https://api.semanticscholar.org/CorpusID:14113767}
}

@InProceedings{Zhou_2018_ECCV,
    author = {Zhou, Bolei and Sun, Yiyou and Bau, David and Torralba, Antonio},
    title = {Interpretable Basis Decomposition for Visual Explanation},
    booktitle = {Proceedings of the European Conference on Computer Vision (ECCV)},
    month = {September},
    year = {2018}
}

@inproceedings{
    gal2023textinverse,
    title={An Image is Worth One Word: Personalizing Text-to-Image Generation using Textual Inversion},
    author={Rinon Gal and Yuval Alaluf and Yuval Atzmon and Or Patashnik and Amit Haim Bermano and Gal Chechik and Daniel Cohen-or},
    booktitle={The Eleventh International Conference on Learning Representations },
    year={2023},
    url={https://openreview.net/forum?id=NAQvF08TcyG}
}

@inproceedings{
    park2023thelinear,
    title={The Linear Representation Hypothesis and the Geometry of Large Language Models},
    author={Kiho Park and Yo Joong Choe and Victor Veitch},
    booktitle={Causal Representation Learning Workshop at NeurIPS 2023},
    year={2023},
    url={https://openreview.net/forum?id=T0PoOJg8cK}
}

@inproceedings{
    bhalla2024cliplinear,
    title={Interpreting {CLIP} with Sparse Linear Concept Embeddings (SpLi{CE})},
    author={Usha Bhalla and Alex Oesterling and Suraj Srinivas and Flavio Calmon and Himabindu Lakkaraju},
    booktitle={The Thirty-eighth Annual Conference on Neural Information Processing Systems},
    year={2024},
    url={https://openreview.net/forum?id=7UyBKTFrtd}
}

@article{zhang2024unlearndiff,
  title={To generate or not? safety-driven unlearned diffusion models are still easy to generate unsafe images... for now},
  author={Zhang, Yimeng and Jia, Jinghan and Chen, Xin and Chen, Aochuan and Zhang, Yihua and Liu, Jiancheng and Ding, Ke and Liu, Sijia},
  journal={European Conference on Computer Vision (ECCV)},
  year={2024}
}

@article{gandikota2024uce,
  title={Unified Concept Editing in Diffusion Models},
  author={Rohit Gandikota and Hadas Orgad and Yonatan Belinkov and Joanna Materzy\'nska and David Bau},
  journal={IEEE/CVF Winter Conference on Applications of Computer Vision},
  year={2024}
}

@inbook{deletion,
    author = {Ginart, Antonio A. and Guan, Melody Y. and Valiant, Gregory and Zou, James},
    title = {Making AI forget you: data deletion in machine learning},
    year = {2019},
    publisher = {Curran Associates Inc.},
    address = {Red Hook, NY, USA},
    booktitle = {Proceedings of the 33rd International Conference on Neural Information Processing Systems},
    articleno = {316},
    numpages = {14}
}

@inproceedings{redteam,
  title={Prompting4Debugging: Red-Teaming Text-to-Image Diffusion Models by Finding Problematic Prompts},
  author={Zhi-Yi Chin and Chieh-Ming Jiang and Ching-Chun Huang and Pin-Yu Chen and Wei-Chen Chiu},
  booktitle={International Conference on Machine Learning (ICML)},
  year={2024},
  url={https://arxiv.org/abs/2309.06135}
}

@inproceedings{schramowski2022safe,
      title={Safe Latent Diffusion: Mitigating Inappropriate Degeneration in Diffusion Models}, 
      author={Patrick Schramowski and Manuel Brack and Björn Deiseroth and Kristian Kersting},
      year={2023},
      booktitle={Proceedings of the {IEEE} Conference on Computer Vision and Pattern Recognition ({CVPR})},
}

@inproceedings{
Hertz2022PrompttoPromptIE,
    title={Prompt-to-Prompt Image Editing with Cross-Attention Control},
    author={Amir Hertz and Ron Mokady and Jay Tenenbaum and Kfir Aberman and Yael Pritch and Daniel Cohen-or},
    booktitle={The Eleventh International Conference on Learning Representations },
    year={2023},
    url={https://openreview.net/forum?id=_CDixzkzeyb}
}

@misc{
    zhang2024the,
    title={The Emergence of Reproducibility and Consistency in Diffusion Models},
    author={Huijie Zhang and Jinfan Zhou and Yifu Lu and Minzhe Guo and Liyue Shen and Qing Qu},
    year={2024},
    url={https://openreview.net/forum?id=UkLSvLqiO7}
}

@ARTICLE{omp,
  author={Tropp, Joel A. and Gilbert, Anna C.},
  journal={IEEE Transactions on Information Theory}, 
  title={Signal Recovery From Random Measurements Via Orthogonal Matching Pursuit}, 
  year={2007},
  volume={53},
  number={12},
  pages={4655-4666},
  doi={10.1109/TIT.2007.909108}}

@article{He2015DeepRL,
  title={Deep Residual Learning for Image Recognition},
  author={Kaiming He and X. Zhang and Shaoqing Ren and Jian Sun},
  journal={2016 IEEE Conference on Computer Vision and Pattern Recognition (CVPR)},
  year={2015},
  pages={770-778},
  url={https://api.semanticscholar.org/CorpusID:206594692}
}

@article{Bau2017NetworkDQ,
  title={Network Dissection: Quantifying Interpretability of Deep Visual Representations},
  author={David Bau and Bolei Zhou and Aditya Khosla and Aude Oliva and Antonio Torralba},
  journal={2017 IEEE Conference on Computer Vision and Pattern Recognition (CVPR)},
  year={2017},
  pages={3319-3327},
  url={https://api.semanticscholar.org/CorpusID:378410}
}

@inproceedings{
    fel2023a,
    title={A Holistic Approach to Unifying Automatic Concept Extraction and Concept Importance Estimation},
    author={Thomas FEL and Victor Boutin and Louis B{\'e}thune and Remi Cadene and Mazda Moayeri and L{\'e}o And{\'e}ol and Mathieu Chalvidal and Thomas Serre},
    booktitle={Thirty-seventh Conference on Neural Information Processing Systems},
    year={2023},
    url={https://openreview.net/forum?id=MziFFGjpkb}
}

@article{46832,title	= {Feature Visualization},author	= {Christopher Olah and Ludwig Schubert and Alexander Mordvintsev},year	= {2017},URL	= {https://distill.pub/2017/feature-visualization/},journal	= {Distill}}

@INPROCEEDINGS {svdiff,
    author = { Han, Ligong and Li, Yinxiao and Zhang, Han and Milanfar, Peyman and Metaxas, Dimitris and Yang, Feng },
    booktitle = {2023 IEEE/CVF International Conference on Computer Vision (ICCV) },
    title = {{ SVDiff: Compact Parameter Space for Diffusion Fine-Tuning }},
    year = {2023},
    volume = {},
    ISSN = {},
    pages = {7289-7300},
    doi = {10.1109/ICCV51070.2023.00673},
    url = {https://doi.ieeecomputersociety.org/10.1109/ICCV51070.2023.00673},
    publisher = {IEEE Computer Society},
    address = {Los Alamitos, CA, USA},
    month =Oct
}

@article{CLIPScore,
  title={CLIPScore: A Reference-free Evaluation Metric for Image Captioning},
  author={Jack Hessel and Ari Holtzman and Maxwell Forbes and Ronan Le Bras and Yejin Choi},
  journal={ArXiv},
  year={2021},
  volume={abs/2104.08718},
  url={https://api.semanticscholar.org/CorpusID:233296711}
}

@inproceedings{
    park2023understanding,
    title={Understanding the Latent Space of Diffusion Models through the Lens of Riemannian Geometry},
    author={Yong-Hyun Park and Mingi Kwon and Jaewoong Choi and Junghyo Jo and Youngjung Uh},
    booktitle={Thirty-seventh Conference on Neural Information Processing Systems},
    year={2023},
    url={https://openreview.net/forum?id=VUlYp3jiEI}
}

@Article{ddpm,
  author  = {Ho, Jonathan and Jain, Ajay and Abbeel, Pieter},
  journal = {Advances in Neural Information Processing Systems},
  title   = {Denoising diffusion probabilistic models},
  year    = {2020},
  pages   = {6840--6851},
  volume  = {33},
}

@inproceedings{
    hspace,
    title={Diffusion Models Already Have A Semantic Latent Space},
    author={Mingi Kwon and Jaeseok Jeong and Youngjung Uh},
    booktitle={The Eleventh International Conference on Learning Representations },
    year={2023},
    url={https://openreview.net/forum?id=pd1P2eUBVfq}
}

@inproceedings{chen2024loco,
    title={Exploring Low-Dimensional Subspace in Diffusion Models for Controllable Image Editing},
    author={Siyi Chen and Huijie Zhang and Minzhe Guo and Yifu Lu and Peng Wang and Qing Qu},
    booktitle={The Thirty-eighth Annual Conference on Neural Information Processing Systems},
    year={2024},
    url={https://arxiv.org/abs/2409.02374}
    }

@article{discovering,
  title={Discovering Interpretable Directions in the Semantic Latent Space of Diffusion Models},
  author={Ren{\'e} Haas and Inbar Huberman-Spiegelglas and Rotem Mulayoff and Tomer Michaeli},
  journal={International Conference on Automatic Face and Gesture Recognition},
  year={2024},
  volume={abs/2303.11073},
  url={https://api.semanticscholar.org/CorpusID:257631803}
}

@inproceedings{stablediffusion,
  title={High-resolution image synthesis with latent diffusion models},
  author={Rombach, Robin and Blattmann, Andreas and Lorenz, Dominik and Esser, Patrick and Ommer, Bj{\"o}rn},
  booktitle={Proceedings of the IEEE/CVF Conference on Computer Vision and Pattern Recognition},
  pages={10684--10695},
  year={2022}
}

@article{LCM,
  title={Latent consistency models: Synthesizing high-resolution images with few-step inference},
  author={Luo, Simian and Tan, Yiqin and Huang, Longbo and Li, Jian and Zhao, Hang},
  journal={ArXiv preprint arXiv:2310.04378},
  year={2023}
}

@misc{xu2024versatilediffusiontextimages,
      title={Versatile Diffusion: Text, Images and Variations All in One Diffusion Model}, 
      author={Xingqian Xu and Zhangyang Wang and Eric Zhang and Kai Wang and Humphrey Shi},
      year={2024},
      eprint={2211.08332},
      archivePrefix={arXiv},
      primaryClass={cs.CV},
      url={https://arxiv.org/abs/2211.08332}, 
}

@article{
    yu2022scaling,
    title={Scaling Autoregressive Models for Content-Rich Text-to-Image Generation},
    author={Jiahui Yu and Yuanzhong Xu and Jing Yu Koh and Thang Luong and Gunjan Baid and Zirui Wang and Vijay Vasudevan and Alexander Ku and Yinfei Yang and Burcu Karagol Ayan and Ben Hutchinson and Wei Han and Zarana Parekh and Xin Li and Han Zhang and Jason Baldridge and Yonghui Wu},
    journal={Transactions on Machine Learning Research},
    issn={2835-8856},
    year={2022},
    url={https://openreview.net/forum?id=AFDcYJKhND},
    note={Featured Certification}
}

@misc{glide,
      title={GLIDE: Towards Photorealistic Image Generation and Editing with Text-Guided Diffusion Models}, 
      author={Alex Nichol and Prafulla Dhariwal and Aditya Ramesh and Pranav Shyam and Pamela Mishkin and Bob McGrew and Ilya Sutskever and Mark Chen},
      year={2022},
      eprint={2112.10741},
      archivePrefix={arXiv},
      primaryClass={cs.CV},
      url={https://arxiv.org/abs/2112.10741}, 
}

@article{DeepFloydIF,
  title={Photorealistic text-to-image diffusion models with deep language understanding},
  author={Saharia, Chitwan and Chan, William and Saxena, Saurabh and Li, Lala and Whang, Jay and Denton, Emily L and Ghasemipour, Kamyar and Gontijo Lopes, Raphael and Karagol Ayan, Burcu and Salimans, Tim and others},
  journal={Advances in Neural Information Processing Systems},
  volume={35},
  pages={36479--36494},
  year={2022}
}

@inproceedings{khachatryan2023text2video,
  title={Text2video-zero: Text-to-image diffusion models are zero-shot video generators},
  author={Khachatryan, Levon and Movsisyan, Andranik and Tadevosyan, Vahram and Henschel, Roberto and Wang, Zhangyang and Navasardyan, Shant and Shi, Humphrey},
  booktitle={Proceedings of the IEEE/CVF International Conference on Computer Vision},
  pages={15954--15964},
  year={2023}
}

@article{dalle2,
  title={Hierarchical text-conditional image generation with clip latents},
  author={Ramesh, Aditya and Dhariwal, Prafulla and Nichol, Alex and Chu, Casey and Chen, Mark},
  journal={ArXiv preprint arXiv:2204.06125},
  year={2022}
}

@inproceedings{imagenet,
  title={Imagenet: A large-scale hierarchical image database},
  author={Deng, Jia and Dong, Wei and Socher, Richard and Li, Li-Jia and Li, Kai and Fei-Fei, Li},
  booktitle={2009 IEEE Conference on Computer Vision and Pattern Recognition},
  pages={248--255},
  year={2009},
  organization={Ieee}
}

@misc{yang2024mmadiffusionmultimodalattackdiffusion,
      title={MMA-Diffusion: MultiModal Attack on Diffusion Models}, 
      author={Yijun Yang and Ruiyuan Gao and Xiaosen Wang and Tsung-Yi Ho and Nan Xu and Qiang Xu},
      year={2024},
      eprint={2311.17516},
      archivePrefix={arXiv},
      primaryClass={cs.CR},
      url={https://arxiv.org/abs/2311.17516}, 
}

@misc{maus2023blackboxadversarialprompting,
      title={Black Box Adversarial Prompting for Foundation Models}, 
      author={Natalie Maus and Patrick Chao and Eric Wong and Jacob Gardner},
      year={2023},
      eprint={2302.04237},
      archivePrefix={arXiv},
      primaryClass={cs.LG},
      url={https://arxiv.org/abs/2302.04237}, 
}

@misc{yang2023sneaky,
      title={SneakyPrompt: Jailbreaking Text-to-image Generative Models}, 
      author={Yuchen Yang and Bo Hui and Haolin Yuan and Neil Gong and Yinzhi Cao},
      year={2023},
      eprint={2305.12082},
      archivePrefix={arXiv},
      primaryClass={cs.LG},
      url={https://arxiv.org/abs/2305.12082}, 
}

@inproceedings{zhu2023boundary,
  title={Boundary Guided Learning-Free Semantic Control with Diffusion Models},
  author={Zhu, Ye and Wu, Yu and Deng, Zhiwei and Russakovsky, Olga and Yan, Yan},
  booktitle={Conference on Neural Information Processing Systems (NeurIPS)},
  year={2023}
}

@article{wang2023conceptalgebra,
  title={Concept Algebra for (Score-Based) Text-Controlled Generative Models},
  author={Wang, Zihao and Gui, Lin and Negrea, Jeffrey and Veitch, Victor},
  journal={arXiv preprint arXiv:2302.03693},
  year={2023},
  url={https://arxiv.org/abs/2302.03693}
}

@inproceedings{clip,
  title={Learning transferable visual models from natural language supervision},
  author={Radford, Alec and Kim, Jong Wook and Hallacy, Chris and Ramesh, Aditya and Goh, Gabriel and Agarwal, Sandhini and Sastry, Girish and Askell, Amanda and Mishkin, Pamela and Clark, Jack and others},
  booktitle={International Conference on Machine Learning},
  pages={8748--8763},
  year={2021},
  organization={PMLR}
}

@article{zhang2024id,
  title={ID-Patch: Robust ID Association for Group Photo Personalization},
  author={Zhang, Yimeng and Zhi, Tiancheng and Liu, Jing and Sang, Shen and Jiang, Liming and Yan, Qing and Liu, Sijia and Luo, Linjie},
  journal={arXiv preprint arXiv:2411.13632},
  year={2024}
}

@article{zhang2024unlearncanvas,
  title={Unlearncanvas: A stylized image dataset to benchmark machine unlearning for diffusion models},
  author={Zhang, Yihua and Zhang, Yimeng and Yao, Yuguang and Jia, Jinghan and Liu, Jiancheng and Liu, Xiaoming and Liu, Sijia},
  journal={arXiv e-prints},
  pages={arXiv--2402},
  year={2024}
}

\clearpage

\appendix

\section*{Broader Impact Statement}
This work exposes that concepts ``erased'' by existing unlearning methods often persist as an interpretable residual subspace, and introduces an attack (SubAttack) that reads it out to regenerate the erased concept. Like any offensive-security research, this carries a misuse risk, which we mitigate on several fronts: the contribution is fundamentally diagnostic and is paired with a defense (SubDefense) that projects the same subspace out to reduce regeneration; the method assumes white-box access and grants no new capability against black-box or API-gated deployments; and we redact sensitive imagery and release code for defenders rather than to lower the barrier for misuse. We hope this encourages treating concept erasure as an ongoing robustness problem addressed with subspace-level defenses.

\section*{Appendix Overview}
\label{appsec:overview}

The appendix is organized as follows:
\begin{itemize}
    \item \textbf{\Cref{appsec:related_works} (Related Works).} Prior work on
    T2I diffusion models and machine unlearning, jailbreaking attacks and
    defenses on unlearned models, diffusion-model interpretability, and the
    linear representation hypothesis that motivates our approach.
    \item \textbf{\Cref{appsec:exp_setting} (Experiment Settings).} Full attack
    and defense protocols: victim models, the attacking dataset, MLP learning
    details, how each attack modifies the prompt, and the evaluation metrics
    (ASR classifiers, CLIP score, FID).
    \item \textbf{\Cref{appsec:aux_attack} (Auxiliary Attack Results).}
    Additional interpretations of the learned attack token embeddings across
    models/concepts, extended ASR results on more unlearned models and settings
    (MACE, SA, AC, SalUn, EraseDiff, RECE), and transfer-attack comparisons.
    \item \textbf{\Cref{appsec:aux_defense} (Auxiliary Defense Results).}
    Detailed RECE-vs-SubDefense baseline comparison on UCE, defense against
    UnlearnDiff on the I2P dataset, defense against SubAttack across concepts and
    models, and exploratory settings (black-box Ring-A-Bell defense, standalone
    SubDefense).
    \item \textbf{\Cref{appsec:ablations} (Ablations).} Attack ablations (number
    of attack tokens $K$, orthogonality constraint, vocabulary size) and defense
    ablations (utility degradation vs.\ number of blocked tokens, defending
    against CCE).
    \item \textbf{\Cref{app-sec:sparse} (Sparsity of Attack Token Embeddings).}
    Analysis of the sparsity of the learned coefficients $\alpha_i$ and how it
    evolves over the iterative attack.
    \item \textbf{\Cref{appsec:uce_vis} (Image Generation Quality After
    Defense).} Qualitative COCO generations before/after SubDefense.
    \item \textbf{\Cref{appsec:attack_vis} (More Attack Visualizations).}
    Additional qualitative attack results across concepts and models.
    \item \textbf{\Cref{appsec_discussion} (Future Directions).} Limitations and
    directions for future work.
\end{itemize}

\section{Related Works}
\label{appsec:related_works}


%

\paragraph{T2I Diffusion Models and Machine Unlearning.} Text-to-image (T2I) diffusion models \cite{stablediffusion,muse_t2i,LCM,DeepFloydIF,scenet2i,dalle2,yu2022scaling,xu2024versatilediffusiontextimages} can take prompts as input and generate desired images following the prompt. There are several different types of T2I models, such as stable diffusion \cite{stablediffusion}, latent consistency model \cite{LCM}, and DeepFloyd \cite{DeepFloydIF}. Despite their generation ability, safety concerns arise since these models have also gained the ability to generate unwanted images that are harmful or violate copyright. To solve this problem, some early works deploy safety filters \cite{glide,stablediffusion} or modified inference guidance \cite{schramowski2022safe} but exhibit limited robustness \cite{redteam,yang2024mmadiffusionmultimodalattackdiffusion}. Recently, machine unlearning (MU) \cite{musurvey,deletion} is one of the major strategies that makes the model ``forget'' one specific concept via fine-tuning, and most MU works build on the widely used latent diffusion models (LDM), specifically stable diffusion (SD) models. Most diffusion machine unlearning works finetune the denoising UNets \cite{esd,zhang2023fmn,lyu2023spm,ac,gandikota2024uce,fan2024salun,receler,heng2023selective}. Although MU is a more practical solution than filtering datasets and retraining models from scratch, the robustness of MU still needs careful attention. Although current diffusion unlearning methods typically target the removal of a single concept per model, the need to preserve safe concept generation makes complete removal a challenging problem.

\paragraph{Jailbreaking Attacks and Defenses on Unlearned Models.} Recent works explore jailbreaking attacks on unlearned diffusion models, which aim to make unlearned models regenerate unwanted concepts. Such attacks can serve as a way to evaluate the robustness of unlearned diffusion models. For example, UnlearnDiff \cite{zhang2024unlearndiff} learns an adversarial attack prompt and appends the prompt before the original text prompt to do attacks, along a similar line of prior attack works \cite{yang2023sneaky, maus2023blackboxadversarialprompting,chin2024prompting4debugging,tsai2024ringabell,zhuang2023a}. Besides, the most related work to ours is \cite{pham2024circumventing}, utilizing Textual Inversion \cite{gal2023textinverse}. It also learns a token embedding that represents the target concept. Though we experimentally show CCE is in nature global to both text prompts and random noise as well, but is less transferable to different unlearned models. Prior jailbreaking attacks also do not consider the interpretability of the resulting attack prompts, thus offering limited insights into the underlying causes of the deficiencies in current unlearning methods, nor do they explore the potential for defense. In contrast, our attack token embeddings are interpretable and reveal the human-interpretable associations remained in unlearned diffusion models to ``remember'' the target concepts. Also, our method can be easily extended to learn a diverse set of attack token embeddings independent of each other. This diversity sheds light on the volume of the inner space where the target concept is still hidden. This motivates us to propose a simple yet effective defense method against existing attack methods. To the best of our knowledge, the defense of unlearned models is an underexplored problem in the field. A recent work, RECE \cite{gong2024reliableefficientconcepterasure}, targets a specific unlearned model (i.e., UCE \cite{gandikota2024uce}), and focuses on defending it against adversarial attacks (i.e., UnlearnDiff). Defending a broader range of unlearned models against diverse attack types remains a challenging problem—one we aim to address by leveraging our defense.

\paragraph{Diffusion Model Interpretability.} To understand the semantics within diffusion models for applications such as image editing and decomposition, a series of works have attempted to interpret the representation space within diffusion models \cite{hspace,park2023understanding,chen2024loco,chefer2024the}. For example, \cite{hspace} studies the semantic correspondences in the middle layer of the denoising UNet in diffusion models, while \cite{chen2024loco} investigates the low-rank subspace spanned in the noise space. \rebuttal{Related work also discovers interpretable semantic directions and subspace boundaries in diffusion representations \citep{discovering,zhu2023boundary}, while Concept Algebra formalizes concepts as subspaces in score-based text-controlled generative models \citep{wang2023conceptalgebra}.} Some works \cite{Hertz2022PrompttoPromptIE,svdiff} focus on the visualization of attention maps with respect to input texts, while other works study the generalization and memorization perspective of diffusion models \cite{zhang2024the}. The most related work to ours is \cite{chefer2024the}, which decomposes a single concept as a combination of a weighted combination of interpretable elements, in line with the concept decomposition and visualization works in a wider domain \cite{46832,fel2023a,Bau2017NetworkDQ}. Inspired by \cite{chefer2024the} as well as other prior works, we attack unlearned diffusion models by learning interpretable representations, which leads to further investigation on the root of failures for existing unlearned diffusion models, as well as a defense method.

\paragraph{Linear Representation Hypothesis.} In large language models (LLMs), the linear representation hypothesis posits that certain features and concepts learned by LLMs are encoded as linear vectors in their high-dimensional embedding spaces. This is supported by the fact that adding or subtracting specific vectors can manipulate a sentence's sentiment or extract specific semantic meanings \cite{park2023thelinear}. The linear property has been further explored for understanding, detoxing, and controlling the generation of LLMs \cite{liu2024incontextvectorsmakingcontext}. Similarly, other works investigating the representations of multimodal models find that concepts are encoded additively \cite{clip, yuksekgonul2023when}, and concepts can be decomposed by human-interpretable words \cite{bhalla2024cliplinear}. Moreover, in stable diffusion models, \cite{chefer2024the} finds that concepts can be decomposed in the CLIP token embedding space in a bag-of-words manner. Based on these works, and considering the flexibility of the token embedding space in diffusion personalization \cite{gal2023textinverse} and attacking \cite{pham2024circumventing}, we specifically investigate interpretable jailbreaking attacks and defenses for diffusion model unlearning by learning an attack token embedding that is a linear combination of existing token embeddings.


\section{Experiment Settings}
\label{appsec:exp_setting}


\subsection{Attack}
\label{exp_setting_attack}

\paragraph{Unlearned LDMs as Victim Models.} The field of diffusion unlearning is evolving rapidly, and there is a wide range of unlearning methods, most of which finetune the stable diffusion model. Most of the existing methods focus on single-concept unlearning. Following the protocol of \cite{zhang2024unlearndiff}, we select several unlearned diffusion models that have an open-source and reproducible codebase, reasonable unlearning performance, and reasonable generation quality. This selection includes three widely used models from prior jailbreaking studies, namely ESD \cite{esd}, FMN \cite{zhang2023fmn}, and UCE \cite{gandikota2024uce}, as well as more recent or complementary settings such as SPM \citep{lyu2023spm}, MACE \citep{lu2024macemassconcepterasure}, SA \citep{heng2023selective}, AC \citep{ac}, SalUn \citep{fan2023salun}, and EraseDiff \citep{wu2024erasingundesirableinfluencediffusion}. These methods fine-tune the denoising UNet for unlearning while freezing other components. In our study, the unlearned models are fine-tuned on Stable Diffusion v1.4, and hence, they share the same CLIP text encoders.

\paragraph{Attacking Dataset.} Our learned token embedding represents the target concept, so the attack token embedding in nature can attack the victim model with different initial noise and text prompts. Thus, we construct a dataset to test such global attacking ability. To facilitate reproducibility, we follow the dataset construction protocol of UnlearnDiff as follows. We study three kinds of target concepts: ``nudity'' for NSFW, ``Van Gogh'' for artistic styles, and ``church'', ``garbage truck'', ``parachute'', and ``tench'' for objects. For each of ``nudity'', ``Van Gogh'', and ``church'', we prepare a corresponding dataset containing 900 (prompt, seed) pairs, and mainly use these concepts for baseline comparisons with other attacks. For each of the other concepts, we prepare a dataset of size 300. Each prompt contains the target concept to attack - for instance, ``a photo of a nude woman in a sunlit garden'' is an example prompt in the ``nudity'' dataset. Each prompt is associated with 10 - 30 different random seeds controlling the initial noise, and this results in a total of 300 - 900 (prompt, seed) pairs for each concept. Each pair is verified to produce the target concept with the original SD v1.4. Our dataset is approximately six times larger than that used in UnlearnDiff, enabling more reliable evaluation.

\paragraph{Learning Details.} We use SD 1.4 to generate 100 images containing the target concept as the training image dataset. The prompt used to generate images for each concept is similar to ``A photo of a [target concept]''. After that, to optimize each of the attack token embeddings for conducting SubAttack, we train an MLP network using the AdamW optimizer for 500 epochs with a batch size of 6. The MLP consists of two linear layers with ReLU activation applied after each layer. The first layer maps from 768 to 100 dimensions, and the second maps from 100 to 1. Experimental results confirm that this design has sufficient capacity to learn the scalar $\alpha_i$ for each embedding in the vocabulary. All experiments are conducted on a single NVIDIA A40 GPU.

\paragraph{Attacking Details.} For NoAttack, the original text prompts and seeds are passed to the victim model. In SubAttack and CCE attacks, we replace the target concept in the text prompt with the special token associated with the learned attack token embedding (For example, change ``a photo of a nude woman'' to ``a photo of a <$\bm v_{att}$>''). In UnlearnDiff, we modify each text prompt by appending the corresponding learned adversarial prompt before it. For each attacking method and each concept, we generate 300-900 images using the resulting (prompt, seed) pairs for testing attack performance. 

\paragraph{Evaluation Protocols.} (\emph{i}) After image generation, we use pretrained classifiers to detect the percentage of images containing the target concept following UnlearnDiff, and report it as the attacking success rate (ASR). For nudity, we use NudeNet \cite{zhang2024unlearndiff} to detect the existence of nudity subjects. For Van Gogh, we deploy the style classifier finetuned on the WikiArt dataset and released by \cite{zhang2024unlearndiff}. We report the Top-3 ASR for style, i.e., if Van Gogh is predicted within the Top-3 style classes for a generated image, the image is viewed as a successful attack for Van Gogh style. For church, the object classifier pretrained on ImageNet \cite{imagenet} using the ResNet-50 \cite{He2015DeepRL} architecture is utilized. \rebuttal{(\emph{ii}) Runtime includes attack construction and image generation. For CCE and SubAttack, the reusable concept-level construction cost is amortized over the 900 prompt--seed pairs in each main comparison: $T_{\mathrm{avg}}(N)=(T_{\mathrm{construct}}+N T_{\mathrm{apply}})/N$, with $N=900$. CCE learns one global token and generates one candidate per pair; default SubAttack learns five global directions and generates five candidates per pair. UnlearnDiff instead optimizes a new local prompt for every pair, so its optimization is not amortized. A direct stage-wise breakdown appears in \Cref{appsubsec:efficiency_breakdown}.}

\subsection{Defense}
\label{exp_setting_defense}

\paragraph{Basics.} We follow the defending strategy presented in \Cref{sec:defend} by blocking a list of token embeddings for the entire CLIP vocabulary. SubDefense is plugged into UCE, ESD, FMN, and SPM. Defense performance is mainly assessed on concepts ``nudity'', ``Van Gogh'', and ``church'' using our constructed dataset. \rebuttal{RECE is compared with UCE+SubDefense at rank 20 under the same controlled setting; all broader evaluations use a fixed rank 100. Rank is not selected per prompt. The choice follows the sensitivity study in \Cref{appsubsec:ablation_defense}: utility changes gradually through rank 100 and degrades more sharply thereafter.} Both before and after cleaning up the token embedding space, we conduct independent attacks following the setting in \Cref{exp_setting_attack}.

\paragraph{Metrics.} An effective defense strategy should reduce the attack success rate while preserving the generation quality of safe concepts. Hence, we use the following metrics. (\emph{i}) ASR. Various jailbreaking attacks are conducted before and after applying defenses, and the corresponding ASR is reported. Specifically for SubAttack, $K=5$ is used consistently before and after defense to ensure a fair comparison. (\emph{ii}) CLIP Score and FID are evaluated to test the generation quality of the defended model. MSCOCO \cite{coco} contains image and text caption pairs. Following \cite{zhang2024unlearndiff,zhang2024defensive}, we use 10k MSCOCO text captions to generate images before and after defense. Then, we report the mean CLIP score \cite{CLIPScore} of generated images with their corresponding text captions to test the defended models' ability to follow these harmless prompts. And we report the FID between generated images and original MSCOCO images to test the quality of generated images.


\section{Auxiliary Attack Results}
\label{appsec:aux_attack}
\subsection{More Interpretation Results on Attack Token Embeddings}
\label{app-subsec:interpret}

First of all, we show detailed results of transferring token embeddings from unlearned models to the original SD in \textbf{\Cref{tab:to_sd}}, emphasizing that these embeddings are inherited from the original SD.

\begin{table}[h!]
\centering
\caption{\textbf{Token embeddings learned by SubAttack originate from the original SD.} This is evidenced by the successful transfer of attack token embeddings from unlearned models to the original SD with high ASR.}
\resizebox{0.7\textwidth}{!}{%
\begin{tabular}{@{}l|cccc@{}}
\toprule
\midrule
\textbf{Scenarios:} & \textbf{ESD$\rightarrow$SD} & \textbf{FMN$\rightarrow$SD} & \textbf{UCE$\rightarrow$SD} & \textbf{SPM$\rightarrow$SD} \\ \midrule
Nudity           &  97.44\%   &  97.78\%   &  95.89\%   &  86.11\%   \\
Van Gogh         &  86\%   &  84\%   &  88.44\%   &  93.11\%   \\
Church           &  87.22\%   &  92.56\%   &  85.56\%   &   84.33\%  \\ 
\midrule
\bottomrule
\end{tabular}
}
\label{tab:to_sd}
\end{table}

\begin{figure}[t]
\centering
\begin{subfigure}[b]{0.19\textwidth}
         \centering
        \includegraphics[width=\linewidth]{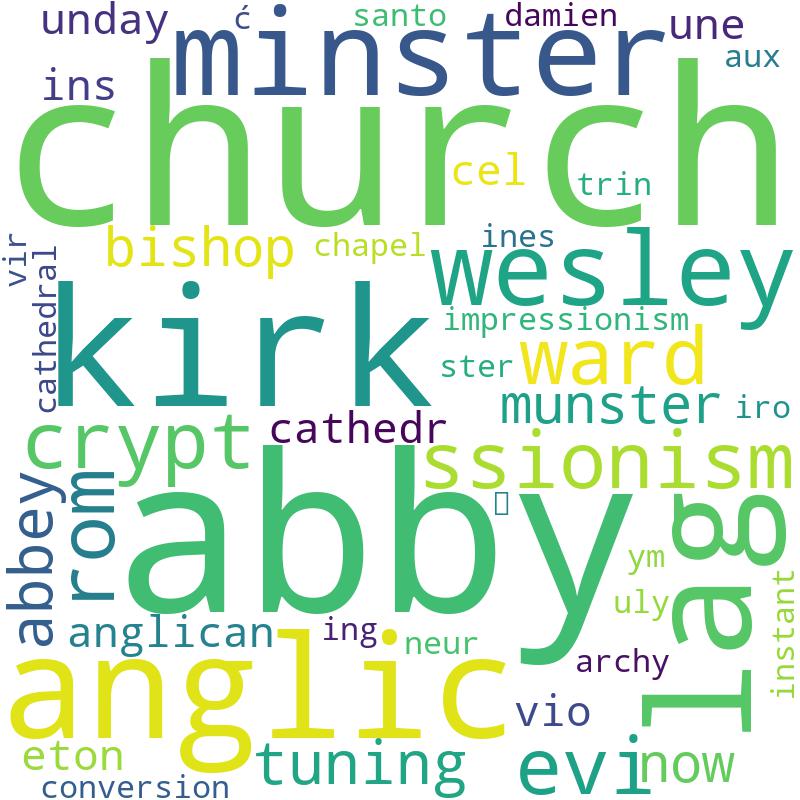}
        \caption{SD}
        \label{fig:sd_church}
     \end{subfigure}
\begin{subfigure}[b]{0.19\textwidth}
         \centering
        \includegraphics[width=\linewidth]{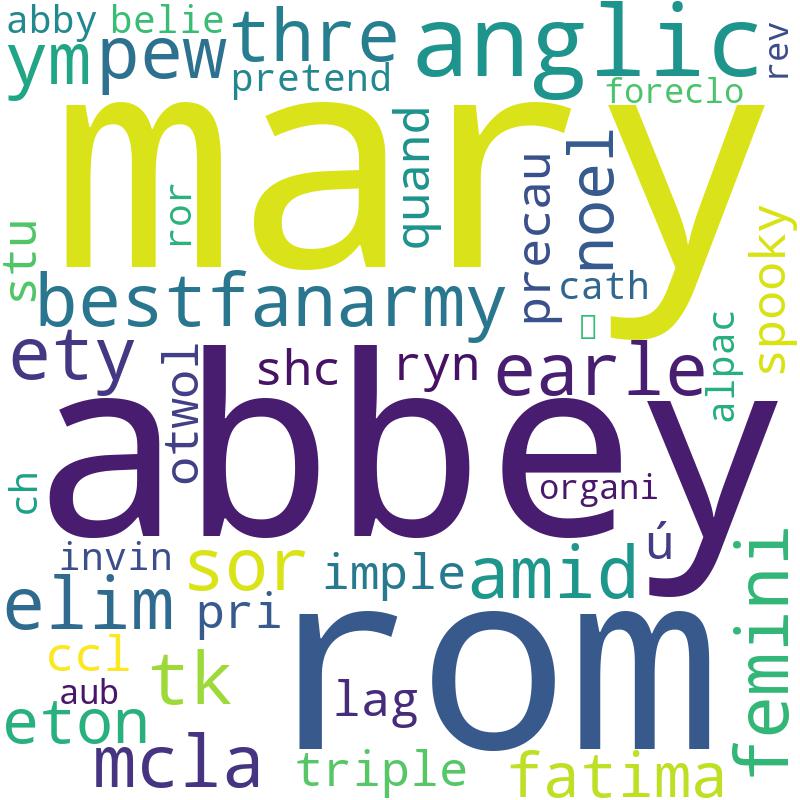}
        \caption{ESD}
        \label{fig:esd_church}
     \end{subfigure}
\begin{subfigure}[b]{0.19\textwidth}
         \centering
        \includegraphics[width=\linewidth]{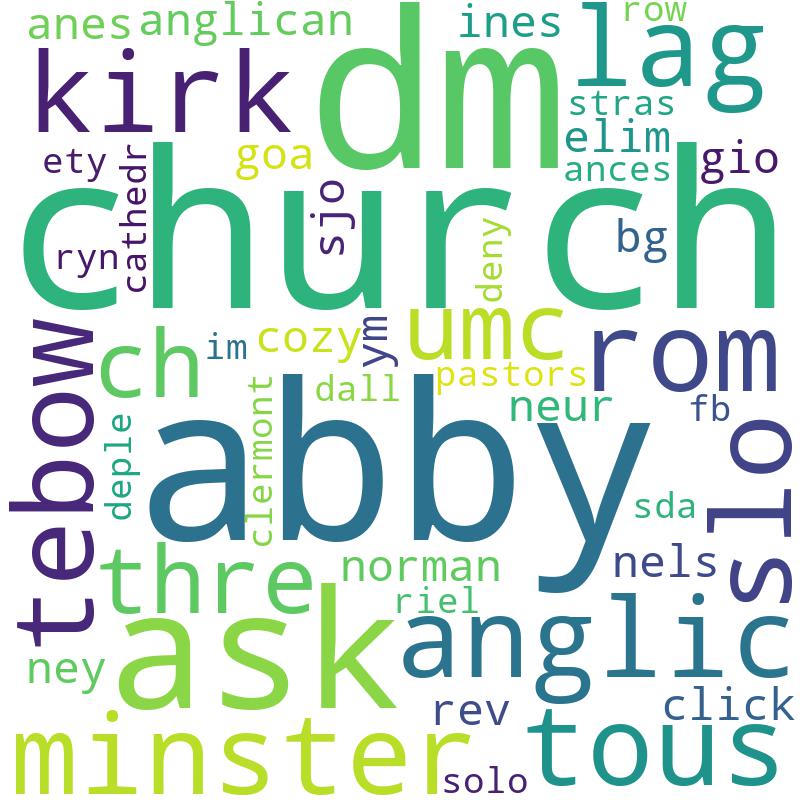}
        \caption{FMN}
        \label{fig:fmn_church}
     \end{subfigure}
\begin{subfigure}[b]{0.19\textwidth}
         \centering
        \includegraphics[width=\linewidth]{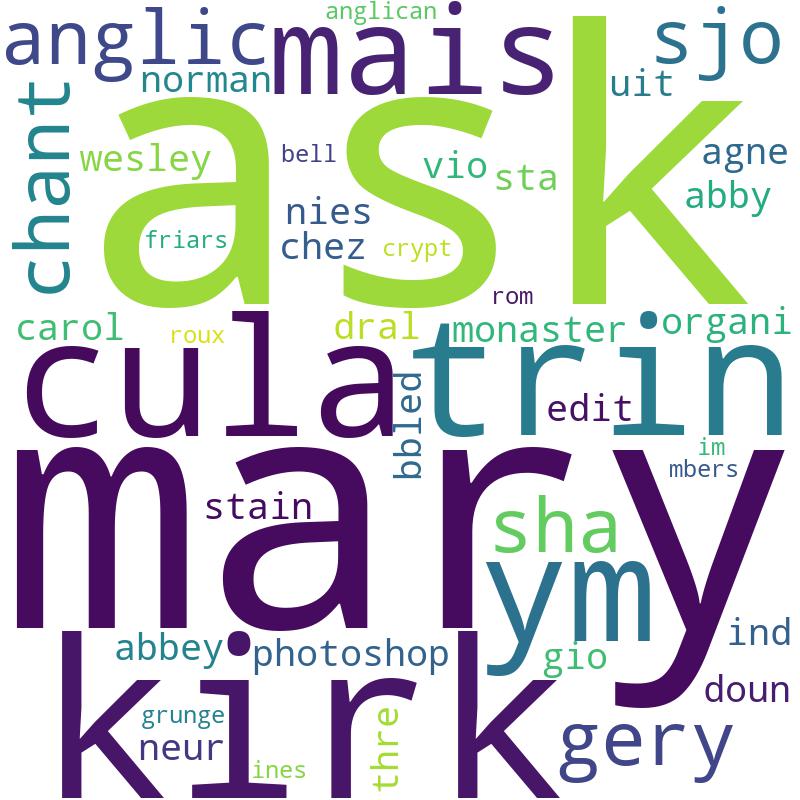}
        \caption{UCE}
        \label{fig:uce_church}
     \end{subfigure}
\begin{subfigure}[b]{0.19\textwidth}
         \centering
        \includegraphics[width=\linewidth]{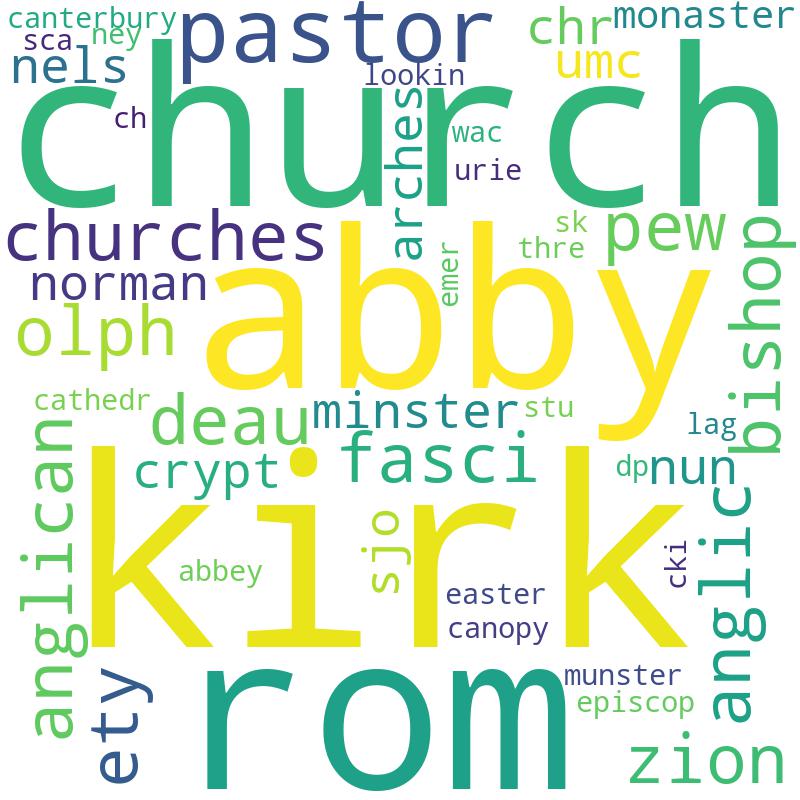}
        \caption{SPM}
        \label{fig:spm_church}
     \end{subfigure}
\caption{Interpreting attack token embeddings for the concept ``church''.}
\label{fig:app_interpret_church}
\end{figure}

\begin{figure}[t]
\centering
\begin{subfigure}[b]{0.19\textwidth}
         \centering
        \includegraphics[width=\linewidth]{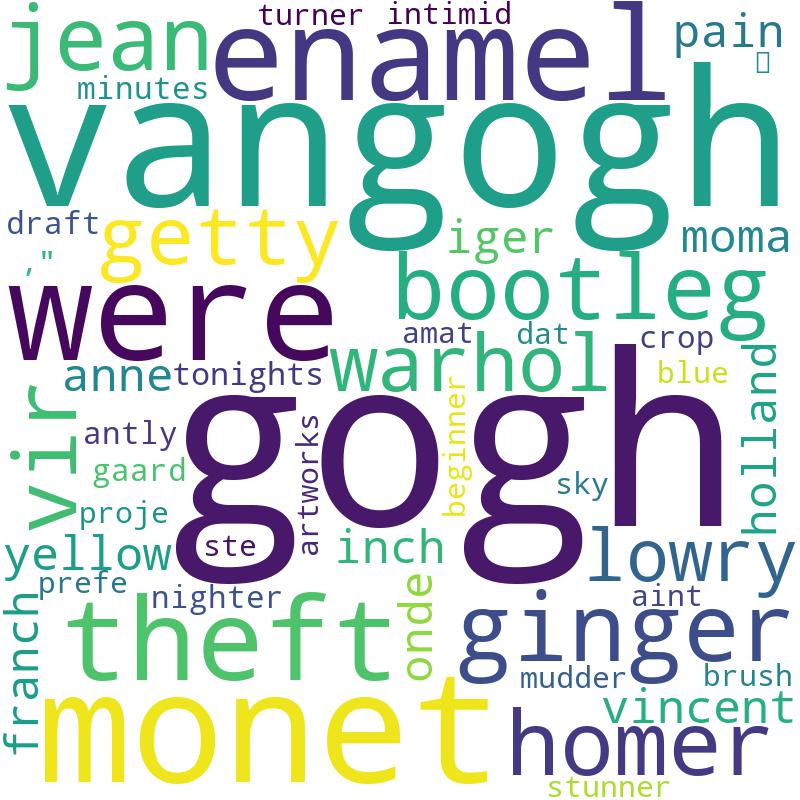}
        \caption{SD}
        \label{fig:sd_van}
     \end{subfigure}
\begin{subfigure}[b]{0.19\textwidth}
         \centering
        \includegraphics[width=\linewidth]{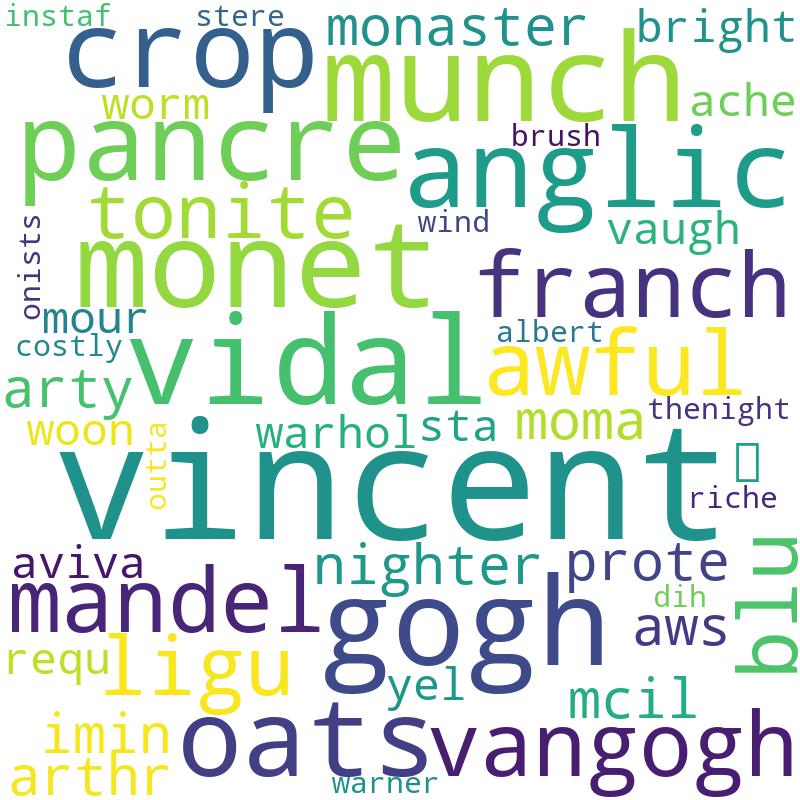}
        \caption{ESD}
        \label{fig:esd_van}
     \end{subfigure}
\begin{subfigure}[b]{0.19\textwidth}
         \centering
        \includegraphics[width=\linewidth]{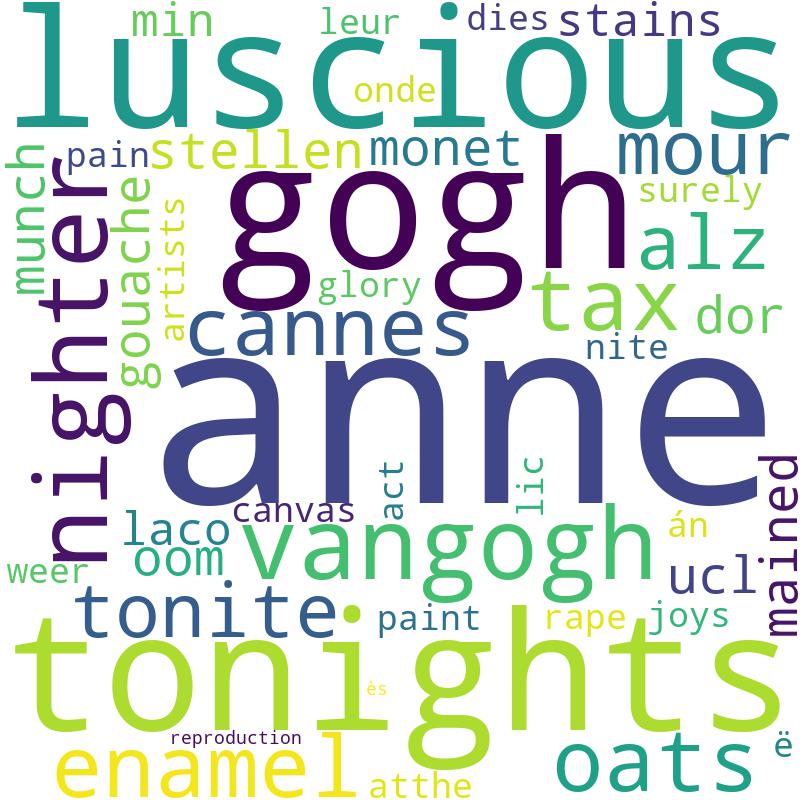}
        \caption{FMN}
        \label{fmn_van}
     \end{subfigure}
\begin{subfigure}[b]{0.19\textwidth}
         \centering
        \includegraphics[width=\linewidth]{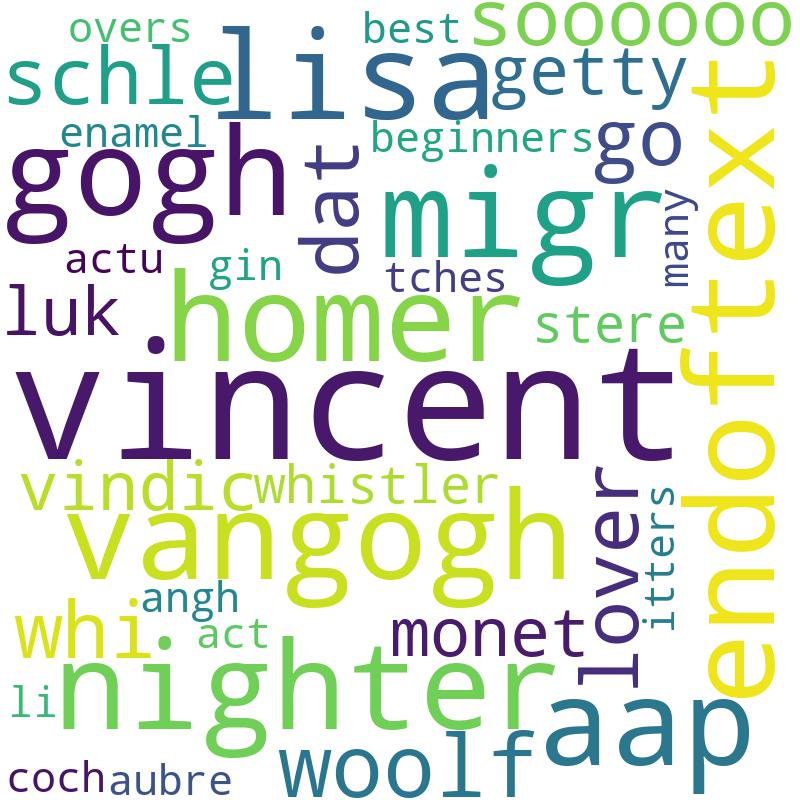}
        \caption{UCE}
        \label{fig:uce_van}
     \end{subfigure}
\begin{subfigure}[b]{0.19\textwidth}
         \centering
        \includegraphics[width=\linewidth]{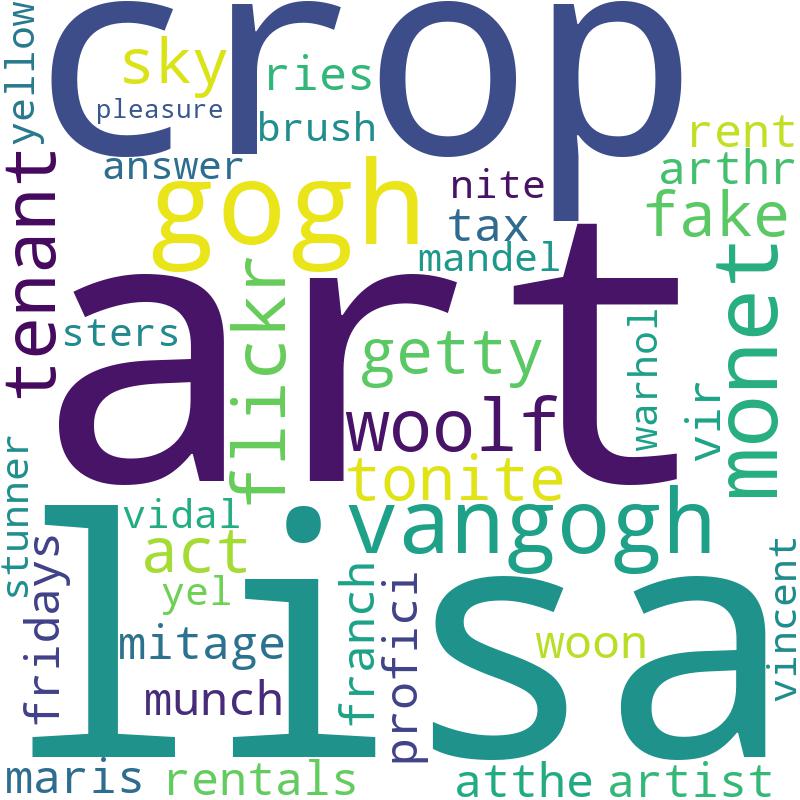}
        \caption{SPM}
        \label{fig:spm_van}
     \end{subfigure}
\caption{Interpreting attack token embeddings for the concept ``Van Gogh''.}
\label{fig:app_interpret_vangogh}
\end{figure}

Moreover, we should provide additional interpretation of the sets of learned attack token embeddings for ``church'' and ``Van Gogh'' across different unlearned LDMs in \textbf{\Cref{fig:app_interpret_church}} and \textbf{\Cref{fig:app_interpret_vangogh}}, showing observations on \textbf{interpretable associations} similar to that of ``nudity''. 

For example, for ``church'', ESD (\Cref{fig:esd_church}) and UCE (\Cref{fig:uce_church}) majorly relate it with \textbf{religious concepts}, including names (``mary''), places (``abbey'', ``abby'', ``rom'' for ``rome''), etc. Interestingly, in \textbf{Scotland and Northern England English}, "kirk" is the traditional word for ``church'' - this may be integrated into LDM during the training of large-scale datasets, but not removed during existing diffusion unlearning methods. As for FMN (\Cref{fig:fmn_church}) and SPM (\Cref{fig:spm_church}), the \textbf{explicit concept ``church''} itself is a significant component. Notably, FMN and SPM also exhibit higher ASR with no attack as presented in \Cref{fig:subattack} and \Cref{tab:attack_result_app}. Under NoAttack, both of them achieve ASR greater than 40\%, but ASR for ESD and UCE is less than 10\%. This also emphasizes that explicit associations also remain in some unlearned LDMs.

As for the concept ``Van Gogh'', when interpreting the sets of embeddings collectively, more \textbf{explicit words} are exposed for existing unlearned models such as ``vincent'', ``gogh'', ``vangogh'', along with \textbf{implicit words} ``art'', ``artist'', ``munch'' (Edvard Munch is an impressionist sharing similar themes and styles with Van Gogh, and the Van Gogh Museum in Amsterdam and the Munch Museum have collaborated to give a joint exhibition, "Munch: Van Gogh".) ``monet'' (also an impressionist), ``nighter'' and ``oats'' (concepts commonly in Van Gogh's paintings), etc. Although UCE, which shows the highest ASR with no attack, has the largest amount of explicitly associated concepts, other unlearned models all show explicit words more or less. This suggests that current unlearning methods retain more explicit associations with the target concept when applied to styles, compared to their application to NSFW and object concepts.

\subsection{More ASR Results}
\label{appsubsec:more_asr}

We present SubAttack ASR details with K=5 on different models across six concepts in \textbf{\Cref{tab:attack_result_app}}. 
Moreover, we show ASR on a broader range of unlearned LDMs and settings such as massive concepts in \textbf{\Cref{tab:mace_extended}}, \textbf{\Cref{tab:defense_salun_erasediff}}, and \textbf{\Cref{tab:defense_rece}}.
Further more, we show transfer attack performance details from ESD to other unlearned models using different attack methods across the three concepts (nudity, Van Gogh, church) in \textbf{\Cref{tab:compare_transfer}}. Moreover, we present additional transfer results between other unlearned model pairs using SubAttack with K=5 in \textbf{\Cref{tab:more_transfer}}.

\begin{table}[h!]
\centering
\caption{\textbf{Attack success rates (ASR)} targeting different unlearned diffusion models across different concept unlearning tasks (NSFW, artist style, object).}
\resizebox{\textwidth}{!}{%
\begin{tabular}{@{}l| cccc | cccc@{}}
\toprule
\midrule
 \textbf{Attacks:} 
 &   \multicolumn{4}{|c|}{NoAttack}
 & \multicolumn{4}{|c}{Ours}  
 \\ \midrule

\textbf{Victim Model:}  
& ESD  & FMN   & UCE & SPM 
& ESD  & FMN  & UCE & SPM \\ \midrule  

Nudity  
&  18.78\%  &  90\%  &  23\%   &  22.56\%    
&  97.56\%  &  100.00\%  &  81.67\%   &  74.89\%  
\\

Van Gogh   
&  5.78\%  &  21.56\%  &   71.44\%  &  43.78\%  
&   81\% &  96.33\%  &  98.33\%   &  82.78\%  
\\

Church 
&  9.33\%  & 51.56\%   &  6.55\%   &  43.78\%  
&  91.33\%  &  97.78\%  &  82.67\%   & 84.89\%   
\\

Garbage Truck 
&  4\%  &  41.33\%   &  11.33\%   &  12.67\%  
&  31.33\%  &  91.67\%  &  44\%   & 77.67\%   
\\

Parachute 
&  4\%  & 63.67\%   &  1.3\%   &  30.67\%  
&  88.67\%  & 100\%  &  67\%   & 97\%   
\\

Tench 
&  1.67\%  & 40\%   &  0\%  &  14.33\%  
&  26.67\%  &  80\%  &  49\%   & 84.33\%   
\\
\midrule
\bottomrule
\end{tabular}
}
\label{tab:attack_result_app}
\end{table}

\begin{table}[h!]
\centering
\caption{\textbf{Evaluation across diverse concepts and settings including MACE, SA, and AC.}}
\resizebox{\textwidth}{!}{%
\begin{tabular}{@{}l|c|c|c|c|c|c@{}}
\toprule
\midrule
\textbf{Scenarios:} & \textbf{MACE (Nudity)} & \textbf{MACE (Truck)} & \textbf{MACE (Airplane)} & \textbf{MACE (Ship)} & \textbf{SA (Nudity)} & \textbf{AC (Van Gogh)} \\
\midrule
NoAttack & 6.67\% & 10\% & 0\% & 6.67\% & 83.33\% & 21.67\% \\
\rowcolor{gray!10}
SubAttack (Ours) & \textbf{98.33\%} & \textbf{85.56\%} & \textbf{96.67\%} & \textbf{100\%} & \textbf{98.33\%} & \textbf{61.67\%} \\
\midrule
\bottomrule
\end{tabular}
}
\label{tab:mace_extended}
\end{table}

\begin{table}[h!]
\centering
\caption{\textbf{Attack success rates (ASR) against additional unlearned models including SalUn and EraseDiff.}}
\resizebox{0.75\textwidth}{!}{%
\begin{tabular}{@{}l|c|c|c|c@{}}
\toprule
\midrule
\textbf{Scenarios:} & \textbf{Church} & \textbf{Garbage Truck} & \textbf{Parachute} & \textbf{Tench} \\
\midrule
SalUn (NoAttack) & 1.67\% & 5\% & 5\% & 0\% \\
\rowcolor{gray!10}
SalUn (SubAttack) & \textbf{56.67\%} & \textbf{40\%} & \textbf{86.67\%} & \textbf{11.67\%} \\
\midrule
EraseDiff (NoAttack) & 6.67\% & 6.67\% & 3.33\% & 0\% \\
\rowcolor{gray!10}
EraseDiff (SubAttack) & \textbf{31.67\%} & \textbf{38.33\%} & \textbf{78.33\%} & \textbf{15\%} \\
\midrule
\bottomrule
\end{tabular}
}
\label{tab:defense_salun_erasediff}
\end{table}

\begin{table}[h!]
\centering
\caption{\textbf{Attack success rates (ASR) against RECE.}}
\resizebox{0.45\textwidth}{!}{%
\begin{tabular}{@{}l|c|c|c@{}}
\toprule
\midrule
\textbf{Scenarios:} & \textbf{Nudity} & \textbf{Van Gogh} & \textbf{Church} \\
\midrule
NoAttack & 3.33\% & 16.67\% & 3.33\% \\
\rowcolor{gray!10}
SubAttack & \textbf{62.44\%} & \textbf{84.44\%} & \textbf{80.33\%} \\
\midrule
\bottomrule
\end{tabular}
}
\label{tab:defense_rece}
\end{table}

\begin{table}[h!]
\centering
\caption{\textbf{Transfer attack success rate using different attack methods}, transferring from ESD to FMN, UCE, and SPM across the three concepts. SubAttack (Ours) transfers best in almost all settings.}
\resizebox{\textwidth}{!}{%
\begin{tabular}{@{}l|ccc|ccc|ccc@{}}
\toprule
\midrule
& \multicolumn{3}{c|}{\textbf{Nudity}} & \multicolumn{3}{c|}{\textbf{Van Gogh}} & \multicolumn{3}{c}{\textbf{Church}} \\
\textbf{Scenario:} & E$\rightarrow$F & E$\rightarrow$U & E$\rightarrow$S & E$\rightarrow$F & E$\rightarrow$U & E$\rightarrow$S & E$\rightarrow$F & E$\rightarrow$U & E$\rightarrow$S \\
\midrule
NoAttack    & 90\%    & 23\%    & 22.56\% & 21.56\% & 71.44\% & 43.78\% & 51.56\% & 6.55\%  & 43.78\% \\
UnlearnDiff & 93.33\% & 41.33\% & 38.22\% & 12.78\% & 64\%    & 47.11\% & 6.19\%  & 13.33\% & 58\%    \\
CCE         & 93\%    & 18.33\% & 37.56\% & 72.33\% & 43.56\% & 81.33\% & 91\%    & 70.11\% & \textbf{92.78\%} \\
\rowcolor{gray!10}
SubAttack (Ours) & \textbf{96.89\%} & \textbf{77\%} & \textbf{80.44\%} & \textbf{72.67\%} & \textbf{88.89\%} & \textbf{86.89\%} & \textbf{92.89\%} & \textbf{83.77\%} & 92\% \\
\midrule
\bottomrule
\end{tabular}
}
\label{tab:compare_transfer}
\end{table}

\noindent\emph{(Here E, F, U, S abbreviate ESD, FMN, UCE, and SPM, respectively.)}

\begin{table}[h!]
\centering
\caption{\textbf{More SubAttack transfer results across four model pairs.}}
\resizebox{0.8\textwidth}{!}{%
\begin{tabular}{@{}lllll@{}}
\toprule
\midrule
\textbf{Scenario:} & FMN->UCE & UCE->ESD & SPM->UCE & UCE->FMN \\ \midrule
Nudity              & 72\%  &  81.33\%   & 86.11\% & 93.44\% \\ 
Van Gogh              & 91.11\%  &  48.55\%   & 80.55\% & 62.55\%  \\ 
Church              & 79.33\%  &  42.44\%   & 68.33\% & 78.77\%  \\
\midrule
\bottomrule
\end{tabular}
}
\label{tab:more_transfer}
\end{table}

\section{Auxiliary Defense Results}
\label{appsec:aux_defense}

\subsection{Detailed Baseline Comparison of Defending UCE Against UnlearnDiff}


A more detailed comparison results of RECE and SubDefense together with UCE with no defense are presented in \textbf{\Cref{tab:defense_uce}} and \textbf{\Cref{tab:defense_uce_coco}}.

\begin{table}[h!]
\centering
\caption{\textbf{SubDefense is stronger than baseline RECE in defending three concepts on UCE against UnlearnDiff or our SubAttack.}}
\resizebox{\textwidth}{!}{%
\begin{tabular}{@{}l| ccc | ccc@{}}
\toprule
\midrule
 \textbf{Attacks:} 
 &   \multicolumn{3}{|c|}{UnlearnDiff}
 & \multicolumn{3}{|c}{SubAttack}  
 \\ \midrule

\textbf{Scenarios:}  
& UCE  & UCE + SubDefense   & RECE  
& UCE  & UCE + SubDefense  & RECE  \\ \midrule  

Nudity  
&  78.22\%  &  73.55\% (\textbf{-4.67\%})  &  76.44\% (-1.78\%)       
&  81.67\%  &  34.11\% (\textbf{-47.56\%})  &  62.44\% (-19.23\%)    
\\

Van Gogh   
&  100\%  &  52.78\% (\textbf{-47.22\%})  &   61.67\% (-38.33\%)   
&   98.33\% &  29.44\% (\textbf{-68.89\%})  &  84.44\% (-13.89\%)   
\\

Church 
&  61.67\%  & 39.78\% (\textbf{-64.34\%})   &  50.78\% (-10.89\%)   
&  82.67\%  &  5.22\% (\textbf{-77.45\%}) &  80.33\% (-2.34\%)   
\\
\midrule
\bottomrule
\end{tabular}
}
\label{tab:defense_uce}
\end{table}

\begin{table}[h!]
\centering
\caption{\textbf{SubDefense preserves better utility than baseline RECE after defense.}}
\resizebox{.7\textwidth}{!}{%
\begin{tabular}{@{}l| ccc | ccc@{}}
\toprule
\midrule
 \textbf{Metrics:} 
 &   \multicolumn{3}{|c|}{COCO-10k FID ($\downarrow$)}
 & \multicolumn{3}{|c}{COCO-10k CLIP ($\uparrow$)}  
 \\ \midrule

\textbf{Scenarios:}  
& UCE  & UCE + SubDefense   & RECE  
& UCE  & UCE + SubDefense  & RECE  \\ \midrule  

Nudity  
& \textbf{17.14} &  17.51  &  17.57         
&  \textbf{30.86}  &  30.70  &  30.07    
\\

Van Gogh   
&  \textbf{16.64}  & \textbf{16.64}  &  17.11    
&  \textbf{31.14}  &  30.94  & 30.08    
\\

Church 
&  17.84  &  \textbf{17.41}  &  \textbf{17.41}   
&  \textbf{30.95}  & 30.86  &  30.07   
\\

\midrule
\bottomrule
\end{tabular}
}
\label{tab:defense_uce_coco}
\end{table}

\subsection{Defending Against UnlearnDiff on the I2P Dataset for Various Unlearned Models}
\label{appsubsec:defense_i2p}

We construct dataset for concepts belonging to the style and object class following UnlearnDiff but with a larger size. Hence, defending against UnlearnDiff using these datasets can demonstrate the effectiveness of SubDefense in a scenario consistent with UnlearnDiff. However, for NSFW concepts such as nudity, UnlearnDiff filters prompts and seeds from the I2P dataset. Hence, to further test SubDefense's ability in defending against UnlearnDiff in this specific setting, we conduct UnlearnDiff with or without SubDefense using the I2P dataset as well. We report the defense results on ESD, FMN, UCE, and SPM in \textbf{\Cref{tab:i2p_defense}}. We can see that SubDefense (with 100 blocked tokens) consistently reduces ASR on I2P-nudity for all four models, both without attack and under UnlearnDiff.

\begin{table}[h!]
\centering
\caption{\textbf{SubDefense for I2P-nudity against UnlearnDiff}, with 100 blocked tokens. For each unlearned model we report ASR before / after applying SubDefense (absolute change in parentheses).}
\resizebox{\textwidth}{!}{%
\begin{tabular}{@{}l|cc|cc|cc|cc@{}}
\toprule
\midrule
& \multicolumn{2}{c|}{\textbf{ESD}} & \multicolumn{2}{c|}{\textbf{FMN}} & \multicolumn{2}{c|}{\textbf{UCE}} & \multicolumn{2}{c}{\textbf{SPM}} \\
\textbf{Scenario:} & Base & +SubDefense & Base & +SubDefense & Base & +SubDefense & Base & +SubDefense \\
\midrule
NoAttack     & 20.56\% & 9.93\% (-10.63\%)  & 87.94\% & 37.59\% (-50.35\%) & 21.98\% & 13.47\% (-8.51\%)  & 55.31\% & 34.04\% (-21.27\%) \\
UnlearnDiff  & 74.47\% & 41.13\% (-33.34\%) & 97.87\% & 45.39\% (-52.58\%) & 78.72\% & 45.39\% (-33.33\%) & 91.49\% & 58.97\% (-32.52\%) \\
\midrule
\bottomrule
\end{tabular}
}
\label{tab:i2p_defense}
\end{table}

\color{black}
\subsection{Reproducibility of the recovered residual}
\label{appsubsec:stability}
We test whether SubAttack recovers consistent structure across target samples and optimization seeds. We form three deterministic 50/50 partitions of 100 ESD--nudity target images and run three optimization seeds on each disjoint subset, yielding 18 independent rank-100 bases (1,800 learned directions). For orthonormal bases $U,V\in\mathbb{R}^{d\times r}$, we compute principal angles through the singular values of $U^\top V$ and summarize them by normalized projection overlap $\|U^\top V\|_F^2/r$, equivalently the mean squared canonical correlation. Its expectation for independent random rank-$r$ subspaces is $r/d$.

\begin{table}[h!]
\centering
\caption{\textbf{Rank-100 residual-subspace stability.} Mean $\pm$ sample SD across matched pairs. Independent runs recover substantially more common structure than random rank-100 subspaces.}
\begin{tabular}{@{}l|c|c@{}}
\toprule
\textbf{Comparison} & \textbf{Pairs} & \textbf{Projection overlap} \\
\midrule
Disjoint subsets, same optimization seed & 9 & $0.3122\pm0.0038$ \\
Same subset, different optimization seeds & 18 & $0.3202\pm0.0036$ \\
Disjoint subsets, different optimization seeds & 18 & $0.3118\pm0.0037$ \\
Independent random rank-100 subspaces & -- & $0.1302$ (analytic expectation) \\
\bottomrule
\end{tabular}
\label{tab:residual_stability}
\end{table}
The disjoint-subset overlap is close to the same-subset/different-seed overlap, indicating that sample variation is comparable to optimization variation rather than producing unrelated bases. This supports a reproducible attack-accessible linear residual, not a unique or complete representation of the concept.

We also evaluate directions 1--30 from every run on the same 30 prompts and two fixed diffusion seeds (60 examples). Best-of-prefix ASR counts an example as successful if any of the first $K$ directions succeeds.
\begin{table}[h!]
\centering
\caption{\textbf{Functional stability across 18 independent SubAttack runs.} Mean $\pm$ SD across learned runs on the fixed 60-example protocol.}
\begin{tabular}{@{}c|ccccc@{}}
\toprule
First $K$ directions & 1 & 5 & 10 & 20 & 30 \\
\midrule
Best-of-prefix ASR & $61.7\pm7.8$ & $78.0\pm8.6$ & $81.9\pm8.9$ & $86.6\pm6.7$ & $88.7\pm5.9$ \\
\bottomrule
\end{tabular}
\label{tab:functional_stability}
\end{table}
The unattacked ESD baseline is 15.0\%. Attack coverage increases and run-to-run variation decreases as directions accumulate.
\color{black}

\color{black}
\subsection{Nearby benign concepts and defense rank}
\label{appsubsec:nearby_painters}
Aggregate COCO metrics may hide local semantic side effects. We therefore evaluate ESD--Van-Gogh before and after rank-20 and rank-50 SubDefense on seven non-target painters, generic painting prompts, and Van Gogh as a target control. Each arm uses the same 30 contents and two fixed seeds (540 images), enabling paired comparison. \Cref{tab:nearby_painters} reports changes in prompt--image CLIP alignment relative to the undefended model; values closer to zero indicate better preservation.

\begin{table}[h!]
\centering
\caption{\textbf{Nearby-painter preservation under Van-Gogh SubDefense.} Paired change in CLIP alignment relative to the undefended model.}
\begin{tabular}{@{}l|rr@{}}
\toprule
\textbf{Prompt group} & \textbf{Rank 20} & \textbf{Rank 50} \\
\midrule
C\'ezanne & $+0.415$ & $+0.154$ \\
Monet & $-0.492$ & $-0.385$ \\
Picasso & $-0.435$ & $-0.541$ \\
Rembrandt & $-0.307$ & $+0.257$ \\
Generic painting & $-0.023$ & $-0.120$ \\
Van Gogh (target control) & $-0.801$ & $-1.193$ \\
\bottomrule
\end{tabular}
\label{tab:nearby_painters}
\end{table}
Broad painterly alignment is nearly unchanged at ranks 20--50, while effects on individual nearby painters are non-uniform. We therefore view defense rank as a local robustness--utility trade-off rather than claim uniform preservation of every semantically related concept.
\color{black}

\color{black}
\subsection{Variance across independently learned defenses}
\label{appsubsec:defense_variance}
To measure variability in the defense intervention, we independently construct three rank-100 ESD--nudity SubDefense bases and evaluate each on the same 60 prompt--seed pairs.
\begin{table}[h!]
\centering
\caption{\textbf{SubDefense variance across independently learned rank-100 bases.}}
\begin{tabular}{@{}c|cc@{}}
\toprule
\textbf{Optimization seed} & \textbf{Direct defended ASR} & \textbf{CLIP alignment} \\
\midrule
1024 & 6.67\% & 28.362 \\
2026 & 0.00\% & 28.401 \\
4096 & 3.33\% & 28.897 \\
\midrule
Mean $\pm$ SD & $3.33\pm3.33$\% & $28.553\pm0.298$ \\
\bottomrule
\end{tabular}
\label{tab:defense_variance}
\end{table}
This experiment measures reproducibility of the projected defense under a fixed direct evaluation protocol; it is distinct from the independently re-optimized adaptive attacks reported in the main defense results.
\color{black}

\subsection{Defending Against SubAttack on Various Concepts for Various Unlearned Models}
\label{appsubsec:defense_subattack}

Apart from the major baseline comparison of defense on UCE, and the defense results against different attacks on ESD presented in the main paper, we provide additional defense results of various concepts and unlearned models against SubAttack in this section. The results are shown in \textbf{\Cref{tab:more_subattack_defense}}. Notice that ASR on various concepts is reduced with SubDefense, while ASR reduction on ``Van Gogh'' is the most significant. It is worth exploring in the future to design new methods and make the defense more robust for other concepts as well.

\begin{table}[h!]
\centering
\caption{\textbf{SubDefense for three concepts against SubAttack}, with 100 blocked tokens. For each unlearned model we report ASR before / after applying SubDefense (absolute change in parentheses).}
\resizebox{\textwidth}{!}{%
\begin{tabular}{@{}l|cc|cc|cc|cc@{}}
\toprule
\midrule
& \multicolumn{2}{c|}{\textbf{ESD}} & \multicolumn{2}{c|}{\textbf{FMN}} & \multicolumn{2}{c|}{\textbf{UCE}} & \multicolumn{2}{c}{\textbf{SPM}} \\
\textbf{Concept:} & Base & +SubDefense & Base & +SubDefense & Base & +SubDefense & Base & +SubDefense \\
\midrule
Nudity   & 97.56\% & 42.33\% (-55.23\%) & 100\%    & 62.89\% (-37.11\%) & 81.67\% & 28\% (-53.67\%)    & 74.89\% & 50.78\% (-24.11\%) \\
Van Gogh & 81\%    & 17\% (-64\%)       & 96.33\%  & 22.78\% (-73.55\%) & 93.78\% & 14.33\% (-79.45\%) & 82.78\% & 12.33\% (-70.45\%) \\
Church   & 91.33\% & 40.22\% (-51.11\%) & 82.67\%  & 13.78\% (-68.89\%) & 82.67\% & 3.22\% (-79.45\%)  & 84.89\% & 23.78\% (-61.11\%) \\
\midrule
\bottomrule
\end{tabular}
}
\label{tab:more_subattack_defense}
\end{table}

\subsection{Defending Results on Other Exploratory Settings}
\label{appsubsec:defense_other}

\paragraph{Defending against black-box attack.}

We also conducted exploratory experiments on defending against Ring-A-Bell \citep{tsai2024ringabell}, a classic black-box attack. Our results in \textbf{\Cref{tab:blackbox_defense}} show that SubDefense reduces ASR across several unlearned models, including MACE, FMN, SPM, and ESD. These findings suggest that SubDefense can provide robustness in black-box scenarios, although our main focus remains on white-box settings.

\begin{table}[h!]
\centering
\caption{\textbf{Exploratory defense results against the black-box Ring-A-Bell (Nudity) attack.}}
\resizebox{0.8\textwidth}{!}{%
\begin{tabular}{@{}l|c|c|c|c@{}}
\toprule
\midrule
\textbf{Scenarios:} & \textbf{MACE} & \textbf{FMN} & \textbf{SPM} & \textbf{ESD} \\
\midrule
Ring-A-Bell ASR & 11.58\% & 95.79\% & 34.74\% & 57.89\% \\
\rowcolor{gray!10}
+ SubDefense & \textbf{5.26\%} (k=10) & \textbf{54.75\%} (k=100) & \textbf{14.74\%} (k=100) & \textbf{4.21\%} (k=100) \\
\midrule
\bottomrule
\end{tabular}
}
\label{tab:blackbox_defense}
\end{table}

\paragraph{Standalone performance of SubDefense.}

Although SubDefense was primarily designed as a plug-in defense to enhance the robustness of existing unlearned models (similar to RECE operating on UCE), we also explored its effectiveness as a standalone unlearning method. Specifically, we applied SubDefense directly to the original Stable Diffusion (SD) model without any prior unlearning. Results are promising: as shown in \textbf{\Cref{tab:subdefense_standalone}}, SubDefense reduces ASR under both black-box (Ring-A-Bell, Nudity) and white-box (SubAttack, Church) attacks.

\begin{table}[h!]
\centering
\caption{\textbf{Standalone performance of SubDefense.}}
\resizebox{.9\textwidth}{!}{%
\begin{tabular}{@{}l|c|c|c|c|c|c|c@{}}
\toprule
\midrule
\textbf{Scenarios:} & \textbf{SD} & \textbf{K=10} & \textbf{K=20} & \textbf{K=50} & \textbf{K=100} & \textbf{K=150} & \textbf{K=200} \\
\midrule
Ring-A-Bell (Nudity) & 97.89\% & 89.47\% & 76.84\% & 60\% & 38.94\% & 23.16\% & 8.42\% \\
\rowcolor{gray!10}
SubAttack (Church)   & 100\% & 80\% & 78.33\% & 55\% & 46.67\% & 21.67\% & 10\% \\
\midrule
\bottomrule
\end{tabular}
}
\label{tab:subdefense_standalone}
\end{table}


\section{Ablations}
\label{appsec:ablations}

\color{black}
\subsection{Standardized attack-cost breakdown}
\label{appsubsec:efficiency_breakdown}
The average runtime in the main table combines two operationally different stages: learning attack tokens and applying them to test prompt--seed pairs. We therefore time ESD--nudity with matched one-GPU resource requests and separate these stages. CCE and SubAttack learn concept-level artifacts that are reused over all 900 evaluation pairs, whereas UnlearnDiff learns a new local prompt for every pair.

\begin{table}[h!]
\centering
\caption{\textbf{Stage-wise attack cost on ESD--nudity.} ``Reuse'' is the number of prompt--seed pairs over which a learned global artifact is amortized in the main comparison.}
\resizebox{0.95\textwidth}{!}{%
\begin{tabular}{@{}l|l|c|c|l@{}}
\toprule
\textbf{Method} & \textbf{Learned attack} & \textbf{Learning time} & \textbf{Reuse} & \textbf{Application per pair} \\
\midrule
CCE & One global token (500 steps) & 656 s/token & 900 & One diffusion generation \\
SubAttack & Five global directions (500 steps each) & $488\pm5$ s/direction; 2,551 s total & 900 & Five generations (one/direction) \\
UnlearnDiff & One local prompt per pair & 103--1,351 s/pair in the timing study & None & Included in per-pair optimization \\
\bottomrule
\end{tabular}}
\label{tab:efficiency_breakdown}
\end{table}
The five SubAttack directions take 495, 483, 485, 488, and 491 seconds. Their total learning cost contributes 2.83 s/pair when amortized over 900 pairs; CCE's contributes 0.73 s/pair. UnlearnDiff cannot amortize its optimization: among five completed local runs, two reach detector success after 85 and 384 seconds, while three use the full 1,324--1,351-second budget without success. This breakdown complements, rather than replaces, the end-to-end average in the main table.
\color{black}

\subsection{Attack}
\label{appsubsec:ablation_attack}

\begin{figure*}[h!]
    \centering
    \begin{subfigure}[b]{0.48\textwidth}
        \centering
        \includegraphics[width=\linewidth]{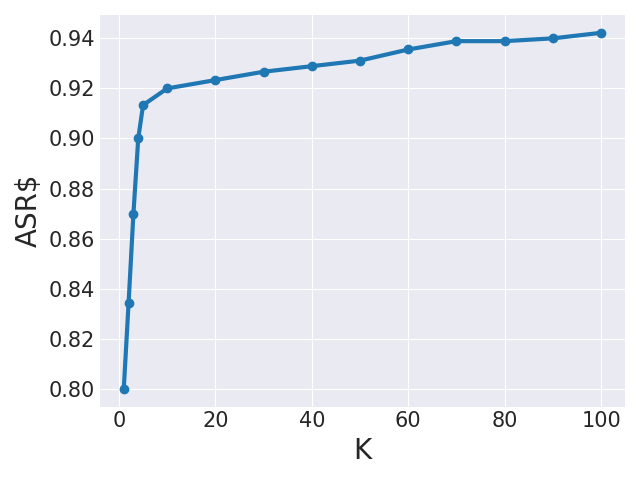}
        \caption{Concept ``church''.}
        \label{fig:asr_k_church}
    \end{subfigure}
    \hfill
    \begin{subfigure}[b]{0.48\textwidth}
        \centering
        \includegraphics[width=\linewidth]{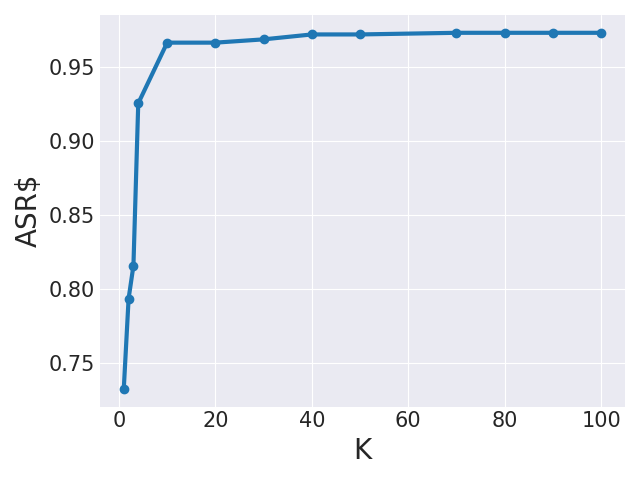}
        \caption{Concept ``nudity''.}
        \label{fig:asr_k_nude}
    \end{subfigure}
    \caption{\reviseq{\textbf{ASR versus $K$} when conducting SubAttack on ESD for the concepts ``church'' and ``nudity''.}}
    \label{fig:asr_k}
\end{figure*}

\paragraph{Number of attack tokens.}
In practice, we use $K = 5$ to conduct SubAttack as it provides strong attack performance while maintaining computational efficiency. Here, we take ESD as an example to show how ASR varies with $K$. To conduct ablations more efficiently, we subsample 300 out of 900 prompts for the concepts ``church'' and ``nudity'' to study the relationship between ASR and $K$. Results are presented in \textbf{\reviseq{\Cref{fig:asr_k}}}. The additional attack time per image caused by each additional token embedding is approximately 10 seconds, which leads to about 3 more hours to attack a single concept having 900 prompts in the dataset. Therefore, considering the needs of attacking multiple concepts and multiple models in practice, we choose $K=5$ where the ASR is approximately stabilized. For some unique scenarios, users can choose to increase $K$ for higher ASR at a cost of longer computation time.

\paragraph{Orthogonility.}
The orthogonality constraint was introduced to encourage diversity among the learned attack embeddings, preventing them from collapsing into a single semantic direction and thereby covering a broader and more effective attack space. To validate this design choice, we conducted an ablation study on the “Nudity” concept. As shown in \textbf{\Cref{tab:orthogonality_ablation}}, enforcing orthogonality consistently improves ASR across multiple unlearned models, supporting the effectiveness of this constraint.

\begin{table}[h!]
\centering
\caption{\textbf{Ablation on the orthogonality constraint.} Enforcing orthogonality improves ASR across unlearned models for the “Nudity” concept.}
\resizebox{0.5\textwidth}{!}{%
\begin{tabular}{@{}l|c|c|c|c@{}}
\toprule
\midrule
\textbf{Scenarios:} & \textbf{ESD} & \textbf{FMN} & \textbf{UCE} & \textbf{SPM} \\
\midrule
With Orthogonality    & \textbf{97.56\%} & \textbf{100\%} & \textbf{81.67\%} & \textbf{74.89\%} \\
\rowcolor{gray!10}
Without Orthogonality & 78.33\% & 100\% & 71.67\% & 63.33\% \\
\midrule
\bottomrule
\end{tabular}
}
\label{tab:orthogonality_ablation}
\end{table}

\paragraph{Vocabulary size.}

We ablate the vocabulary size used for SubAttack by selecting the top-N CLIP tokens most similar to the target concept. As shown in \textbf{\Cref{tab:vocab_ablation}}, ASR improves sharply up to 5000 tokens but declines when the vocabulary grows larger. This indicates that 5000 tokens strike the best balance between diversity and optimization feasibility.

\begin{table}[h!]
\centering
\caption{\textbf{Ablation on vocabulary size.} ASR of SubAttack on ESD “Nudity” with different vocabulary sizes.}
\resizebox{0.7\textwidth}{!}{%
\begin{tabular}{@{}l|c|c|c|c|c@{}}
\toprule
\midrule
\textbf{Vocabulary size} & \textbf{50} & \textbf{500} & \textbf{5000 (default)} & \textbf{10000} & \textbf{Full} \\
\midrule
ASR & 43.33\% & 81.67\% & \textbf{97.56\%} & 70\% & 28.33\% \\
\midrule
\bottomrule
\end{tabular}
}
\label{tab:vocab_ablation}
\end{table}

\subsection{Defense}
\label{appsubsec:ablation_defense}

\paragraph{Gradual degradation of generation utility with stronger defense.}
We show an ablation study on COCO-10k generation CLIP score and FID versus the number of blocked tokens in \textbf{\Cref{tab:clip_fid}} using ESD for the concept of ``nudity''. We can see that, after the number of blocked tokens surpasses 100, there appears to be a significant harm to the CLIP score and FID. In practice, the number of blocked tokens during defense can be selected to balance good generation quality and low ASR according to one's preference. In this paper, we provide an ablation study on ESD as an example, and report ASR majorly with 20 or 100 blocked tokens for different unlearned models and concepts.

\begin{table}[h!]
\centering
\caption{\textbf{SubDefense exhibits gradual degradation of CLIP score and FID when the number of blocked token embeddings increases.}}
\resizebox{0.7\textwidth}{!}{%
\begin{tabular}{@{}l|ccccccc@{}}
\toprule
\midrule
\textbf{\#Blocked Tokens:} & 0 & 20 & 50 & 100 & 200 & 300 & 350 \\ \midrule
CLIP Score ($\uparrow$)       &  30.13 & 30.02 & 29.86 &  29.58 &   28.54  & 26.15    &  24.72   \\
FID    ($\downarrow$)           & 18.23  & 19.02 & 19.09 &  19.20   & 20.92 &  26.42   &   30.33\\ 
\midrule
\bottomrule
\end{tabular}
}
\label{tab:clip_fid}
\end{table}

\paragraph{More results and discussions on defending against CCE.}
Defending against CCE is an underexplored problem in the field, where there are no baselines to compare with, to the best of our knowledge. Hence, we show a detailed study on defense against CCE, along with more discussions to support future research. As shown in \textbf{\Cref{tab:cce_defense_app}}, different from UlearnDiff, CCE requires a large number of tokens to be blocked if we aim to have low ASR. However, lower ASR achieved by more blocked attack tokens leads to a degradation of generation utility, with an increased FID and a decreased CLIP score, referring to \textbf{\Cref{tab:clip_fid}}. Such a phenomenon indicates that the embedding identified by CCE has a complex association with the target concept, sharing components with a variety of interpretable token embeddings found by our method. This suggests that fully understanding the behavior of CCE requires a deeper analysis of how LDMs interpret and generate concepts other than the current approach we use. For example, currently, the interpretability of retained associations of concepts relies on predefined CLIP vocabularies, which may not capture all implicit or nuanced representations retained in unlearned models. While the above question is beyond the scope of the current work, such insights could inform the development of more robust and versatile defense strategies in the future. With improved understanding of LDMs, future research may come up with more efficient and robust defenses against CCE while preserving model utility.

\begin{table}[h!]
\centering
\caption{\textbf{ASR of concept ``nudity'' on CCE after blocking different numbers of token embeddings.}}
\resizebox{0.9\textwidth}{!}{%
\begin{tabular}{@{}l|cccccccc@{}}
\toprule
\midrule
\textbf{\#Blocked Tokens:} & 0 & 100 & 230 & 270 & 320 & 350 & 390  \\ \midrule
CCE ASR              & 85.11\% & 75.67\% &  65.78\%   & 37.44\% &  28.11\%   &   18.11\% &   8.89\%  \\ 
\midrule
\bottomrule
\end{tabular}
}
\label{tab:cce_defense_app}
\end{table}

\section{Sparsity of Attack Token Embeddings}
\label{app-sec:sparse}

Sparsity constraints are widely adopted in prior concept decomposition works - where the linear combination coefficients $\alpha_i$ are forced to be nearly zeros except for dozens of tokens (usually 20-50). However, in our attacks, where the unlearned diffusion models majorly associate the target concept with a set of implicit tokens, removing such sparsity regularization is helpful, especially for attack token embeddings discovered later in the iterative learning process. Hence, we do not impose a sparsity constraint. Yet, it's interesting to find through our learning that a weaker sparse structure still emerges, and such sparsity gradually decreases as we learn more attack token embeddings through the iterative learning process. 

Specifically, for each learned attack token embedding, we normalize $\bm \alpha = [\alpha_1, \dots, \alpha_N]$ to have a unit norm. Then, we find the index $i^*$ such that:
\begin{equation}
    i^* = \argmin_i i, \;\;\text{such that} \;\; \sum_{j = 1}^i \alpha_j^2 \geq 0.9
\end{equation}

Besides, we also count the number of $\alpha_i$ such that $\alpha_i \geq 0.01$. We report the results of the first attack token embedding on ESD for each concept in \textbf{\Cref{tab:sparse}}. Notice the size of the CLIP token vocabulary is more than 40000.

\begin{table}[h!]
\centering
\caption{\textbf{Sparsity of the learned attack token embeddings.}}
\label{tab:sparse}
\resizebox{0.4\textwidth}{!}{%
\begin{tabular}{@{}l|ccc@{}}
\toprule
\midrule
\textbf{Concept:} & Nudity & Van Gogh & Church \\ \midrule
$i^*$              & 1455  &  668   & 547 \\ 
\#$\alpha_i \geq 0.01$              & 1743  &  1023   & 885  \\ 
\midrule
\bottomrule
\end{tabular}
}
\end{table}

During our iterative learning process of a set of tokens for the nudity concept, we observe a decreasing sparsity, as shown in \textbf{\Cref{tab:sparse_itr}}. This is intuitive since later attacking requires more complex associations to the target concept. 

\begin{table}[h!]
\centering
\caption{\textbf{Sparsity of the learned attack token embeddings decreases during the iterative subspace attack process.}}
\resizebox{0.9\textwidth}{!}{%
\begin{tabular}{@{}l|cccccccccc@{}}
\toprule
\midrule
\textbf{\#Itrs} & 1 & 10 & 30 & 50 & 70 & 100 & 130 & 150 & 170 & 200\\ \midrule
$i^*$              & 1455  &  1799  & 1905  & 1784 & 1914 & 2062 & 2062 & 2136 & 2155 & 2115\\ 
\#$\alpha_i \geq 0.01$              & 1743  &  2019  & 2078 & 2009 & 2206 & 2298 & 2328 & 2368 & 2358 & 2326\\ 
\midrule
\bottomrule
\end{tabular}
}
\label{tab:sparse_itr}
\end{table}

\begin{figure*}[h]
    \centering
    \includegraphics[width=\linewidth]{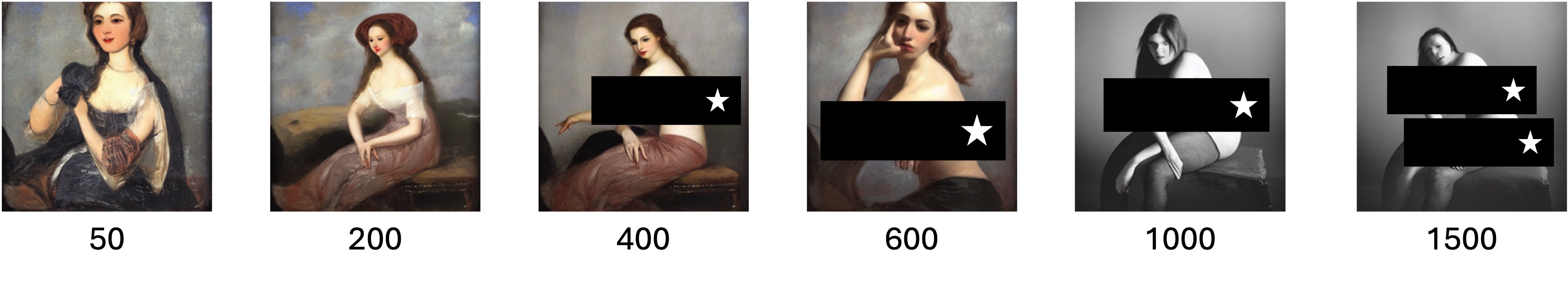}
    \caption{Attacking the concept nudity on ESD when $\bm \alpha$ has different numbers of non-zero entries.}
    \label{fig:app_sparse}
\end{figure*}

Furthermore, we visualize the nudity concept attacking results on ESD by selecting only the largest dozens of $\alpha_i$ within a learned $\bm \alpha$ and setting other entries as zeros. As shown in \textbf{\Cref{fig:app_sparse}}, we see the nudity concept is gradually enhanced as the number of selected $\alpha_i$ increases to 1500: the woman generated happens to wear fewer and fewer clothes until she's completely bare.

\clearpage

\section{Image Generation Quality Visualization After Defense}
\label{appsec:uce_vis}

\begin{figure*}[h!]
    \centering
    \includegraphics[width=.7\linewidth]{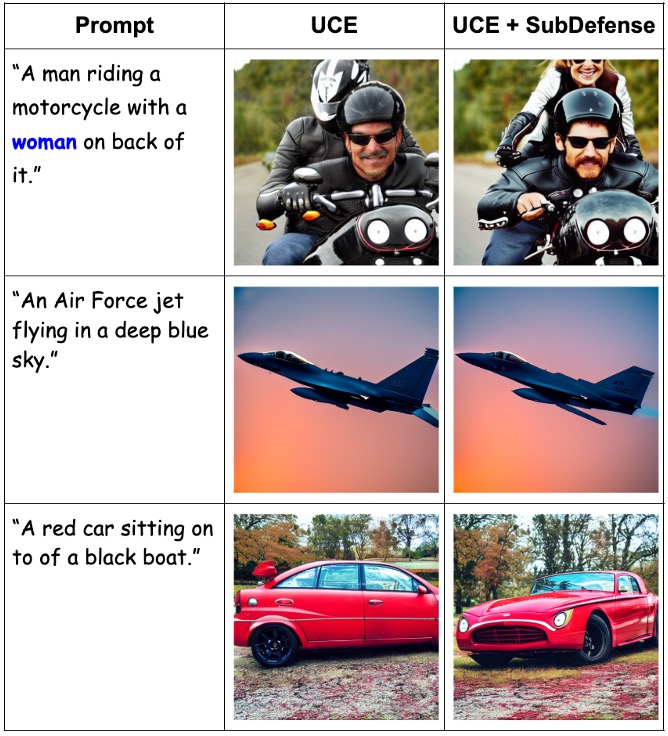}
    \caption{\textbf{More detailed visualization of COCO generation results with or without SubDefense on the concept nudity.}}
    \label{fig:more_uce_coco_nude}
\end{figure*}

\begin{figure*}[h!]
    \centering
    \includegraphics[width=.7\linewidth]{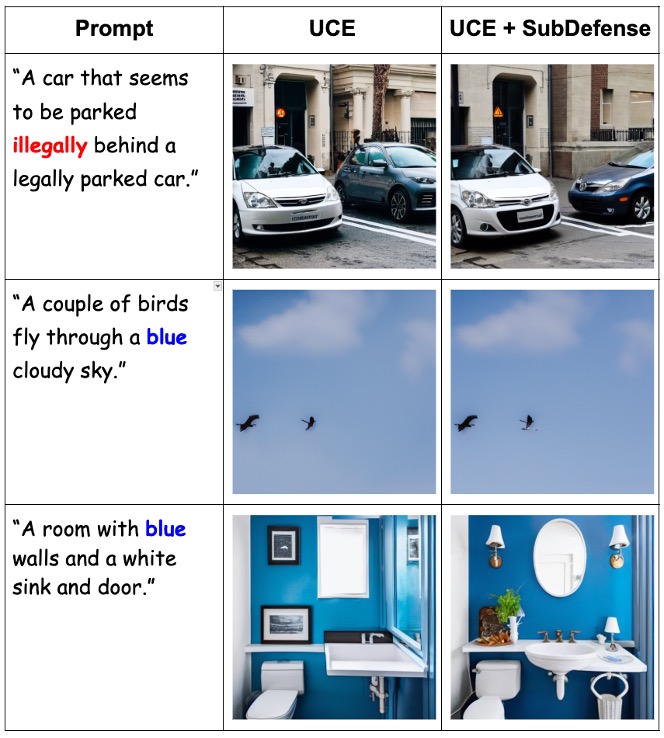}
    \caption{\textbf{More detailed visualization of COCO generation results with or without SubDefense on the concept Van Gogh.}}
    \label{fig:more_uce_coco_vangogh}
\end{figure*}

\begin{figure*}[h!]
    \centering
    \includegraphics[width=.7\linewidth]{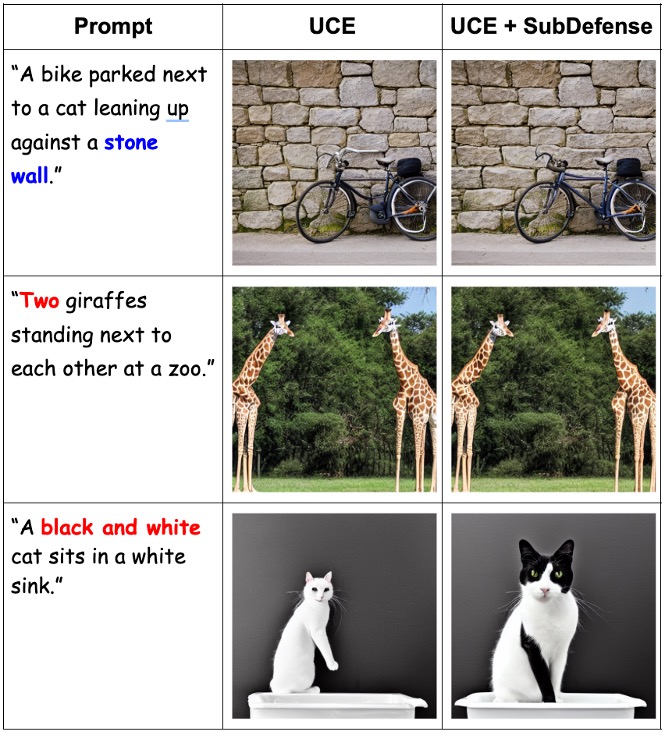}
    \caption{\textbf{More detailed visualization of COCO generation results with or without SubDefense on the concept church.}}
    \label{fig:more_uce_coco_church}
\end{figure*}

\begin{figure*}[h!]
    \centering
    \includegraphics[width=.7\linewidth]{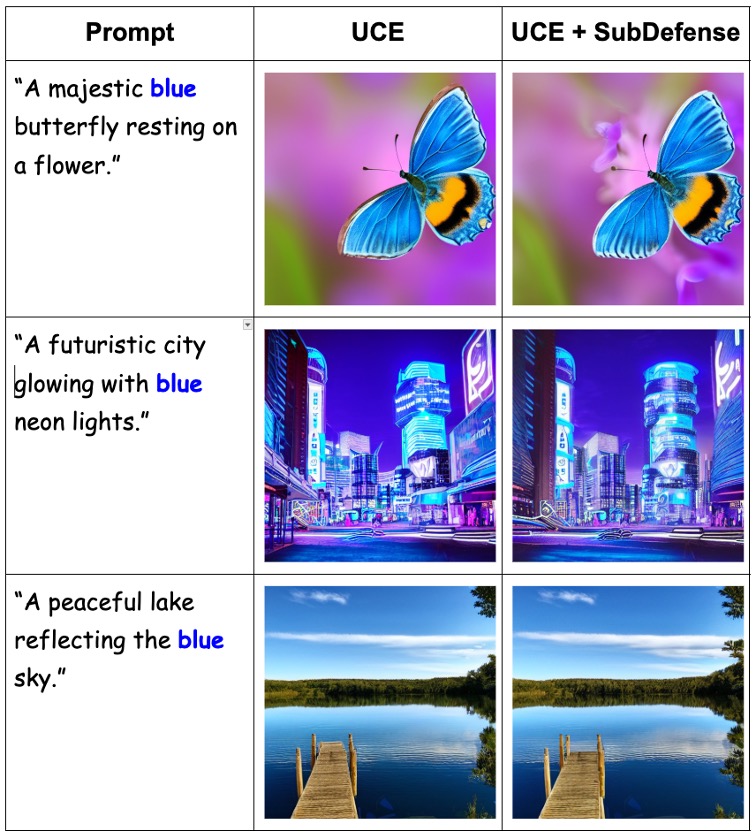}
    \caption{\textbf{Visualization of ``\textcolor{blue}{blue}'' image generation results before and after defending “Van Gogh” on UCE.}}
    \label{fig:uce_blue_vangogh}
\end{figure*}

\begin{figure*}[h!]
    \centering
    \includegraphics[width=.7\linewidth]{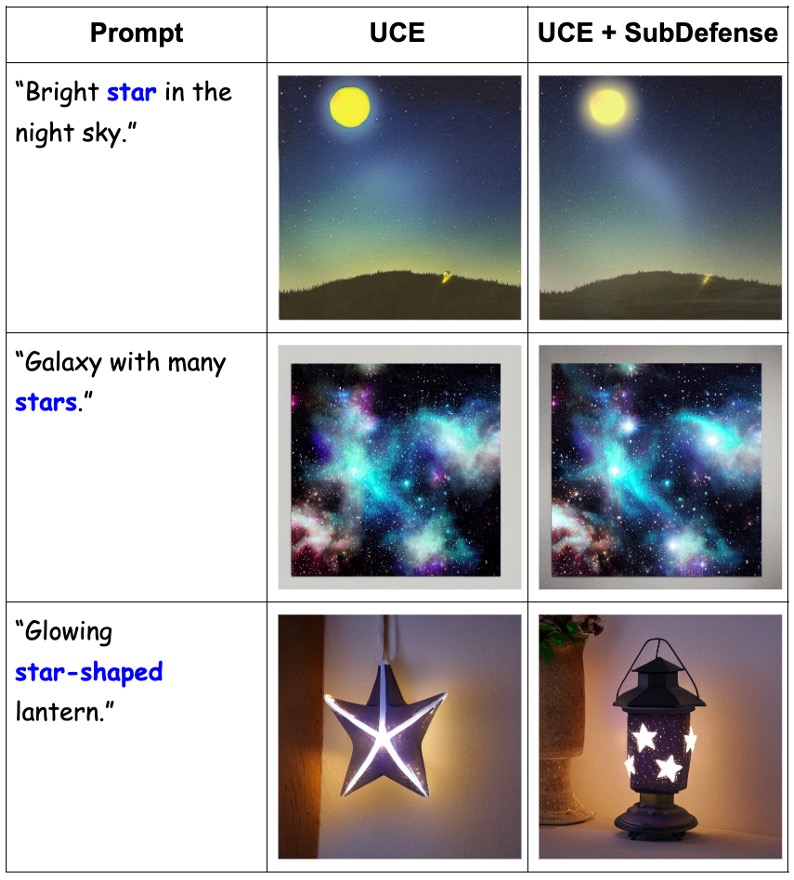}
    \caption{\textbf{Visualization of ``\textcolor{blue}{star}'' image generation results before and after defending “Van Gogh” on UCE.}}
    \label{fig:uce_star_vangogh}
\end{figure*}

In this section, we provide a more detailed study on the generation quality of unlearned models after we plug SubDefense into them. 
First, we provide more detailed MSCOCO prompts and the generated images of UCE and UCE + SubDefense (with 20 blocked tokens) in \textbf{\Cref{fig:more_uce_coco_nude}}, \textbf{\Cref{fig:more_uce_coco_vangogh}}, and \textbf{\Cref{fig:more_uce_coco_church}}. Next, taking UCE and ``Van Gogh'' as an example, whose attack token embeddings are highly related to ``blue'' and ``star'', we study whether SubDefense of ``Van Gogh'' harms the generation of ``blue'' and ``star'' in \textbf{\Cref{fig:uce_blue_vangogh}} and \textbf{\Cref{fig:uce_star_vangogh}}. It turns out that the ability to generate these related concepts is highly preserved, which highlights that subdefense is different from direct token blocking of all related concepts. Instead, SubDefense blocks the composed embeddings, which represent the concept ``Van Gogh'' more accurately.


\clearpage

\section{More Attack Visualizations}
\label{appsec:attack_vis}

\begin{figure*}[h!]
    \centering
    \includegraphics[width=.9\linewidth]{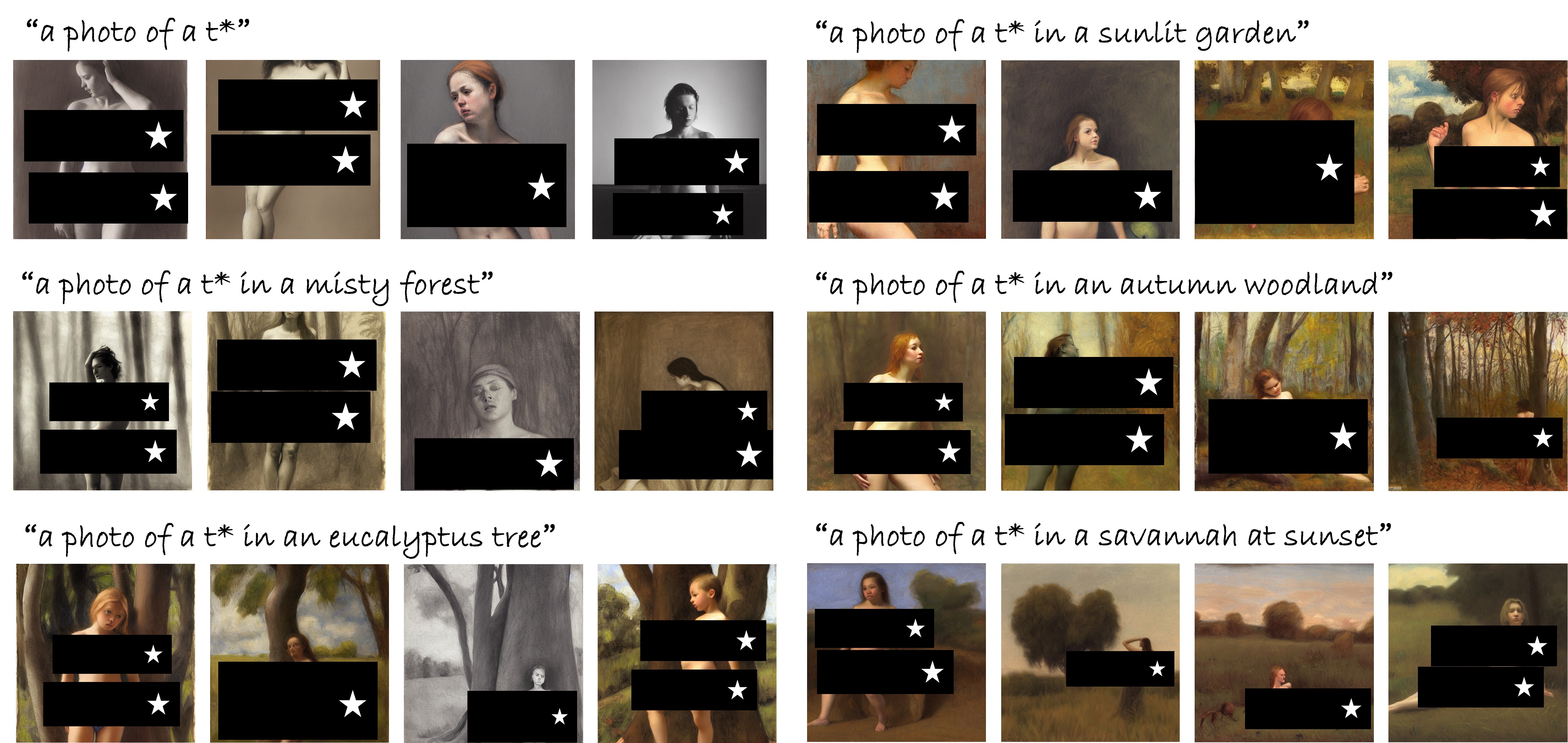}
    \caption{\textbf{Visualizing nudity attacking results on ESD.}}
    \label{fig:app_esd_nude1}
\end{figure*}

\begin{figure*}[h!]
    \centering
    \includegraphics[width=.9\linewidth]{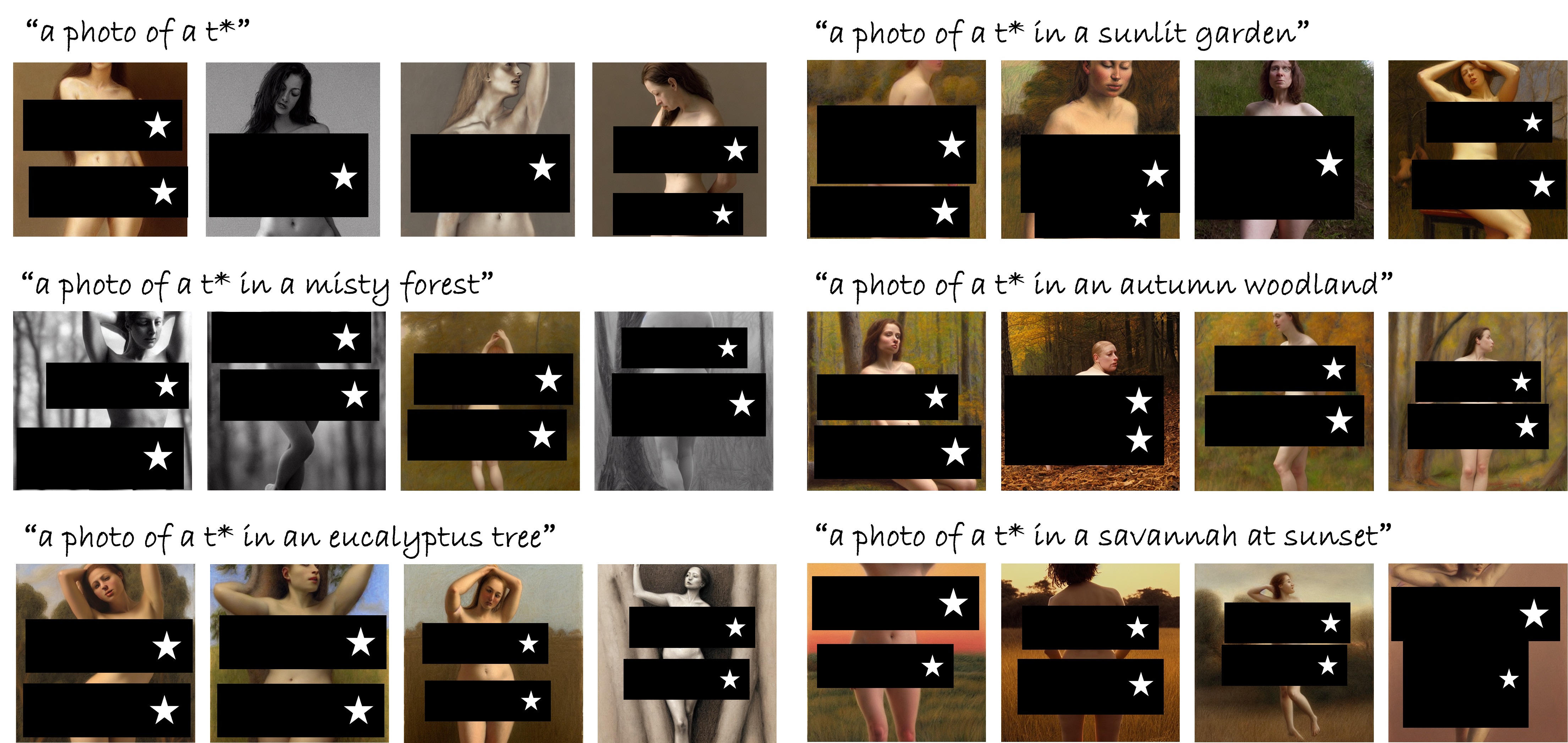}
    \caption{\textbf{Visualizing nudity attacking results on FMN.}}
    \label{fig:app_fmn_nude1}
\end{figure*}

\begin{figure*}[h!]
    \centering
    \includegraphics[width=.9\linewidth]{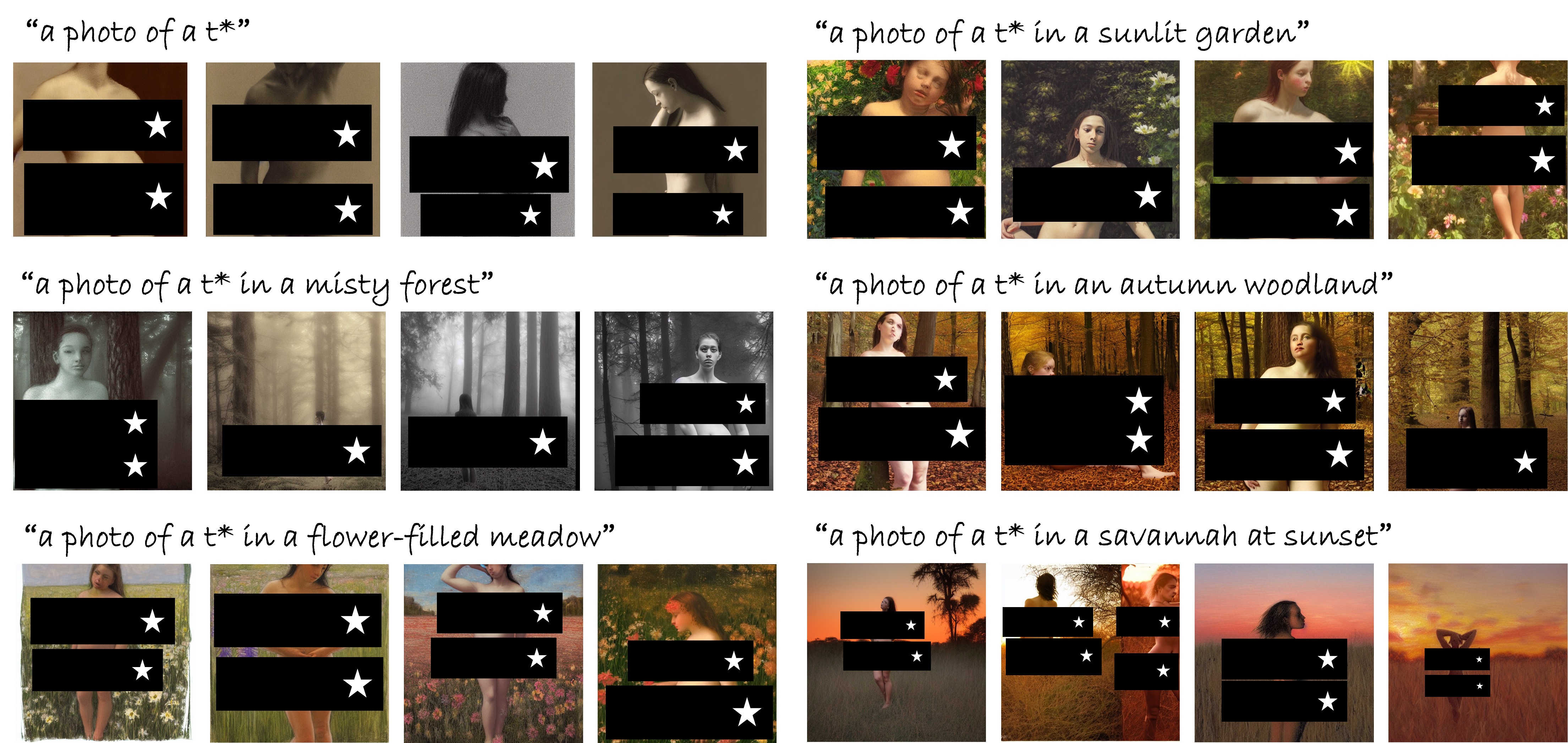}
    \caption{\textbf{Visualizing nudity attacking results on UCE.}}
    \label{fig:app_uce_nude1}
\end{figure*}

\begin{figure*}[h!]
    \centering
    \includegraphics[width=.9\linewidth]{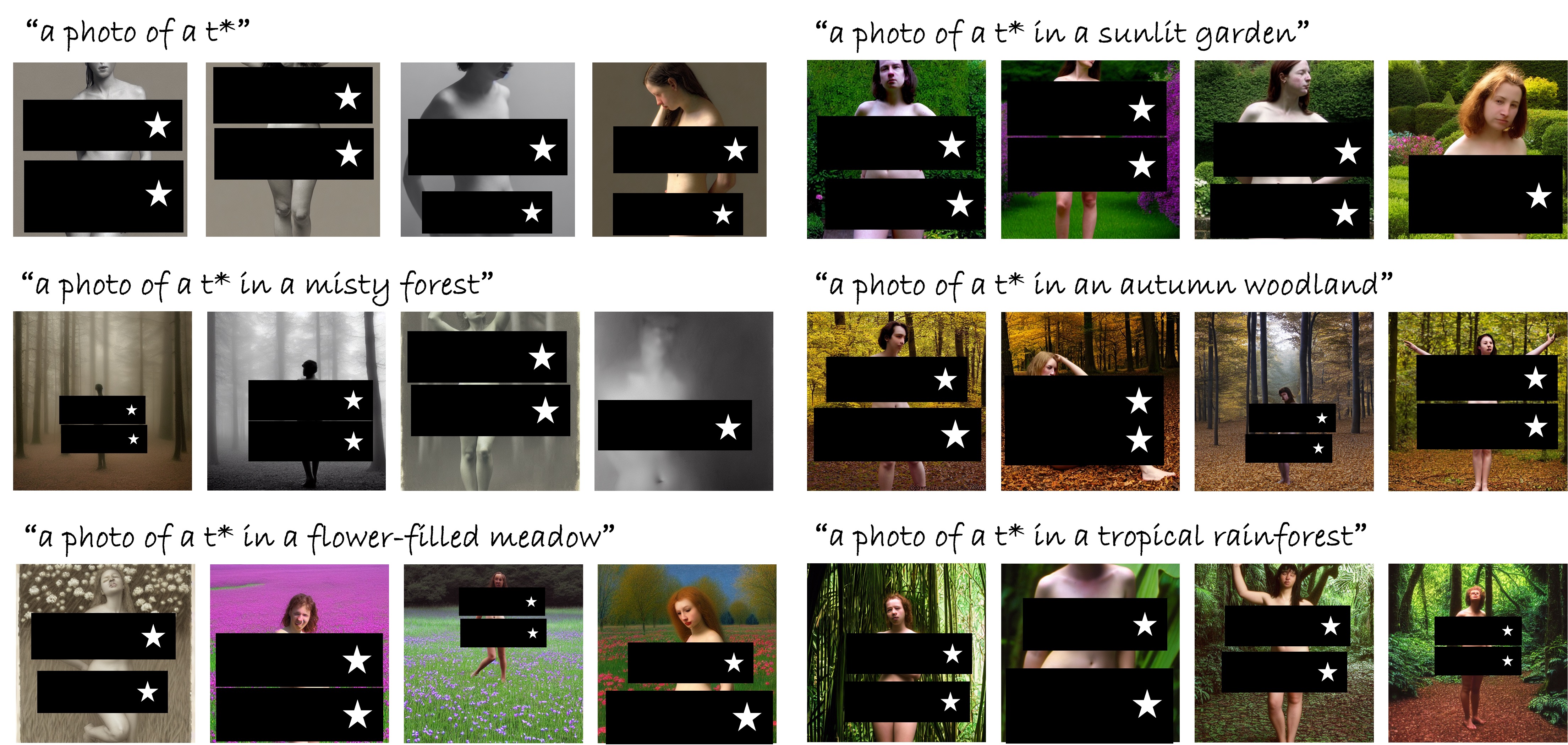}
    \caption{\textbf{Visualizing nudity attacking results on SPM.}}
    \label{fig:app_spm_nude1}
\end{figure*}

\begin{figure*}[h!]
    \centering
    \includegraphics[width=.9\linewidth]{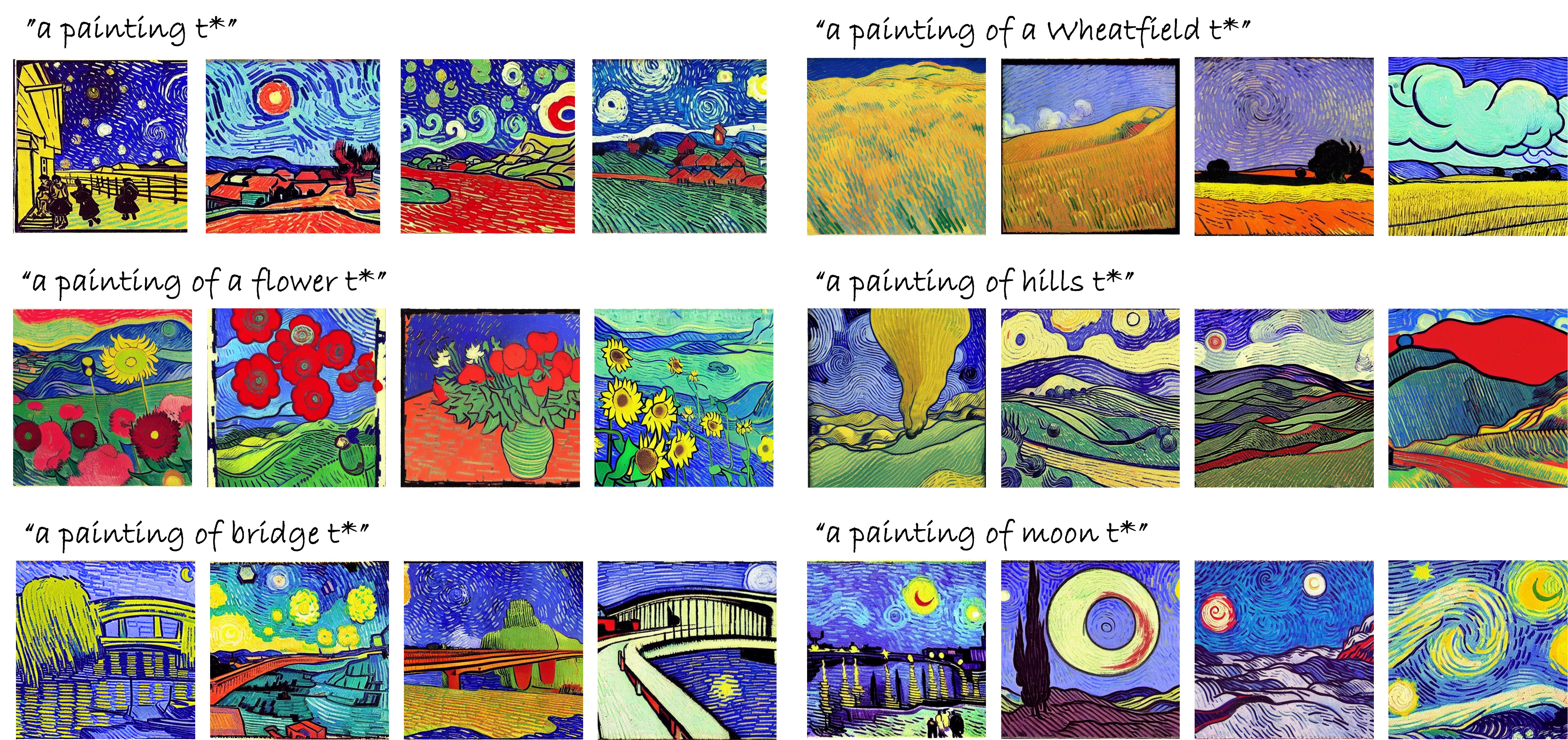}
    \caption{\textbf{Visualizing Van Gogh attacking results on ESD.}}
    \label{fig:app_esd_vangogh1}
\end{figure*}

\begin{figure*}[h!]
    \centering
    \includegraphics[width=.9\linewidth]{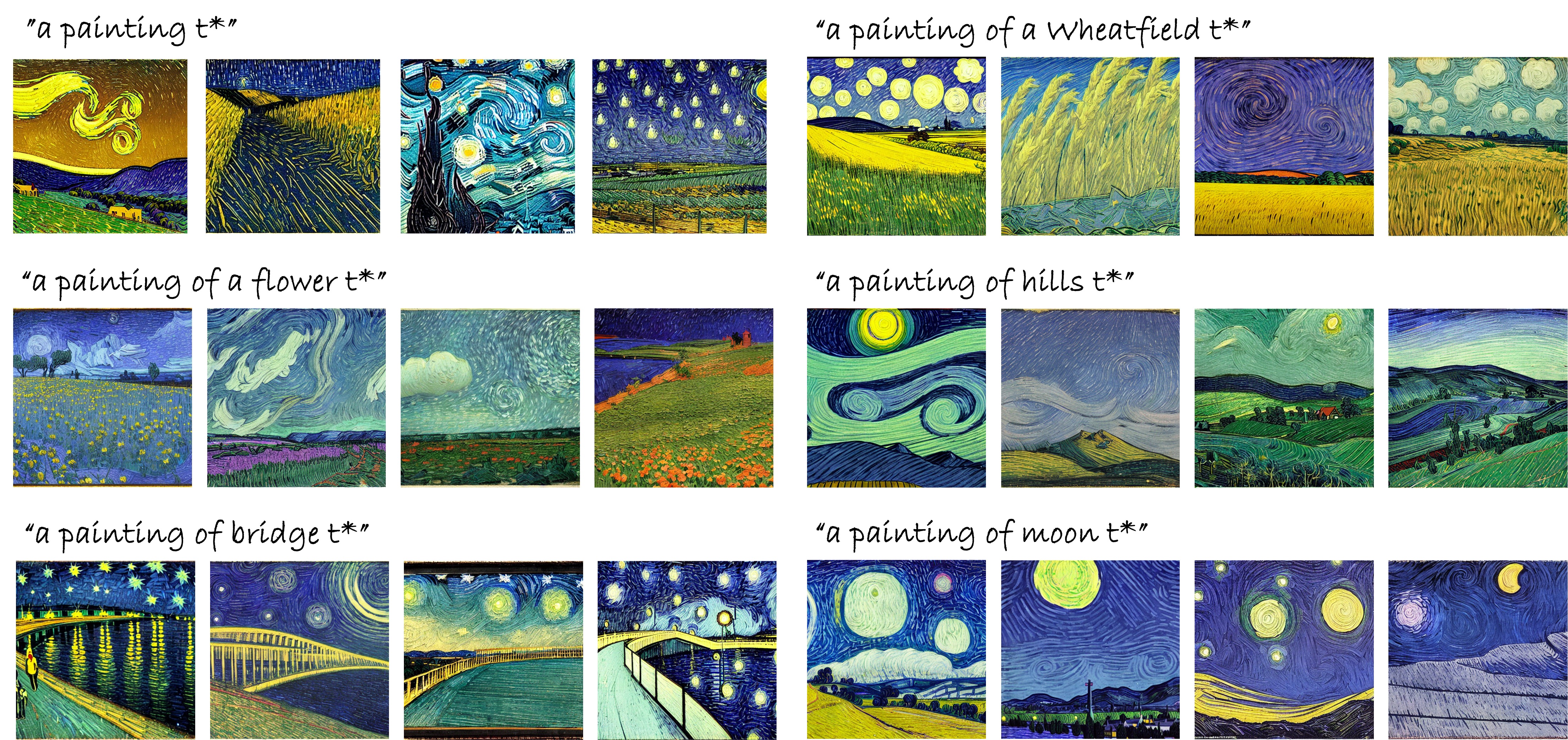}
    \caption{\textbf{Visualizing Van Gogh attacking results on FMN.}}
    \label{fig:app_fmn_vangogh1}
\end{figure*}

\begin{figure*}[h!]
    \centering
    \includegraphics[width=.9\linewidth]{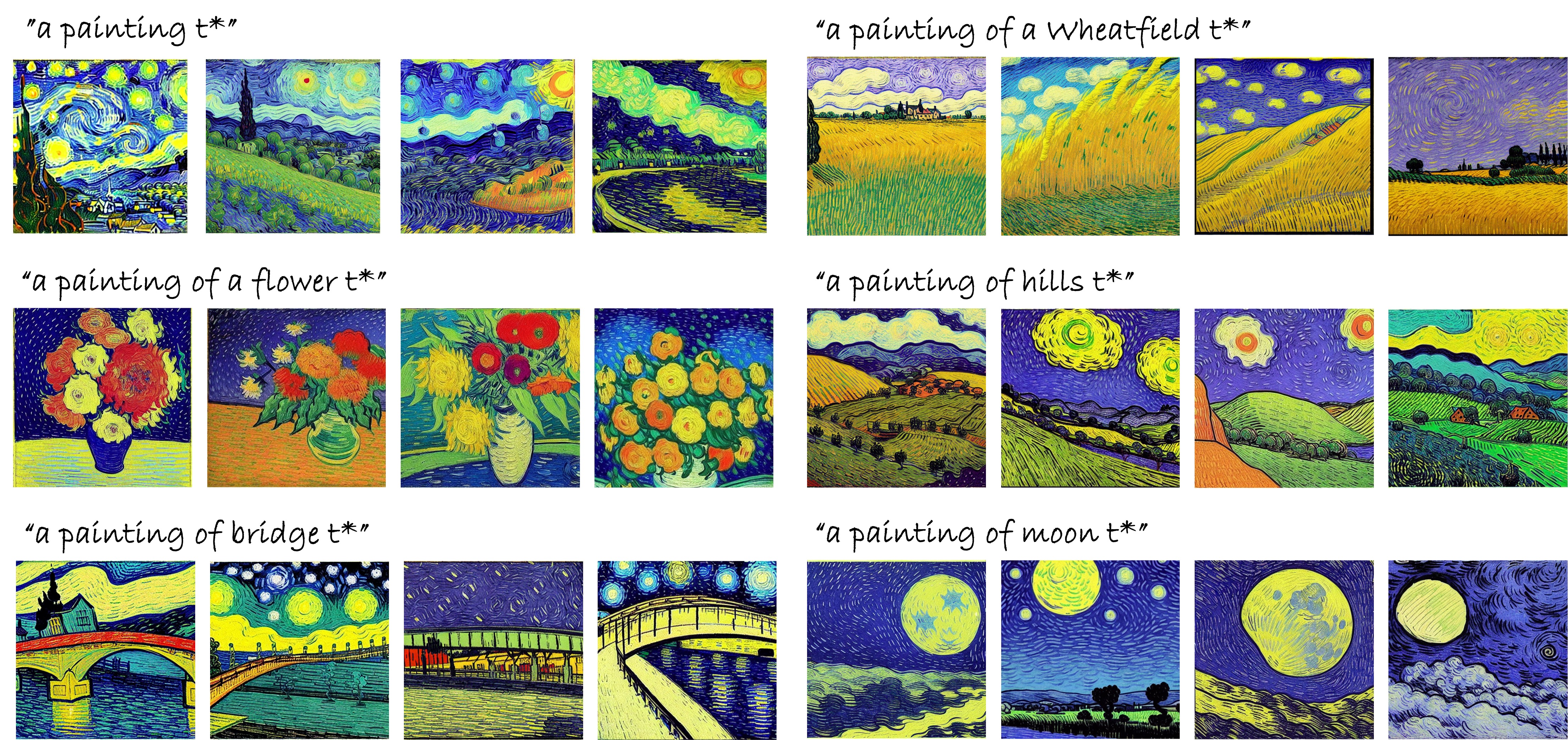}
    \caption{\textbf{Visualizing Van Gogh attacking results on UCE.}}
    \label{fig:app_uce_vangogh1}
\end{figure*}

\begin{figure*}[h!]
    \centering
    \includegraphics[width=.9\linewidth]{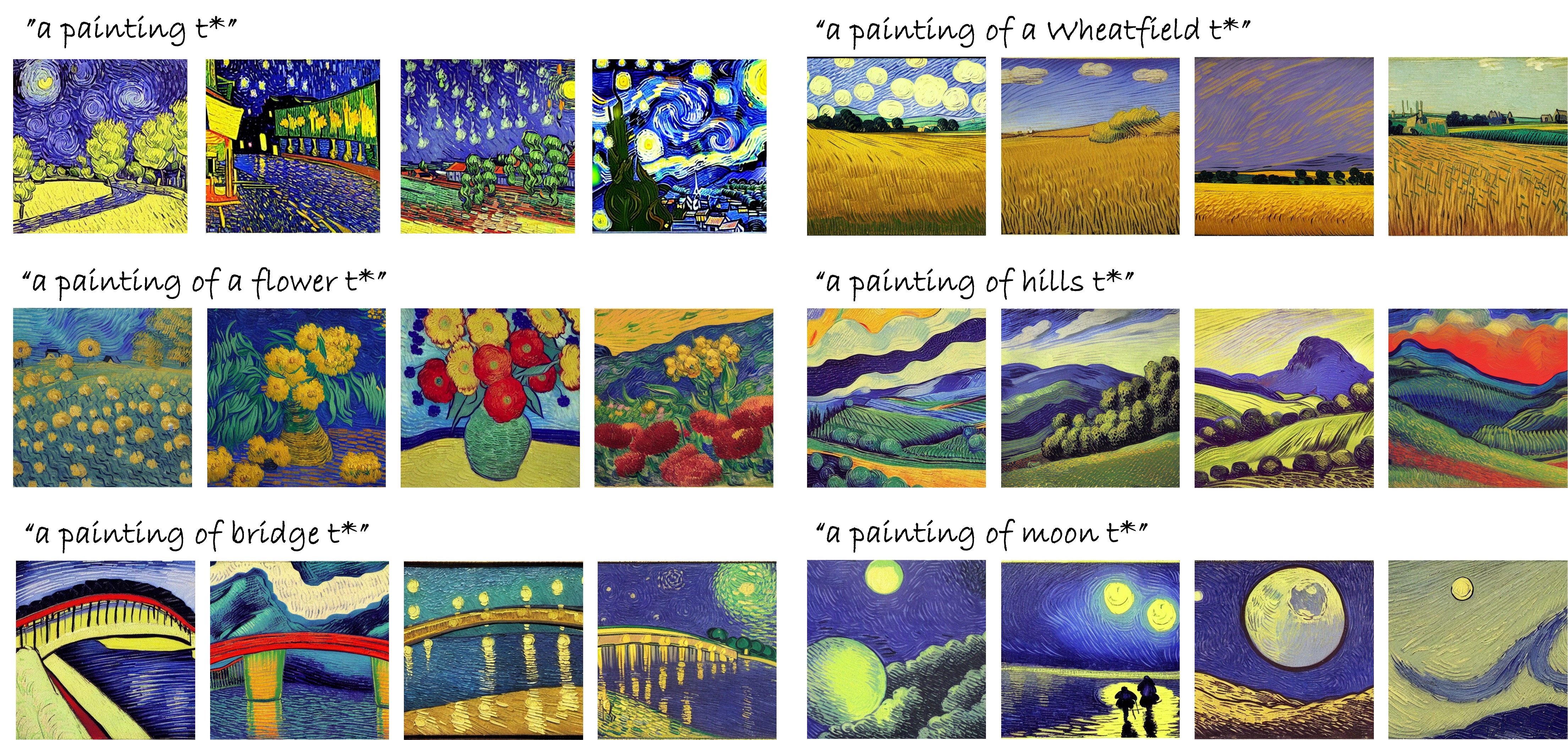}
    \caption{\textbf{Visualizing Van Gogh attacking results on SPM.}}
    \label{fig:app_spm_vangogh1}
\end{figure*}

\begin{figure*}[h!]
    \centering
    \includegraphics[width=.9\linewidth]{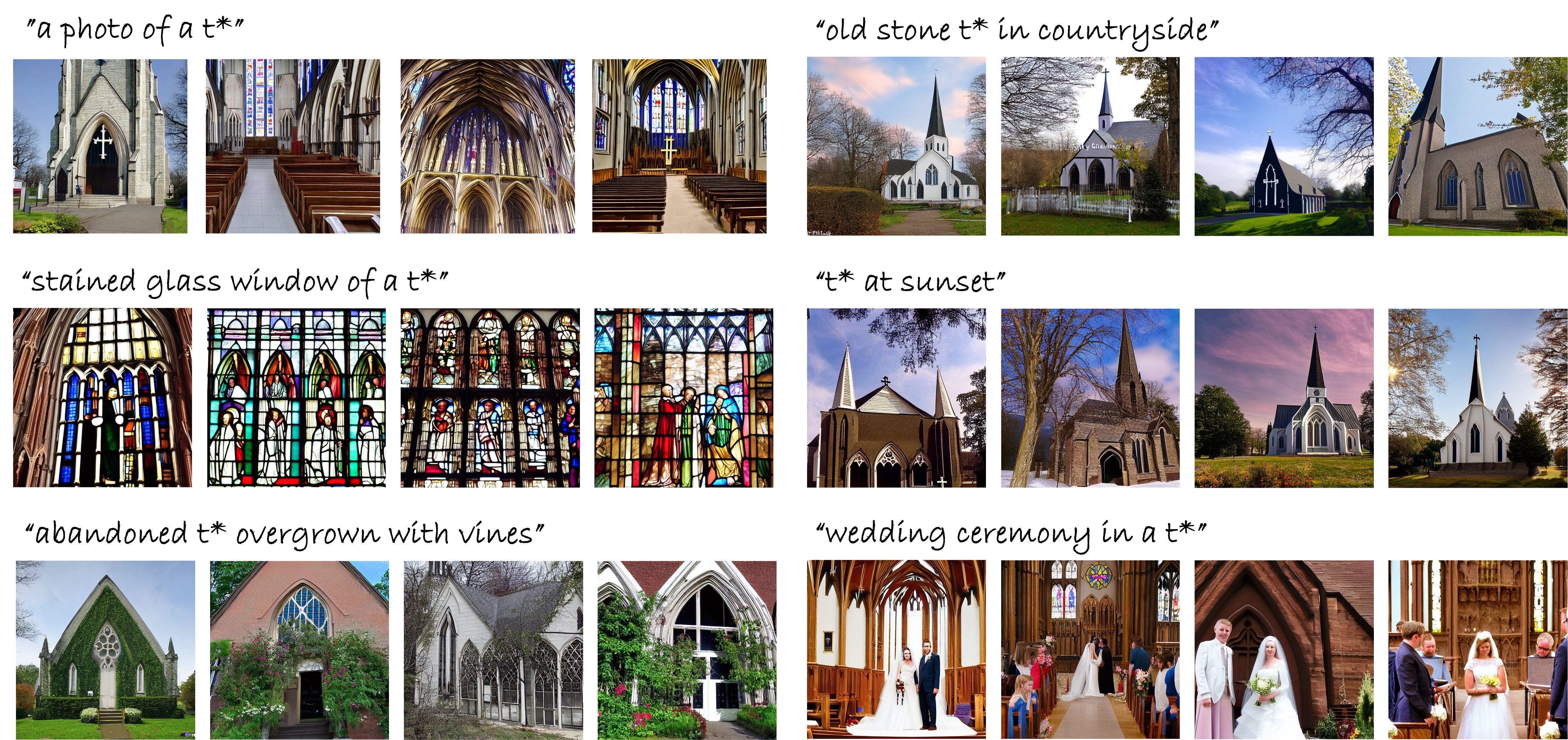}
    \caption{\textbf{Visualizing church attacking results on ESD.}}
    \label{fig:app_esd_church1}
\end{figure*}

\begin{figure*}[h!]
    \centering
    \includegraphics[width=.9\linewidth]{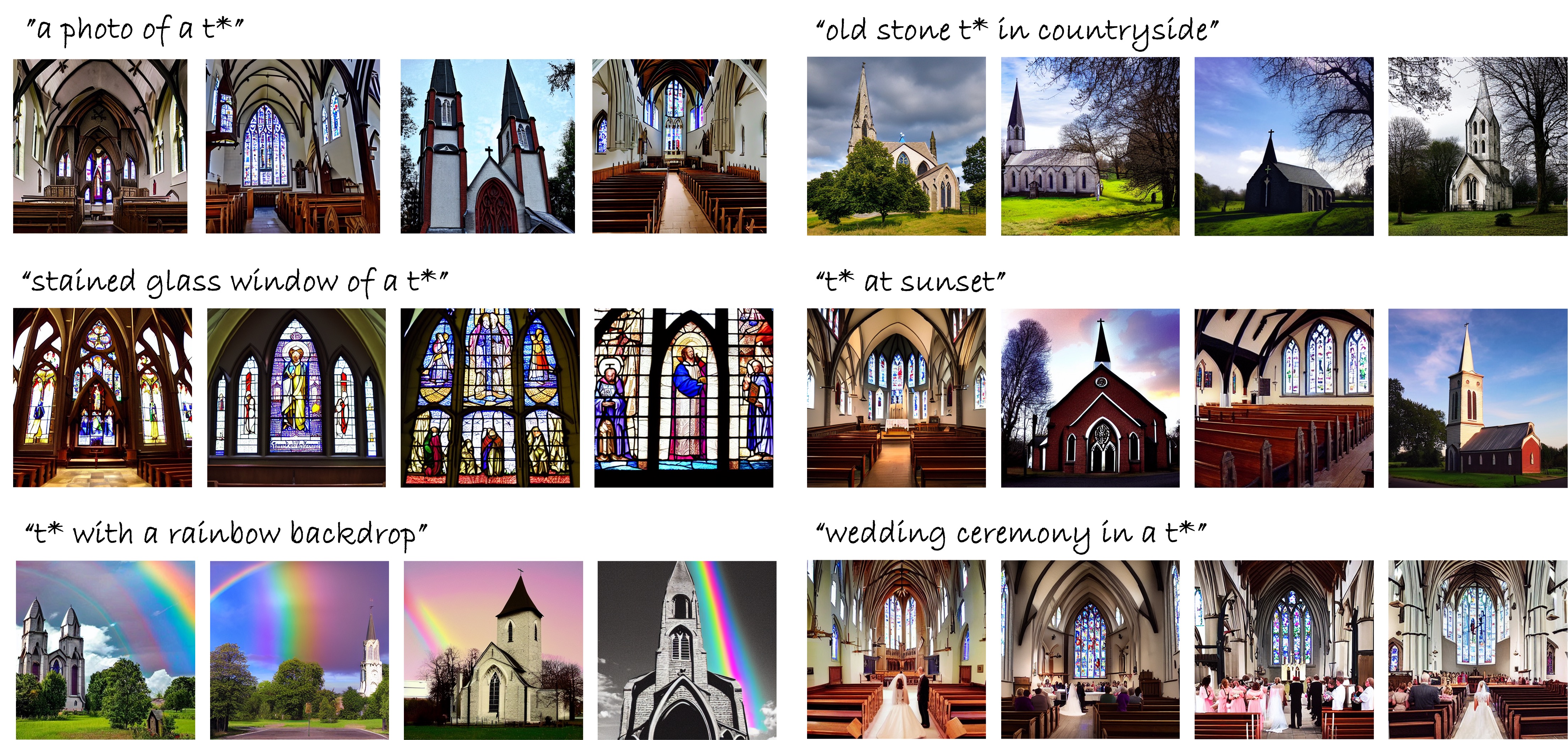}
    \caption{\textbf{Visualizing church attacking results on FMN.}}
    \label{fig:app_fmn_church1}
\end{figure*}

\begin{figure*}[h!]
    \centering
    \includegraphics[width=.9\linewidth]{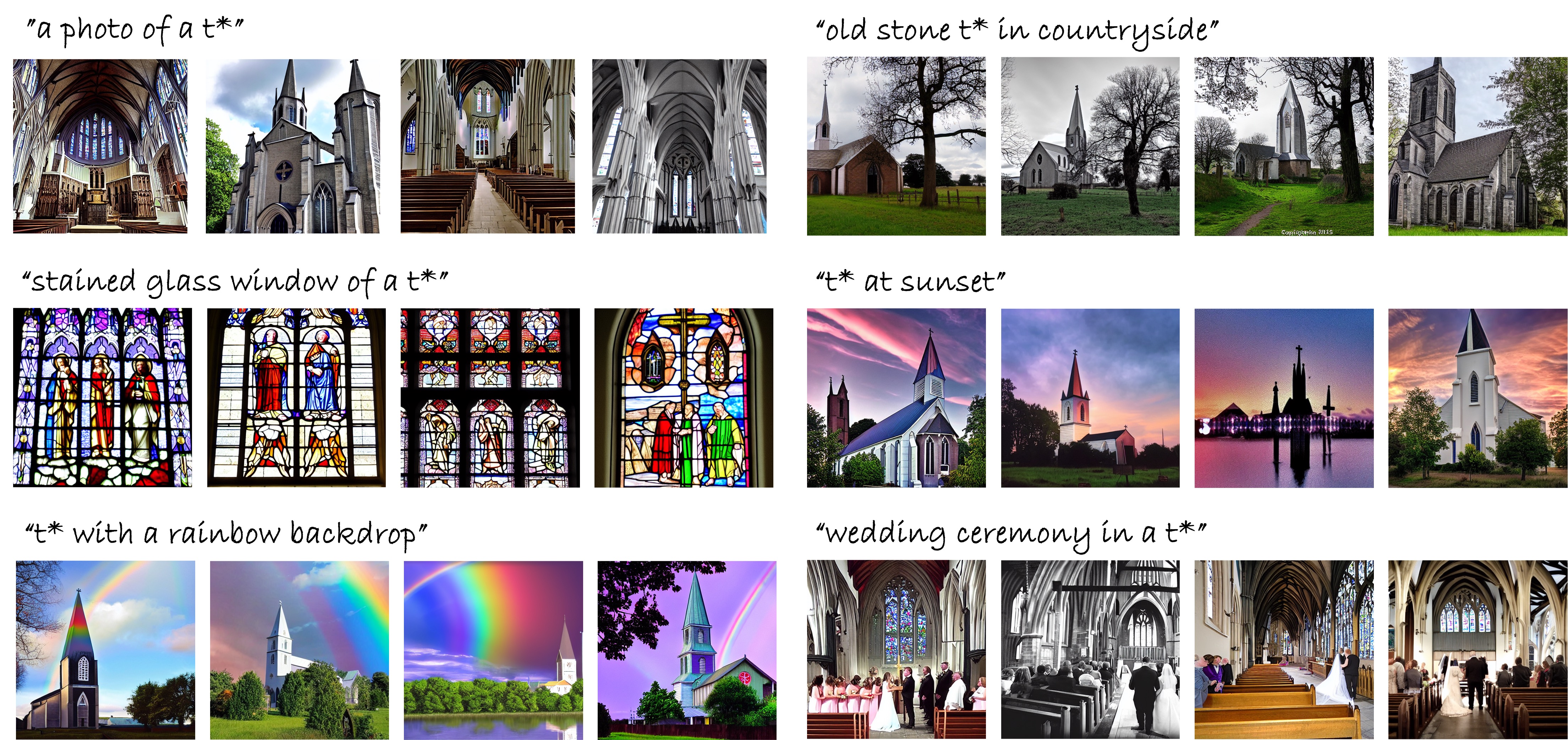}
    \caption{\textbf{Visualizing church attacking results on SPM.}}
    \label{fig:app_spm_church1}
\end{figure*}

\clearpage
\section{Future Directions}
\label{appsec_discussion}

\rebuttal{\paragraph{Limitations.} SubAttack and SubDefense assume white-box access to the diffusion pipeline and a small set of target-concept images. Our evaluation primarily studies Stable Diffusion v1.4 with a CLIP text encoder, and ASR depends on imperfect concept classifiers. SubDefense removes only recovered attack-accessible linear directions at the conditioning interface; adaptive attacks can still exploit residual information, and increasing projection rank trades robustness for both broad and local semantic utility. We therefore view SubDefense as a complementary post-hoc refinement, not complete erasure or a replacement for robust unlearning. Released artifacts will include responsible-use guidance discouraging offensive use against deployed systems.}

We identify the following future directions.
First, future research could explore feature representations in diffusion models beyond the linear structure, which may reveal richer mechanisms underlying unlearning.
Second, efficient, adaptive, and automatic methods could be designed to determine not only the number of blocked tokens but also the specific set to block, for example through learned importance scores or attention-based relevance.
Third, joint visual–textual embeddings could be investigated to better understand and defend against multimodal jailbreaks.
Fourth, as a reference point for defenses against CCE, SubDefense highlights a clear trade-off between robustness and utility; addressing this trade-off remains an important open challenge.
Fifth, extending SubDefense beyond CLIP-based architectures is another promising avenue. The core principle of identifying and nullifying harmful semantic directions in the conditional embedding space could be applied to other text encoders or even to models conditioned on alternative modalities.
Finally, examining residual associations without relying solely on predefined vocabularies may capture more implicit or nuanced concepts retained in unlearned models, improving interpretability and guiding the development of stronger defenses. \rebuttal{The recovered basis could also be incorporated during unlearning as interpretable hard-negative conditioning embeddings: one could alternate between probing for residual directions and updating model weights to suppress target generation along them while preserving benign outputs. We leave evaluation of this training-time extension to future work.}

\end{document}